# Resolution scaling governs DINOv3 transfer performance in chest radiograph classification

Soroosh Tayebi Arasteh (1,2,3,4), Mina Shaigan (5), Christiane Kuhl (2), Jakob Nikolas Kather (6,7,8), Sven Nebelung* (1,2), Daniel Truhn* (1,2)

(1) Lab for AI in Medicine, RWTH Aachen University, Aachen, Germany.
(2) Department of Diagnostic and Interventional Radiology, University Hospital RWTH Aachen, Aachen, Germany.
(3) Department of Urology, Stanford University, Stanford, CA, USA.
(4) Department of Radiology, Stanford University, Stanford, CA, USA.
(5) Institute for Computational Genomics, Joint Research Center for Computational Biomedicine, University Hospital RWTH Aachen, Aachen, Germany.
(6) Else Kroener Fresenius Center for Digital Health, Technical University Dresden, Dresden, Germany.
(7) Department of Medicine I, University Hospital Dresden, Dresden, Germany.
(8) National Center for Tumor Diseases (NCT), University Hospital Heidelberg, Heidelberg, Germany.

* Sven Nebelung and Daniel Truhn are shared senior authors.

## Abstract

Self-supervised learning (SSL) has improved visual representation learning, but its value in chest radiography remains uncertain. DINOv3 extends earlier SSL models through Gram-anchored self-distillation and explicit high-resolution adaptation. Whether these changes improve transfer learning for chest radiograph classification has not been established. We benchmarked DINOv3 against DINOv2 and supervised ImageNet initialization across seven chest radiograph datasets comprising 816,183 radiographs from pediatric and adult cohorts. ViT-B/16 and ConvNeXt-B were evaluated under full fine-tuning at 224 × 224 and 512 × 512 pixels, with targeted 1024 × 1024 experiments on three cohorts. Additional analyses examined parameter-efficient adaptation, synthetic label corruption, external validation, frozen 7B features, and computational efficiency. The primary outcome was mean AUROC across labels. In adult cohorts, DINOv3 did not consistently outperform DINOv2 at 224 × 224 pixels, but became the strongest initialization at 512 × 512 pixels, especially with ConvNeXt-B. Gains were greatest for small focal and boundary-dependent abnormalities, whereas large-structure findings changed little. The pediatric cohort showed no significant benefit from DINOv3, higher resolution, or backbone choice. Scaling to 1024 × 1024 rarely improved performance and markedly increased computational cost. ConvNeXt-B remained superior to ViT-B/16 under both full and parameter-efficient adaptation. External validation preserved the 512 × 512 DINOv3 advantage, whereas synthetic label corruption showed that this benefit should not be interpreted simply as superior noise robustness. Frozen DINOv3-7B features underperformed relative to fully adapted 86 to 89M-parameter backbones. For adult chest radiograph classification, DINOv3 provides its most reliable benefit at 512 × 512 pixels, particularly with ConvNeXt-B. Fully adapted mid-sized models at 512 × 512 pixels provided the best performance-cost trade-off in our benchmark.

**Correspondence**
Soroosh Tayebi Arasteh, Dr.-Ing., Dr. rer. medic. (soroosh.arasteh@rwth-aachen.de)
Lab for AI in Medicine, Department of Diagnostic and Interventional Radiology, University Hospital RWTH Aachen
Pauwelsstr. 30, 52074 Aachen, Germany

---

# Introduction

Chest radiography is the most widely performed imaging examination worldwide and a first-line tool for detecting pulmonary and cardiac abnormalities. Subtle or low-contrast findings, such as interstitial lung disease, reticular changes, or diffuse pulmonary opacification, can be difficult to recognize, motivating the use of automated analysis to assist interpretation and triage. Artificial intelligence (AI) has become an integral component of medical imaging[1–3], with chest radiographs serving as one of the most extensively studied modalities for evaluating new algorithms[4–6]. Early advances relied on supervised deep learning, where models were pretrained on large annotated datasets such as ImageNet[7,8] and then fine-tuned for radiographic tasks. Although this strategy improved performance compared with training from scratch, it remains constrained by the domain mismatch between natural and medical images and by its dependence on costly manual annotations. Constructing large, expertly labeled radiograph collections continues to be a major bottleneck, motivating the exploration of label-efficient alternatives.

Self-supervised learning (SSL) has emerged as a promising response to this challenge. By constructing pretraining objectives that do not depend on manual labels, SSL enables the use of massive unlabeled datasets to learn transferable visual representations[9,10]. Methods such as MoCo[11], SimCLR[12], BYOL[13], and SwAV[14] have shown strong performance on natural images, and their application to medical imaging has yielded encouraging gains in classification and segmentation tasks. However, many medical studies remain limited in scale, often involving tens rather than hundreds of thousands of radiographs, and important questions remain regarding the robustness, generalizability, and practical operating conditions of SSL for clinical imaging[15]. These questions are especially relevant in chest radiography, where diagnostically important structures may occupy only a small fraction of the image and where label quality varies widely across datasets.

The introduction of transformer-based[16] architectures has further accelerated progress. Vision transformers (ViTs)[17] and modern convolutional backbones such as ConvNeXt[18] have reshaped representation learning in computer vision, and their transfer to radiology has highlighted the value of flexible, high-capacity architectures for medical data. Within this landscape, the DINO[19] family of SSL methods has been particularly influential. DINOv2[20], pretrained on hundreds of millions of natural images, established itself as a strong general-purpose representation learner. In our prior work[21], we showed that DINOv2 could match or surpass supervised ImageNet pretraining for chest radiograph classification. Meta's DINOv3[22] builds on this foundation by introducing Gram-anchored self-distillation and explicit high-resolution adaptation, with the goal of preserving fine-grained visual information and improving scaling to larger input sizes. Recent studies[23] have begun to benchmark DINOv3 across multiple medical imaging tasks and modalities[24,25], including chest radiograph classification, but important questions remain regarding its task-specific operating conditions and interpretation in chest radiography. These design choices are especially relevant to chest radiographs, which are acquired at far higher native resolution than is typically used during downstream model training. Yet several clinically important questions remain unresolved: whether DINOv3's benefits emerge only at higher resolution, whether they generalize beyond within-dataset splits, whether they extend equally across adult and pediatric cohorts, whether they depend on backbone architecture or adaptation regime, and whether performance gains on weakly labeled datasets reflect better transfer or simply greater tolerance to label noise.

Here, we present a large-scale, chest-radiograph-focused evaluation of DINOv3 for chest radiograph classification across seven datasets comprising 816,183 anteroposterior (AP) or posteroanterior (PA) radiographs from public and internal cohorts (see **Figure 1**). Our benchmark spans pediatric and adult populations, two backbone families (ViT-B/16 and ConvNeXt-B), and input resolutions from 224 × 224 to 1024 × 1024 pixels, with the 1024-pixel setting used as a targeted probe of further scaling. In addition to the core comparison against DINOv2 and ImageNet initialization, we examine frozen representations from the 7B-parameter DINOv3 teacher, test whether the ConvNeXt-B vs. ViT-B/16 difference persists under parameter-efficient adaptation, assess the effect of controlled synthetic label corruption, and perform external validation across institutions using a harmonized label space. This design allows us to move beyond a simple benchmark and ask when DINOv3 helps, where it fails, and how those gains should be interpreted in practice. As we show, DINOv3 yields its most reproducible benefit in adult cohorts at 512 × 512 pixels, especially with ConvNeXt-B, whereas the pediatric cohort follows a distinct transfer pattern, scaling beyond 512 × 512 provides little additional value for markedly higher cost, and frozen billion-parameter features remain inferior to task-adapted mid-sized models. We expect that these results provide practical guidance for applying modern SSL to chest radiograph analysis and clarify the conditions under which high-resolution self-supervised representations are most useful in radiology.

# Results

We benchmarked ImageNet[7], DINOv2[20], and DINOv3[22] initializations across seven chest radiograph datasets spanning pediatric and adult populations, including six public cohorts and one internal cohort: Pedi-CXR[26] (n = 9,125 radiographs; 3 labels), VinDr-CXR[27] (n = 18,000; 11 labels), ChestX-ray14[28] (n = 112,120; 14 labels), PadChest[29] (n = 110,525; 17 labels), CheXpert[30] (n = 157,865; 10 labels), MIMIC-CXR[6] (n = 215,187; 10 labels), and UKA-CXR[21,31–35] (n = 193,361; 6 labels), for a total of 816,183 radiographs. Full fine-tuning experiments covered two backbone families, ViT-B/16 and ConvNeXt-B, at universal input resolutions of 224 × 224 and 512 × 512 pixels, with targeted 1024 × 1024 evaluation on three representative cohorts. Additional follow-up analyses examined parameter-efficient adaptation, synthetic label corruption, external validation, frozen 7B features, and computational efficiency. Dataset characteristics and study design are summarized in **Table 1**. Unless otherwise specified, all performance values reported in the Results section refer to mean area under the receiver operating characteristic curve (AUROC) across labels and are given as mean ± standard deviation [95% confidence interval] in percent.

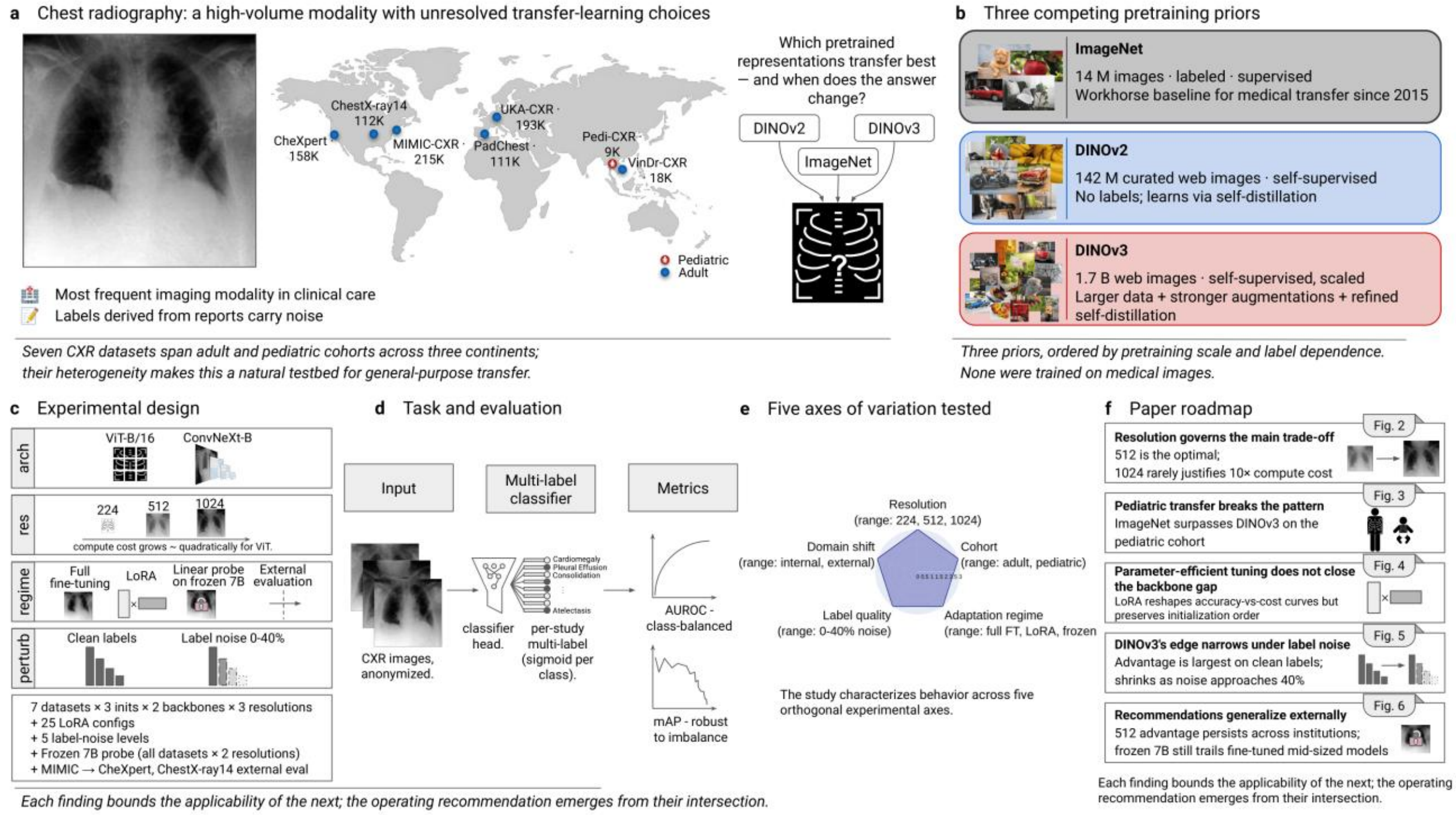

**Figure 1**: Study overview and analytical framework. **a** Chest radiography as a high-volume imaging modality and natural testbed for transfer learning. A representative adult posteroanterior chest radiograph is shown together with the geographic distribution of the seven study cohorts: Pedi-CXR (n = 9,125), VinDr-CXR (n = 18,000), ChestX-ray14 (n = 112,120), PadChest (n = 110,525), CheXpert (n = 157,865), MIMIC-CXR (n = 215,187), and UKA-CXR (n = 193,361). Marker color distinguishes pediatric (red) from adult (blue) cohorts. The panel also frames the central study question of which pretrained representations transfer best to chest radiograph classification and under which conditions that answer changes. **b** Three competing pretraining priors evaluated in this work. ImageNet is shown as a labeled supervised natural-image prior, DINOv2 as a self-supervised web-scale prior, and DINOv3 as a larger self-supervised web-scale prior with refined self-distillation and high-resolution adaptation. None of the evaluated checkpoints were pretrained on medical images. **c** Experimental design. The benchmark varies four main dimensions: architecture (ViT-B/16 and ConvNeXt-B), input resolution (224 x 214, 512 x 512, and targeted 1024 x 1024 pixels), adaptation regime (full fine-tuning, LoRA, frozen 7B linear probing, and external evaluation), and label condition (clean labels versus 0-40% synthetic label noise). The factorial scope of the study is summarized in the boxed panel. **d** Downstream task and evaluation. Chest radiographs were analyzed in a per-study multilabel classification setting with sigmoid outputs per class. Performance was assessed primarily by AUROC and complemented by mAP to account for class imbalance. **e** Five orthogonal axes of variation examined in the study: resolution, cohort type, adaptation regime, label quality, and domain shift. These axes define the conditions under which DINOv3 transfer was evaluated. **f** Paper roadmap.

# DINOv3 gains emerge at 512 × 512 pixels in adult chest radiograph cohorts

Across the six adult cohorts, the relative behavior of ImageNet, DINOv2, and DINOv3 depended strongly on input resolution (**Figure 2, Table 2, Supplementary Table 1, Supplementary Figures 1-7**). At 224 × 224 pixels, DINOv3 did not consistently outperform DINOv2. In the ViT-B/16 setting, DINOv3 was numerically higher in AUROC only in VinDr-CXR (90.2 ± 0.6 vs. 89.2 ± 0.7), whereas DINOv2 remained equal or higher in the other adult datasets, including PadChest (88.0 ± 0.2 vs. 87.4 ± 0.2; p = 0.003), CheXpert (80.3 ± 0.2 vs. 80.0 ± 0.2; p = 0.003), and UKA-CXR (88.1 ± 0.1 vs. 88.0 ± 0.1; p = 0.008). Thus, at standard resolution, DINOv3 improved over ImageNet in all adult ViT-B/16 cohorts (p ≤ 0.012), but did not yet establish a clear advantage over DINOv2.

**Table 1:** Characteristics of the datasets and study protocol utilized in this study. Summary of patient cohorts, image counts, demographics, and label sets for all seven datasets: Pedi-CXR, VinDr-CXR, ChestX-ray14, PadChest, CheXpert, MIMIC-CXR, and UKA-CXR. Reported values include the number of patients and radiographs, split into total, training, validaiton, and test sets, as well as patient age distributions (mean ± standard deviation (SD) and range) and sex ratios (female/male, given). The labels used for multi-label classification are listed as defined in each dataset. Dataset locations and the distribution of image projections (anteroposterior (AP) vs. posteroanterior (PA)) are also reported. Whenever available, the "no finding" label was preserved as a separate category to indicate a completely normal radiograph without any imaging abnormality, not merely the absence of the labels considered in this study. Patient-wise splits were used in all datasets to ensure no overlap between training and test cohorts. Only AP or PA images are considered in this study. For Pedi-CXR and VinDr-CXR, no validation split was created because patient identifiers were unavailable, precluding leakage-safe subdivision of the provided training sets. N/A = not available. * The youngest patients in the Pedi-CXR, PadChest, and UKA-CXR datasets were infants younger than six months. Missing demographic information was handled by exclusion from the corresponding analyses: age information was unavailable for 29 patients in UKA-CXR, 10 patients in ChestX-ray14, 13,772 images in VinDr-CXR, and one image in CheXpert, while sex information was unavailable for 9,392 images in VinDr-CXR.

| Field | Pedi-CXR | VinDr-CXR | ChestX-ray14 | PadChest | CheXpert | MIMIC-CXR | UKA-CXR |
|---|---|---|---|---|---|---|---|
| Cohort type | Pediatric | Adult | Adult | Adult | Adult | Adult | Adult |
| Age (years)<br>Mean ± SD<br>Range | 4 ± 3<br>(0*, 10) | 54 ± 18<br>(2, 90) | 46 ± 17<br>(1, 95) | 56 ± 21<br>(0*, 105) | 60 ± 18<br>(18, 90) | N/A | 66 ± 16<br>(0*, 111) |
| Sex (female/male) | 42% / 58% | 47% / 53% | 44% / 56% | 50% / 50% | 41% / 59% | N/A | 35% / 65% |
| Number of patients (n) | N/A | N/A | 30,805 | 67,205 | 57,872 | 62,094 | 54,176 |
| Number of radiographs (n)<br>Total<br>Training<br>Validation<br>Test | 9,125<br>7,728<br>0<br>1,397 | 18,000<br>15,000<br>0<br>3,000 | 112,120<br>77,870<br>8,654<br>25,596 | 110,525<br>79,697<br>8,783<br>22,045 | 157,865<br>115,449<br>13,098<br>29,318 | 215,187<br>153,255<br>18,139<br>43,793 | 193,361<br>137,902<br>15,353<br>40,106 |
| Number of labels used (n) | 3 | 11 | 14 | 17 | 10 | 10 | 6 |
| Labels used in the main study | No finding, pneumonia, bronchitis/bronchiolitis | Cardiomegaly, pleural effusion, pneumonia, atelectasis, no finding, consolidation, pneumothorax, pleural thickening, lung opacity, pulmonary fibrosis, nodule/mass | Cardiomegaly, effusion, pneumonia, atelectasis, no finding, consolidation, pneumothorax, fibrosis, emphysema, hernia, pleural thickening, edema, nodule, mass | Cardiomegaly, pleural effusion, pneumonia, atelectasis, no finding, consolidation, pneumothorax, emphysema, hernia, scoliosis, congestion, aortic elongation, kyphosis, COPD signs, pleural thickening, nodule mass, infiltrates | Cardiomegaly, pleural effusion, pneumonia, atelectasis, no finding, consolidation, pneumothorax, lung opacity, lung lesion, fracture | Cardiomegaly, pleural effusion, pneumonia, atelectasis, no finding, consolidation, pneumothorax, lung opacity, lung lesion, fracture | Cardiomegaly, congestion, pleural effusion, pneumonic infiltrates, atelectasis, no finding |
| Label provenance | Radiologist annotation | Radiologist annotation | Rule-based NLP from reports | Partially radiologist annotation + partially rule-based NLP | Rule-based NLP from reports | Rule-based NLP from reports | Radiologist annotation |
| Location | Vietnam | Vietnam | USA | Spain | USA | USA | Germany |
| Projections (AP/PA) | 0% / 100% | 0% / 100% | 40% / 60% | 17% / 83% | 85% / 15% | 58% / 42% | 47% / 53% |
| Public or internal | Public | Public | Public | Public | Public | Public | Internal |

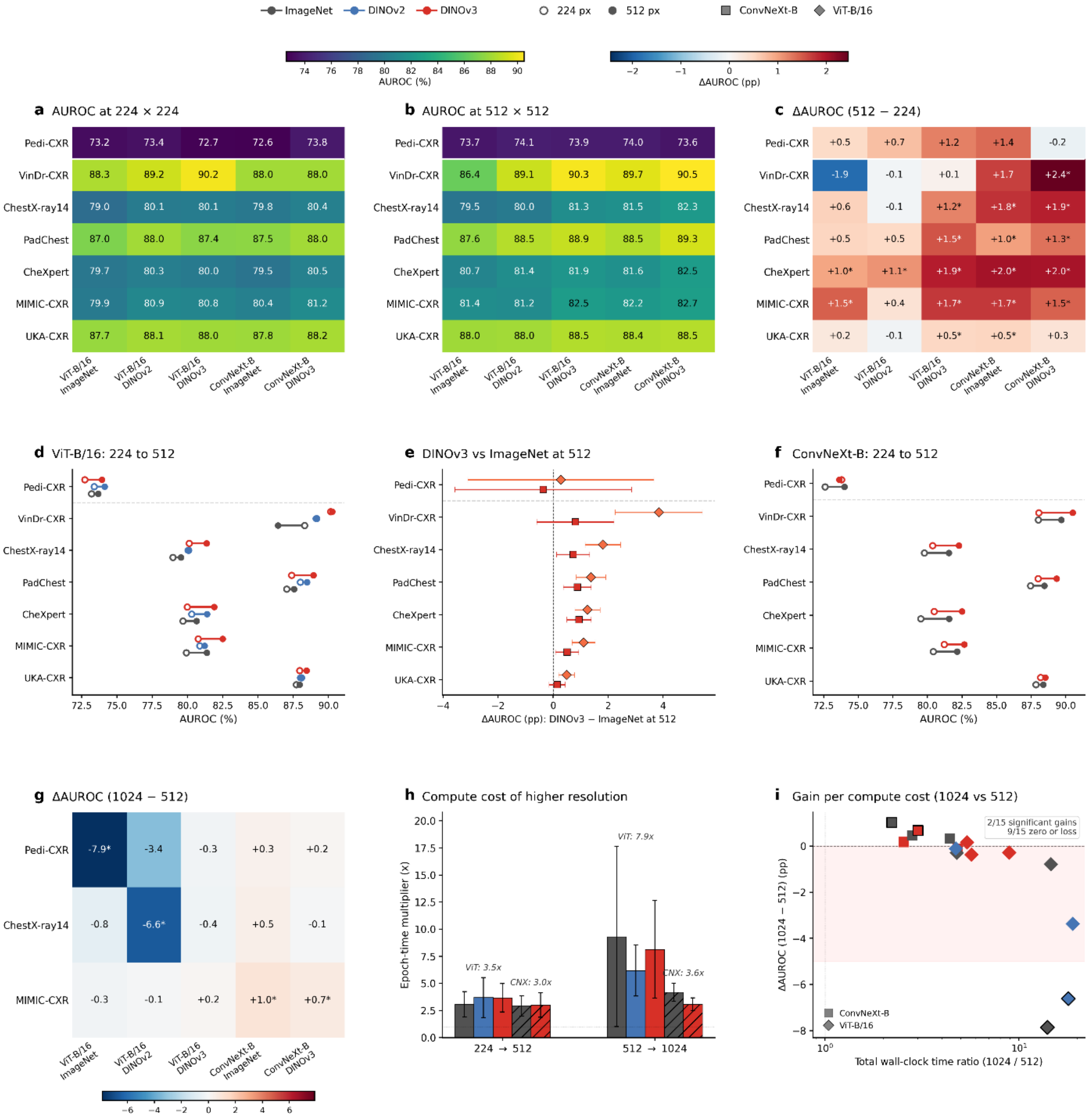

**Figure 2**: Resolution scaling and computational cost-benefit analysis. **a** AUROC at 224×224 resolution across seven chest radiograph datasets and five configurations (three ViT-B/16 initializations plus two ConvNeXt-B initializations). Heatmap cells show mean AUROC (%). The horizontal line separates Pedi-CXR (top) from adult datasets. **b** AUROC at 512×512 resolution for the same configurations. DINOv3 initializations achieve the highest performance across most datasets at this resolution. **c** Delta AUROC (512×512 minus 224×224) showing performance gains from quadrupling resolution. Red indicates positive gains; blue indicates losses. Stars mark statistically significant changes (non-overlapping 95% CIs). Most configurations show 1-2 pp gains; notable exceptions include Pedi-CXR where several ViT configurations show minimal or negative change, and ChestX-ray14 where gains exceed 2 pp for some configurations. **d** Per-dataset dumbbell plot for ViT-B/16 showing 224×224 (open circles) to 512×512 (filled circles) transitions across three initializations. Pedi-CXR shows atypical behavior with ImageNet outperforming self-supervised initializations at 512×512. **e** Forest plot comparing DINOv3 vs. ImageNet at 512×512 across all datasets. Squares represent ConvNeXt-B; diamonds represent ViT-B/16. Error bars show 95% CIs on the difference. DINOv3 outperforms ImageNet on six of seven datasets; Pedi-CXR is the exception. **f** ConvNeXt-B resolution scaling from 224×224 to 512×512 for ImageNet and DINOv3 initializations. Both initializations show consistent gains across all datasets. **g** Delta AUROC (1024×1024

minus 512×512) for the three datasets with 1024 resolution data (Pedi-CXR, ChestX-ray14, MIMIC-CXR). Stars indicate significant changes. Only 2 of 15 cells show significant positive gains (both MIMIC-CXR ConvNeXt configurations: +1.0 pp for ImageNet, +0.7 pp for DINOv3). Striking negative outliers include Pedi-CXR ViT-B/16 ImageNet (−7.9 pp) and ChestX-ray14 ViT-B/16 DINOv2 (−6.6 pp). **h** Computational cost of higher resolution quantified as epoch-time multipliers. Bars show mean scaling ratios across datasets with error bars representing standard deviation across configurations. 224→512 scaling averages 3.5× for ViT-B/16 and 3.0× for ConvNeXt-B. 512→1024 scaling shows greater variance: 7.9× for ViT-B/16 (SD 5.1) vs. 3.6× for ConvNeXt-B (SD 0.9). Summary annotations show aggregate means per backbone. **i** Cost-benefit analysis for 1024×1024 vs. 512×512. Each point represents one (dataset, backbone, initialization) combination. X-axis (log scale) shows total wall-clock time ratio (time to convergence at 1024 divided by time at 512); y-axis shows AUROC gain in percentage points. Marker shape indicates backbone (squares for ConvNeXt-B, diamonds for ViT-B/16); color indicates initialization; thick black edges mark statistically significant changes. Red shaded region indicates loss zone (more compute, zero or negative gain). Text annotation: "2/15 significant gains, 9/15 zero or loss." Most points cluster at high compute cost (5-20× time multiplier) with near-zero or negative AUROC change, demonstrating that scaling to 1024×1024 yields limited returns for the majority of configurations. Results are shown for all seven datasets: Pedi-CXR (training n=7,728; test n=1,397), VinDr-CXR (training n=15,000; test n=3,000), ChestX-ray14 (training n=77,870; validation n=8,654; test n=25,596), PadChest (training n = 79,697; validation n=8,783; test n=22,045), CheXpert (training n=115,449; validation n=13,098; test n=29,318), MIMIC-CXR (training n=153,255; validation n=18,139; test n=43,793), and UKA-CXR (training n=137,902; validation n=15,353; test n=40,106).

At 512 × 512 pixels, this pattern changed. For ViT-B/16, DINOv3 outperformed DINOv2 in five of six adult cohorts, including VinDr-CXR (90.3 ± 0.5 vs. 89.1 ± 0.5; p = 0.019), ChestX-ray14 (81.3 ± 0.2 vs. 80.0 ± 0.2; p = 0.006), CheXpert (81.9 ± 0.2 vs. 81.4 ± 0.2; p = 0.006), MIMIC-CXR (82.5 ± 0.1 vs. 80.7 ± 0.2; p = 0.006), and UKA-CXR (88.5 ± 0.1 vs. 88.0 ± 0.1; p = 0.006). The strongest overall results were obtained with ConvNeXt-B, where DINOv3 at 512 × 512 gave the highest mean AUROC in all six adult datasets, reaching 90.5 ± 0.5 in VinDr-CXR, 82.3 ± 0.2 in ChestX-ray14, 89.3 ± 0.2 in PadChest, 82.5 ± 0.1 in CheXpert, 82.7 ± 0.2 in MIMIC-CXR, and 88.5 ± 0.1 in UKA-CXR. Relative to ConvNeXt-B with ImageNet initialization, these gains were significant in all six adult cohorts (p ≤ 0.013).

The key result was therefore not simply that higher resolution improved accuracy, but that it changed which initialization was preferable. Across adult datasets, the most consistent 224-to-512 gains were concentrated in DINOv3-initialized models, particularly with ConvNeXt-B (**Figure 2c–f**). This resolution dependence was also reflected at the finding-group level: the largest average gains were observed for small focal lesions and boundary-dependent findings, whereas large-structure findings changed little between 224 × 224 and 512 × 512 pixels (**Supplementary Table 2, Supplementary Figure 8**). Representative example radiographs for findings that showed the largest apparent resolution sensitivity are provided in **Supplementary Figure 8**. In contrast, the pediatric cohort did not follow this pattern. Pedi-CXR showed minimal separation between configurations at either resolution, with no significant advantage for DINOv3 over ImageNet or DINOv2 at 224 × 224 or 512 × 512 (all p ≥ 0.31). Because this behavior differed qualitatively from the adult cohorts, we analyzed the pediatric dataset separately in the next section.

**Table 2:** Core benchmark performance across datasets, initialization strategies, backbone families, and input resolutions. Average area under the receiver operating characteristic curve (AUROC) derived from 1,000 bootstrap resamples for full fine-tuning of ViT-B/16 and ConvNeXt-B backbones across the universal input resolutions of 224 × 224 and 512 × 512 pixels for all seven datasets: Pedi-CXR (training n=7,728; test n=1,397), VinDr-CXR (training n=15,000; test n=3,000), ChestX-ray14 (training n=77,870; validation n=8,654; test n=25,596), PadChest (training n = 79,697; validation n=8,783; test n=22,045), CheXpert (training n=115,449; validation n=13,098; test n=29,318), MIMIC-CXR (training n=153,255; validation n=18,139; test n=43,793), and UKA-CXR (training n=137,902; validation n=15,353; test n=40,106). Results are reported for models initialized from ImageNet, DINOv2, and DINOv3. For each evaluated configuration, results should be reported as mean ± standard deviation with 95% confidence intervals (CIs).

| Configuration | Pedi-CXR | VinDr-CXR | ChestX-ray14 | PadChest | CheXpert | MIMIC-CXR | UKA-CXR |
|---|---|---|---|---|---|---|---|
| ViT-B ImageNet 224 | 73.2 ± 1.2<br>[70.9, 75.5] | 88.3 ± 0.6<br>[87.0, 89.5] | 79.0 ± 0.2<br>[78.6, 79.4] | 87.0 ± 0.2<br>[86.7, 87.4] | 79.7 ± 0.2<br>[79.4, 80.0] | 79.9 ± 0.2<br>[79.6, 80.2] | 87.7 ± 0.1<br>[87.5, 87.9] |
| ViT-B ImageNet 224 P-value | 0.31 | 0.006 | 0.006 | 0.012 | 0.006 | 0.006 | 0.006 |
| ViT-B DINOv2 224 | 73.4 ± 1.2<br>[71.1, 75.6] | 89.2 ± 0.7<br>[87.5, 90.5] | 80.1 ± 0.2<br>[79.6, 80.5] | 88.0 ± 0.2<br>[87.6, 88.4] | 80.3 ± 0.2<br>[80.0, 80.6] | 80.9 ± 0.2<br>[80.6, 81.2] | 88.1 ± 0.1<br>[87.9, 88.3] |
| ViT-B DINOv2 224 P-value | 0.61 | 0.11 | 0.006 | 0.006 | 0.006 | 0.006 | 0.006 |
| ViT-B DINOv3 224 | 72.7 ± 1.2<br>[70.2, 75.1] | 90.2 ± 0.6<br>[89.0, 91.2] | 80.1 ± 0.2<br>[79.7, 80.6] | 87.4 ± 0.2<br>[87.0, 87.8] | 80.0 ± 0.2<br>[79.7, 80.3] | 80.8 ± 0.1<br>[80.5, 81.0] | 88.0 ± 0.1<br>[87.8, 88.2] |
| ConvNeXt-B ImageNet 224 | 72.6 ± 1.2<br>[70.2, 75.0] | 88.0 ± 0.6<br>[86.8, 89.1] | 79.8 ± 0.2<br>[79.4, 80.2] | 87.5 ± 0.2<br>[87.0, 87.8] | 79.5 ± 0.2<br>[79.2, 79.8] | 80.4 ± 0.1<br>[80.1, 80.7] | 87.8 ± 0.1<br>[87.6, 88.0] |
| ConvNeXt-B ImageNet 224 P-value | 0.08 | 0.45 | 0.001 | 0.001 | 0.001 | 0.001 | 0.001 |
| ConvNeXt-B DINOv3 224 | 73.8 ± 1.1<br>[71.6, 76.0] | 88.0 ± 0.6<br>[87.0, 89.1] | 80.4 ± 0.2<br>[80.0, 80.8] | 88.0 ± 0.2<br>[87.6, 88.3] | 80.5 ± 0.2<br>[80.2, 80.8] | 81.2 ± 0.1<br>[80.9, 81.5] | 88.2 ± 0.1<br>[88.0, 88.4] |
| ViT-B ImageNet 512 | 73.7 ± 1.2<br>[71.1, 76.0] | 86.4 ± 0.7<br>[85.1, 87.6] | 79.5 ± 0.2<br>[79.1, 80.0] | 87.6 ± 0.2<br>[87.2, 88.0] | 80.7 ± 0.2<br>[80.3, 81.0] | 81.4 ± 0.1<br>[81.1, 81.7] | 88.0 ± 0.1<br>[87.8, 88.2] |
| ViT-B ImageNet 512 P-value | 0.37 | 0.006 | 0.006 | 0.006 | 0.006 | 0.006 | 0.006 |
| ViT-B DINOv2 512 | 74.1 ± 1.2<br>[72.0, 76.4] | 89.1 ± 0.5<br>[88.0, 90.1] | 80.0 ± 0.2<br>[79.6, 80.4] | 88.5 ± 0.2<br>[88.1, 88.8] | 81.4 ± 0.2<br>[81.1, 81.7] | 81.2 ± 0.1<br>[80.9, 81.5] | 88.0 ± 0.1<br>[87.8, 88.2] |
| ViT-B DINOv2 512 P-value | 0.32 | 0.006 | 0.006 | 0.006 | 0.006 | 0.044 | 0.26 |
| ViT-B DINOv3 512 | 73.9 ± 1.2<br>[71.6, 76.3] | 90.3 ± 0.5<br>[89.3, 91.2] | 81.3 ± 0.2<br>[80.9, 81.8] | 88.9 ± 0.2<br>[88.6, 89.3] | 81.9 ± 0.2<br>[81.6, 82.2] | 82.5 ± 0.1<br>[82.2, 82.8] | 88.5 ± 0.1<br>[88.2, 88.7] |
| ConvNeXt-B ImageNet 512 | 74.0 ± 1.1<br>[71.8, 76.2] | 89.7 ± 0.5<br>[88.6, 90.7] | 81.5 ± 0.2<br>[81.1, 81.9] | 88.5 ± 0.2<br>[88.0, 88.8] | 81.6 ± 0.2<br>[81.2, 81.9] | 82.2 ± 0.1<br>[81.9, 82.4] | 88.4 ± 0.1<br>[88.2, 88.6] |
| ConvNeXt-B ImageNet 512 P-value | 0.66 | 0.013 | 0.001 | 0.001 | 0.001 | 0.001 | 0.001 |
| ConvNeXt-B DINOv3 512 | 73.6 ± 1.2<br>[71.2, 75.9] | 90.5 ± 0.5<br>[89.5, 91.5] | 82.3 ± 0.2<br>[81.8, 82.7] | 89.3 ± 0.2<br>[89.0, 89.7] | 82.5 ± 0.1<br>[82.2, 82.8] | 82.7 ± 0.2<br>[82.3, 83.0] | 88.5 ± 0.1<br>[88.3, 88.7] |
| Best-performing configuration | ConvNeXt-B ImageNet 512 | ConvNeXt-B DINOv3 512 | ConvNeXt-B DINOv3 512 | ConvNeXt-B DINOv3 512 | ConvNeXt-B DINOv3 512 | ConvNeXt-B DINOv3 512 | ConvNeXt-B DINOv3 512 |

## The pediatric cohort shows a distinct transfer pattern that does not mirror adult resolution scaling

Pedi-CXR was not simply the weakest-performing dataset, but the only cohort that failed to follow the adult resolution-scaling pattern (**Table 2**). Whereas adult configurations showed a consistent positive lift when moving from 224 × 224 to 512 × 512 pixels, pediatric configurations clustered around zero, with a median gain of only +0.74 percentage points compared with +1.05 percentage points across adult cohorts (**Figure 3a**). This difference was also evident when plotting performance at 224 × 224 vs. 512 × 512 pixels directly: adult configurations lay predominantly above the identity line, whereas the pediatric points remained close to it (**Figure 3b**).

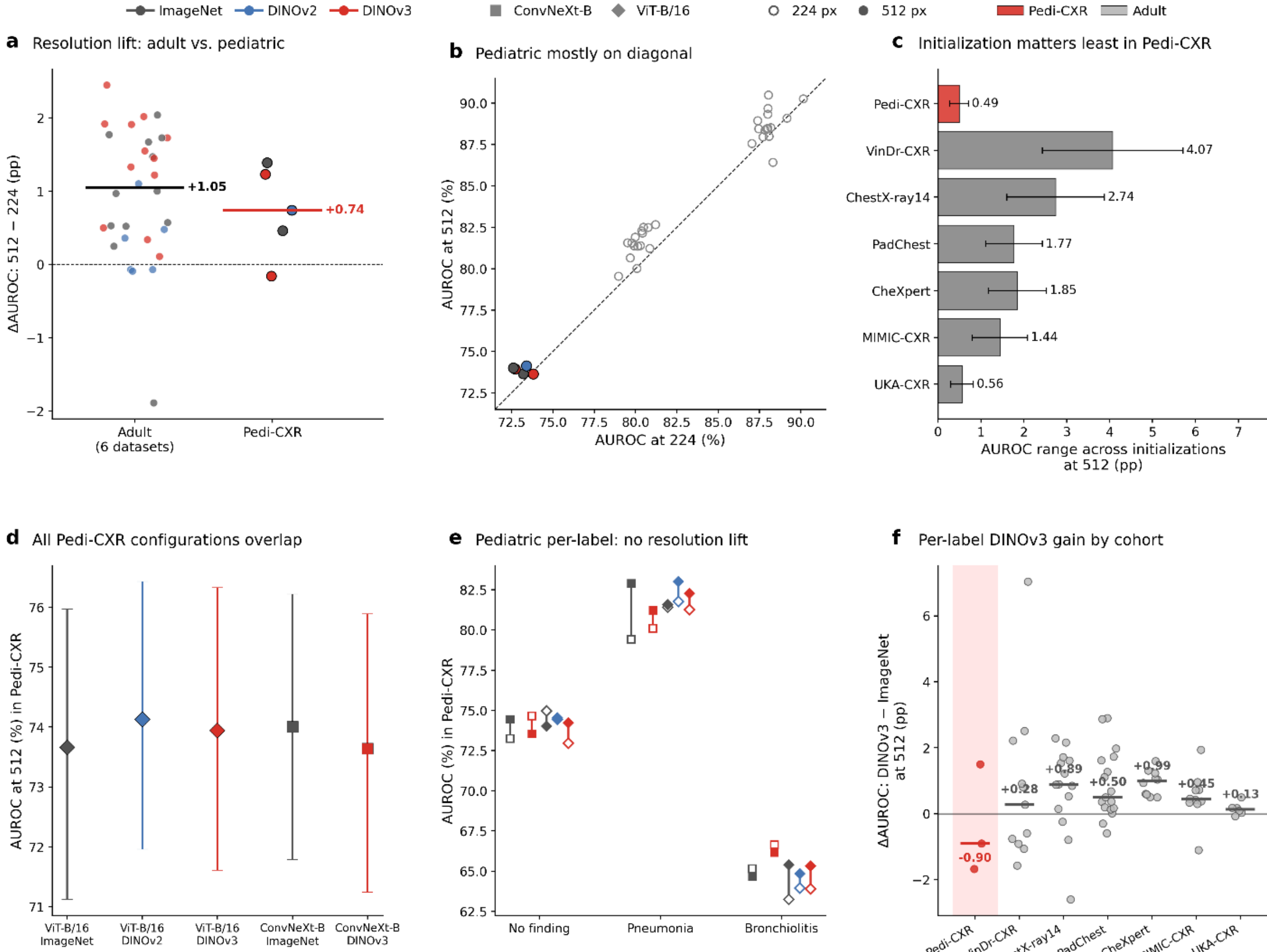

**Figure 3**: The pediatric cohort shows a distinct transfer pattern that does not follow the adult resolution-scaling behavior. Results are shown for all seven datasets: Pedi-CXR (training n=7,728; test n=1,397), VinDr-CXR (training n=15,000; test n=3,000), ChestX-ray14 (training n=77,870; validation n=8,654; test n=25,596), PadChest (training n = 79,697; validation n=8,783; test n=22,045), CheXpert (training n=115,449; validation n=13,098; test n=29,318), MIMIC-CXR (training n=153,255; validation n=18,139; test n=43,793), and UKA-CXR (training n=137,902; validation n=15,353; test n=40,106). **a** Distribution of AUROC change when increasing resolution from 224 to 512 pixels for every backbone-initialization combination, separated by adult cohorts (6 datasets, 30 configurations) and Pedi-CXR (5 configurations). Each point represents one configuration within one dataset. Horizontal bars mark the median per group, annotated in bold. Adult configurations show a consistently positive resolution lift; Pedi-CXR configurations cluster around zero. **b** AUROC at 224 pixels against AUROC at 512 pixels for every configuration across all seven datasets. The dashed line denotes equality. Adult configurations (grey) sit above the line; pediatric configurations (colored by initialization) sit on it. **c** Range of mean AUROC across the five initialization-backbone configurations at 512 pixels for each dataset, with standard deviation error bars. Pedi-CXR exhibits the narrowest range (0.49 pp), indicating that initialization choice has minimal influence on pediatric performance. **d** Mean AUROC at 512 pixels with 95% bootstrap confidence intervals for all five Pedi-CXR configurations. All intervals overlap, confirming that no initialization or backbone offers a statistically distinguishable advantage. **e** Per-label dumbbell plot of the 224-to-512 AUROC shift for each of the three pediatric labels (Any finding, Pneumonia, Bronchiolitis). Open markers denote 224-pixel performance; filled markers denote 512-pixel performance. Colors denote initialization; marker shapes denote backbone (squares for ConvNeXt-B, diamonds for ViT-B/16). No label shows a consistent resolution-driven gain. **f** Distribution of per-label AUROC differences (DINOv3 minus ImageNet at 512, ConvNeXt-B) across all seven datasets. Each point represents one pathology label; horizontal bars mark the median. Pedi-CXR is the only cohort with a negative median, confirming that the DINOv3 advantage seen at the label level in adult cohorts does not transfer to the pediatric cohort.

At the dataset level, performance differences between initialization and backbone choices were minimal in Pedi-CXR. At 512 × 512 pixels, all five evaluated configurations fell within a narrow AUROC range of 0.49 percentage points, far smaller than in any adult cohort (**Figure 3c**). Their uncertainty intervals also overlapped broadly (**Figure 3d**), and none of the pairwise comparisons between DINOv3 and ImageNet or DINOv2 reached significance at either 224 × 224 or 512 × 512 pixels (all $p \geq 0.31$; **Table 2, Supplementary Table 1**). Even the nominally highest pediatric configuration, ViT-B/16 with DINOv2 at 512 × 512, reached only 74.1 ± 1.2, while ConvNeXt-B with DINOv3 reached 73.6 ± 1.2.

This absence of a pediatric resolution effect also held at the label level. None of the three Pedi-CXR labels showed a consistent upward shift from 224 to 512 pixels across configurations (**Figure 3e**). Moreover, when comparing DINOv3 with ImageNet at 512 × 512 using ConvNeXt-B, Pedi-CXR was the only cohort with a negative median label-level effect, whereas all adult datasets showed positive medians (**Figure 3f**). Thus, the adult result that DINOv3 benefits from higher resolution did not transfer to the pediatric setting. These findings indicate that the adult DINOv3-at-512 pattern should not be generalized automatically to pediatric chest radiography. In this cohort, increasing resolution and changing initialization or backbone had little measurable impact, suggesting that the determinants of transfer performance differ materially between pediatric and adult chest X-ray analysis.

## Scaling beyond 512 × 512 yields limited benefit despite sharply increased computational cost

We next examined whether the adult 512 × 512 advantage extended to 1024 × 1024 training in the three cohorts evaluated at this resolution, namely Pedi-CXR, ChestX-ray14, and MIMIC-CXR (**Figure 2g-i, Table 3, Supplementary Table 3**). Formal 512 vs. 1024 comparison showed that additional scaling rarely improved performance. Across the 15 evaluated dataset-configuration pairs, only two yielded significant gains, both in MIMIC-CXR with ConvNeXt-B: ImageNet increased from 82.2 ± 0.1 to 83.2 ± 0.1 (ΔAUROC = +1.0 ± 0.2) and DINOv3 from 82.7 ± 0.2 to 83.3 ± 0.1 (ΔAUROC = +0.7 ± 0.2). All remaining comparisons were neutral or negative, including clear degradations for several ViT-B/16 settings, such as Pedi-CXR with ImageNet (73.7 ± 1.2 to 65.8 ± 1.4; ΔAUROC = -7.9 ± 1.8) and ChestX-ray14 with DINOv2 (80.0 ± 0.2 to 73.4 ± 0.2; ΔAUROC = -6.6 ± 0.3).

The overall pattern was therefore not one of continued monotonic improvement with increasing resolution. In Pedi-CXR, 1024 × 1024 provided no measurable benefit for any configuration. In ChestX-ray14, performance changes were small for ConvNeXt-B and negative for all ViT-B/16 initializations. Only MIMIC-CXR showed a modest positive shift, and even there the gains were confined to ConvNeXt-B and remained well below one AUROC point for DINOv3 (**Figure 2g, Table 3**). Thus, while DINOv3 retained its strong relative standing at 1024 × 1024 wherever tested against alternative initializations at the same resolution, scaling beyond 512 × 512 did not produce a consistent cross-dataset accuracy benefit.

**Table 3:** Targeted high-resolution comparison and computational efficiency. Focused 512 × 512 vs. 1024 × 1024 comparison for the three representative datasets evaluated at 1024 × 1024 resolution, namely Pedi-CXR (training n=7,728; test n=1,397), ChestX-ray14 (training n=77,870; validation n=8,654; test n=25,596), and MIMIC-CXR (training n=153,255; validation n=18,139; test n=43,793). For each model configuration, the table reports AUROC at 512 × 512 and 1024 × 1024 as mean ± SD [95% CI]. For non-DINOv3 initialization strategies, the corresponding adjusted pairwise bootstrap p-value vs. DINOv3 at the same dataset and resolution is appended to the absolute AUROC estimate. The table further reports the absolute cross-resolution change in AUROC (ΔAUROC = AUROC-1024 – AUROC-512) as mean ± SD [95% CI], and the relative total training-time ratio (1024 / 512). Pairwise bootstrap p-values were computed as two-sided p-values from paired bootstrap tests using 1,000 resampled AUROC pairs per model, with identical resampling across initialization strategies to ensure fair comparison. The total training-time ratio is reported once per backbone-initialization configuration as mean ± SD across the three cohorts evaluated at both resolutions, rather than separately for each dataset.

| Configuration | Pedi-CXR | ChestX-ray14 | MIMIC-CXR |
|---|---|---|---|
| ViT-B ImageNet: AUROC at 512 × 512 | 73.7 ± 1.2 [71.1, 76.0] (p = 0.37) | 79.5 ± 0.2 [79.1, 80.0] (p = 0.006) | 81.4 ± 0.1 [81.1, 81.7] (p = 0.006) |
| ViT-B ImageNet: AUROC at 1024 × 1024 | 65.8 ± 1.4 [63.0, 68.4] (p = 0.001) | 78.8 ± 0.2 [78.3, 79.2] (p = 0.002) | 81.1 ± 0.2 [80.8, 81.4] (p = 0.002) |
| ViT-B ImageNet: ΔAUROC (1024 - 512) | -7.9 ± 1.8 [-11.5, -4.2] | -0.8 ± 0.3 [-1.4, -0.1] | -0.3 ± 0.2 [-0.7, 0.1] |
| ViT-B ImageNet: total training-time ratio (1024 / 512) | 11.1 ± 5.5 | 11.1 ± 5.5 | 11.1 ± 5.5 |
| ViT-B DINOv2: AUROC at 512 × 512 | 74.1 ± 1.2 [72.0, 76.4] (p = 0.54) | 80.0 ± 0.2 [79.6, 80.4] (p = 0.006) | 81.2 ± 0.1 [80.9, 81.5] (p = 0.006) |
| ViT-B DINOv2: AUROC at 1024 × 1024 | 70.8 ± 1.2 [68.3, 73.2] (p = 0.001) | 73.4 ± 0.2 [72.9, 73.8] (p = 0.002) | 81.1 ± 0.1 [80.8, 81.4] (p = 0.002) |
| ViT-B DINOv2: ΔAUROC (1024 - 512) | -3.4 ± 1.7 [-6.7, -0.0] | -6.6 ± 0.3 [-7.2, -6.0] | -0.1 ± 0.2 [-0.5, 0.3] |
| ViT-B DINOv2: total training-time ratio (1024 / 512) | 13.9 ± 7.9 | 13.9 ± 7.9 | 13.9 ± 7.9 |
| ViT-B DINOv3: AUROC at 512 × 512 | 73.9 ± 1.2 [71.6, 76.3] | 81.3 ± 0.2 [80.9, 81.8] | 82.5 ± 0.1 [82.2, 82.8] |
| ViT-B DINOv3: AUROC at 1024 × 1024 | 73.7 ± 1.2 [71.4, 75.9] | 81.0 ± 0.2 [80.5, 81.4] | 82.7 ± 0.1 [82.4, 83.0] |
| ViT-B DINOv3: ΔAUROC (1024 - 512) | -0.3 ± 1.7 [-3.5, 3.0] | -0.4 ± 0.3 [-1.0, 0.3] | 0.2 ± 0.2 [-0.2, 0.6] |
| ViT-B DINOv3: total training-time ratio (1024 / 512) | 6.7 ± 1.9 | 6.7 ± 1.9 | 6.7 ± 1.9 |
| ConvNeXt-B ImageNet: AUROC at 512 × 512 | 74.0 ± 1.1 [71.8, 76.2] (p = 0.66) | 81.5 ± 0.2 [81.1, 81.9] (p = 0.001) | 82.2 ± 0.1 [81.9, 82.4] (p = 0.001) |
| ConvNeXt-B ImageNet: AUROC at 1024 × 1024 | 74.3 ± 1.2 [72.1, 76.7] (p = 0.76) | 82.0 ± 0.2 [81.6, 82.4] (p = 0.002) | 83.2 ± 0.1 [82.9, 83.5] (p = 0.038) |
| ConvNeXt-B ImageNet: ΔAUROC (1024 - 512) | 0.3 ± 1.6 [-2.9, 3.5] | 0.5 ± 0.3 [-0.1, 1.0] | 1.0 ± 0.2 [0.6, 1.4] |
| ConvNeXt-B ImageNet: total training-time ratio (1024 / 512) | 3.1 ± 1.1 | 3.1 ± 1.1 | 3.1 ± 1.1 |
| ConvNeXt-B DINOv3: AUROC at 512 × 512 | 73.6 ± 1.2 [71.2, 75.9] | 82.3 ± 0.2 [81.8, 82.7] | 82.7 ± 0.2 [82.3, 83.0] |
| ConvNeXt-B DINOv3: AUROC at 1024 × 1024 | 73.8 ± 1.2 [71.5, 76.1] | 82.2 ± 0.2 [81.7, 82.6] | 83.3 ± 0.1 [83.1, 83.6] |
| ConvNeXt-B DINOv3: ΔAUROC (1024 - 512) | 0.2 ± 1.7 [-3.1, 3.5] | -0.1 ± 0.3 [-0.8, 0.5] | 0.7 ± 0.2 [0.3, 1.1] |
| ConvNeXt-B DINOv3: total training-time ratio (1024 / 512) | 3.5 ± 1.2 | 3.5 ± 1.2 | 3.5 ± 1.2 |

These limited returns contrasted sharply with the computational cost of 1024-pixel training. Across all full fine-tuning experiments, moving from 512 × 512 to 1024 × 1024 increased per-epoch training time by 6.2 ± 4.4-fold overall, with a larger penalty for ViT-B/16 than for ConvNeXt-B (7.9 ± 5.1-fold vs. 3.6 ± 0.9-fold; **Supplementary Table 3**). The effect on end-to-end training time was even more pronounced: the median total wall-clock ratio for 1024 × 1024 vs. 512 × 512 was 4.8 overall, but 8.9 for ViT-B/16 and 2.9 for ConvNeXt-B. In the targeted 1024 × 1024 experiments, this translated to mean total training-time ratios of 11.1 ± 5.5 for ViT-B/ImageNet, 13.9 ± 7.9 for ViT-B/DINOv2, 6.7 ± 1.9 for ViT-B/DINOv3, 3.1 ± 1.1 for ConvNeXt-B/ImageNet, and 3.5 ± 1.2 for ConvNeXt-B/DINOv3 (**Table 3, Figure 2h,i**). Under the present training setup, 1024 × 1024 training rarely improves classification performance enough to justify its substantially higher cost. The practical operating point of the benchmark therefore remained at 512 × 512 pixels, where DINOv3 already achieved its most reproducible gains across adult cohorts without incurring the marked efficiency penalty of further resolution scaling.

## ConvNeXt-B remains superior to ViT-B/16 under both full and parameter-efficient adaptation

Under full fine-tuning, ConvNeXt-B consistently outperformed ViT-B/16 across adult datasets and resolutions (**Table 2**), with the gap most apparent at 512 × 512 pixels. For example, with DINOv3 initialization at 512 × 512, ConvNeXt-B reached 82.7 ± 0.2 vs. 82.5 ± 0.1 in MIMIC-CXR, 82.3 ± 0.2 vs. 81.3 ± 0.2 in ChestX-ray14, and 82.5 ± 0.1 vs. 81.9 ± 0.2 in CheXpert. Thus, the main benchmark already indicated a stable backbone advantage under standard end-to-end optimization. We then asked whether this result could be explained simply by the use of full fine-tuning by replacing it with a parameter-efficient adaptation regime, implemented here with low-rank adaptation (LoRA)[36] (**Figure 4a-c, Supplementary Table 4**). Across all evaluated dataset-configuration pairs, LoRA reduced AUROC relative to full fine-tuning. The magnitude of this loss ranged from 1.3 ± 0.3 to 5.1 ± 0.3 percentage points for ViT-B/16 and from 2.1 ± 0.2 to 4.1 ± 0.3 percentage points for ConvNeXt-B. For DINOv3 specifically, the drop under LoRA was larger for ViT-B/16 than for ConvNeXt-B in all three datasets: 5.1 ± 0.3 vs. 2.8 ± 0.3 in ChestX-ray14, 4.3 ± 0.2 vs. 3.3 ± 0.2 in CheXpert, and 4.1 ± 0.2 vs. 2.1 ± 0.2 in MIMIC-CXR. Thus, reduced adaptation capacity did not preferentially benefit ViT-B/16.

Accordingly, the backbone gap did not narrow under parameter-efficient adaptation. Instead, ConvNeXt-B remained superior to ViT-B/16 in five of six backbone-initialization pairs, and the gap widened in several cases (**Figure 4c**). For DINOv3, the ConvNeXt-B minus ViT-B/16 difference increased from 1.0 to 3.2 percentage points in ChestX-ray14 and from 0.2 to 2.2 percentage points in MIMIC-CXR, while remaining positive in CheXpert (0.6 to 1.6 percentage points). The same qualitative pattern held for ImageNet initialization. These findings indicate that the ConvNeXt advantage cannot be attributed merely to having used full end-to-end fine-tuning in the main benchmark.

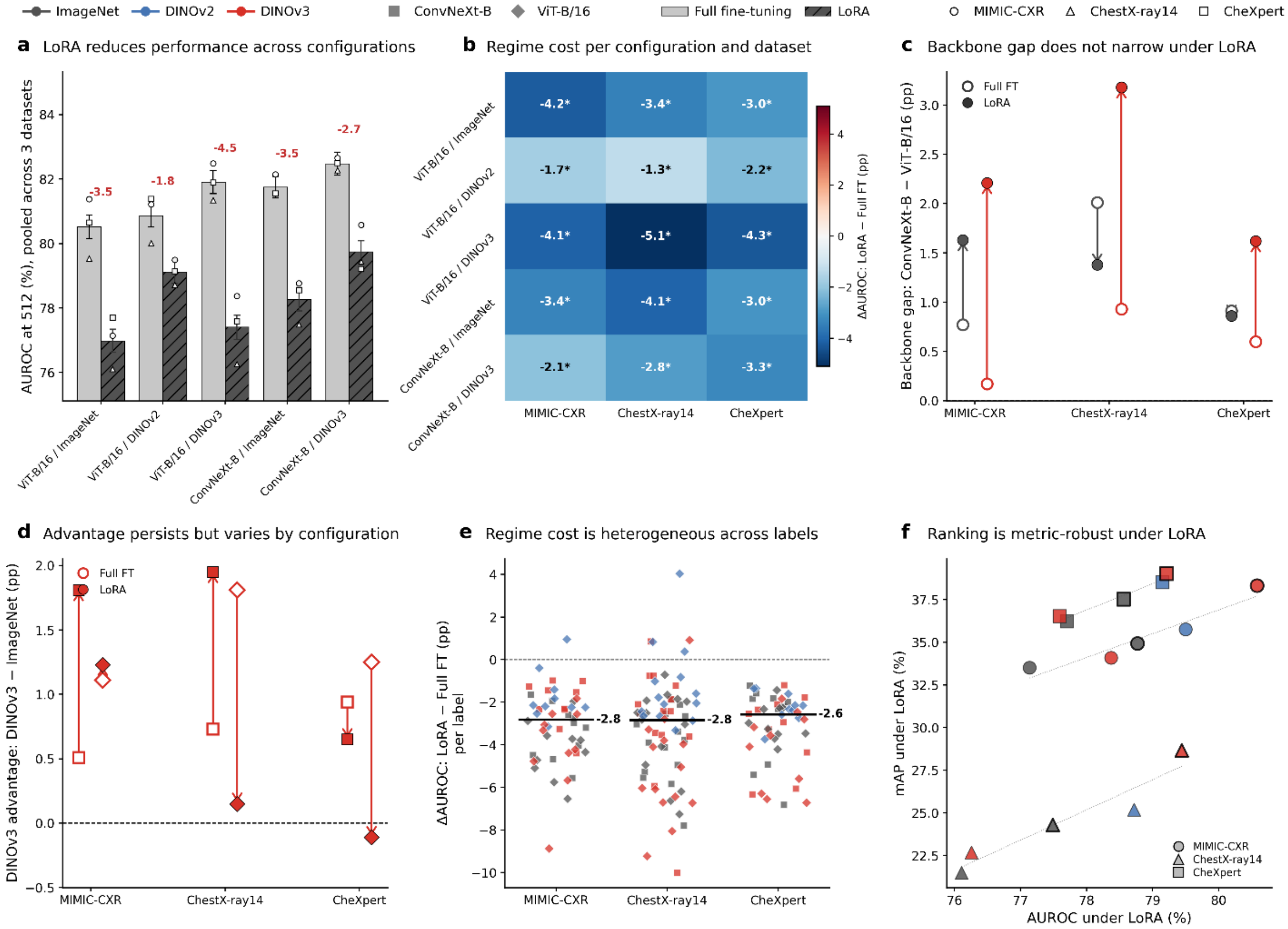

**Figure 4**: Parameter-efficient fine-tuning (LoRA) does not narrow the ConvNeXt-B over ViT-B/16 gap; the conclusion is robust to class-imbalance-aware metrics. All panels use the three adult datasets where LoRA was evaluated: MIMIC-CXR (training n=153,255; validation n=18,139; test n=43,793), ChestX-ray14 (training n=77,870; validation n=8,654; test n=25,596), CheXpert (training n=115,449; validation n=13,098; test n=29,318) at 512 pixels. **a** Mean AUROC across the three datasets for each of the five backbone-initialization configurations, under full fine-tuning (light bars) and LoRA (dark hatched bars). Individual dataset values are overlaid as markers (circles, triangles, squares for MIMIC-CXR, ChestX-ray14, CheXpert). Error bars denote averaged 95% bootstrap confidence intervals. Red annotations show the AUROC change from full fine-tuning to LoRA. **b** Heatmap of the per-configuration, per-dataset AUROC change when switching from full fine-tuning to LoRA. Asterisks denote non-overlapping 95% confidence intervals between regimes. **c** Backbone gap (ConvNeXt-B minus ViT-B/16 AUROC) under full fine-tuning (open circles) and LoRA (filled circles), for each dataset and initialization (excluding DINOv2, which is not evaluated on ConvNeXt-B). Arrows show the direction of change. The gap widens or persists in five of six pairs under LoRA. **d** DINOv3 advantage (DINOv3 minus ImageNet AUROC) under each regime, for each dataset and backbone. The advantage is preserved or grows in four of six pairs under LoRA and collapses for ViT-B/16 on ChestX-ray14 and CheXpert. **e** Distribution of per-label AUROC changes (LoRA minus full fine-tuning) across all available configurations, per dataset. Each point represents one pathology label under one configuration; colours denote initialization and shapes denote backbone. Horizontal bars mark the median. Regime cost is concentrated between zero and −4 pp but with substantial label-level heterogeneity. **f** Mean average precision (mAP) against AUROC under LoRA for every configuration-dataset combination. Marker shapes denote datasets (circles for MIMIC-CXR, triangles for ChestX-ray14, squares for CheXpert).

The initialization story also remained broadly stable. Under parameter-efficient adaptation, DINOv3 retained an advantage over ImageNet for ConvNeXt-B in all three datasets and for ViT-B/16 in MIMIC-CXR, but this advantage weakened or disappeared for ViT-B/16 in ChestX-ray14 and CheXpert (**Figure 4d**). This suggests that ViT-B/16 was more sensitive to the restricted adaptation regime, whereas ConvNeXt-B preserved both its stronger absolute performance and its relative benefit from DINOv3 initialization.

At the label level, parameter-efficient adaptation produced heterogeneous but predominantly negative changes, with median per-label AUROC shifts of about -2.8 percentage points in MIMIC-CXR, -2.8 percentage points in ChestX-ray14, and -2.6 percentage points in CheXpert (**Figure 4e**). Importantly, these shifts did not alter the overall ranking of configurations. Class-imbalance-aware evaluation supported the same conclusion: under LoRA, AUROC and mAP remained strongly aligned across configurations (Spearman $\rho = 1.00$ for MIMIC-CXR, $\rho = 1.00$ for ChestX-ray14, and $\rho = 0.90$ for CheXpert; all $p \leq 0.037$), indicating that the backbone comparison was not an artifact of AUROC alone (**Figure 4f**). The superiority of ConvNeXt-B over ViT-B/16 cannot be explained simply by the use of full fine-tuning. Instead, it persisted under both full and parameter-efficient adaptation, supporting a genuine backbone-level difference in transfer behavior for chest radiograph classification.

## DINOv3's clean-label advantage diminishes under synthetic label corruption

To test whether DINOv3's gains on adult datasets could be explained by superior robustness to weak supervision, we performed a controlled label-noise experiment on VinDr-CXR using ViT-B/16 at 512 × 512 pixels (**Figure 5, Supplementary Table 5**). Under clean labels, DINOv3 gave the strongest performance, reaching 90.3 ± 0.5 AUROC, compared with 89.0 ± 0.6 for DINOv2 and 86.4 ± 0.6 for ImageNet. The same ordering was seen for class-imbalance-aware metrics, with DINOv3 also leading in PR-AUC and mAP (47.7 ± 1.4 and 48.3 ± 1.4, respectively).

This advantage did not persist as label corruption increased. DINOv3 degraded more steeply than the other initializations, falling to 84.7 ± 0.8 at 10% noise, 84.6 ± 0.8 at 20%, 78.9 ± 1.0 at 30%, and 77.9 ± 0.9 at 40% noise (**Figure 5a**). By comparison, DINOv2 retained higher AUROC at 20%, 30%, and 40% noise, and ImageNet surpassed DINOv3 at 40% noise (80.7 ± 0.9 vs. 77.9 ± 0.9). The same crossover was visible for mAP, where DINOv3 decreased from 48.3 ± 1.4 under clean labels to 29.0 ± 0.9 at 40% noise, below both DINOv2 (37.0 ± 1.1) and ImageNet (33.4 ± 1.0) (**Figure 5b,f**).

Viewed relative to the clean-label baseline, DINOv3 also showed the largest overall loss. Its AUROC dropped by 5.6 points at 10% noise and by 12.4 points at 40% noise, compared with drops of 8.7 points for DINOv2 and 5.6 points for ImageNet at the highest corruption level (**Figure 5c**). Retained-performance analysis showed the same pattern: at 40% noise, DINOv3 preserved 86.3 ± 1.0 of its clean-label AUROC, compared with 90.2 ± 1.1 for DINOv2 and 93.5 ± 1.1 for ImageNet (**Supplementary Table 5**). Thus, although DINOv3 started from the highest clean-label baseline, it was not the most stable initialization under progressive label corruption.

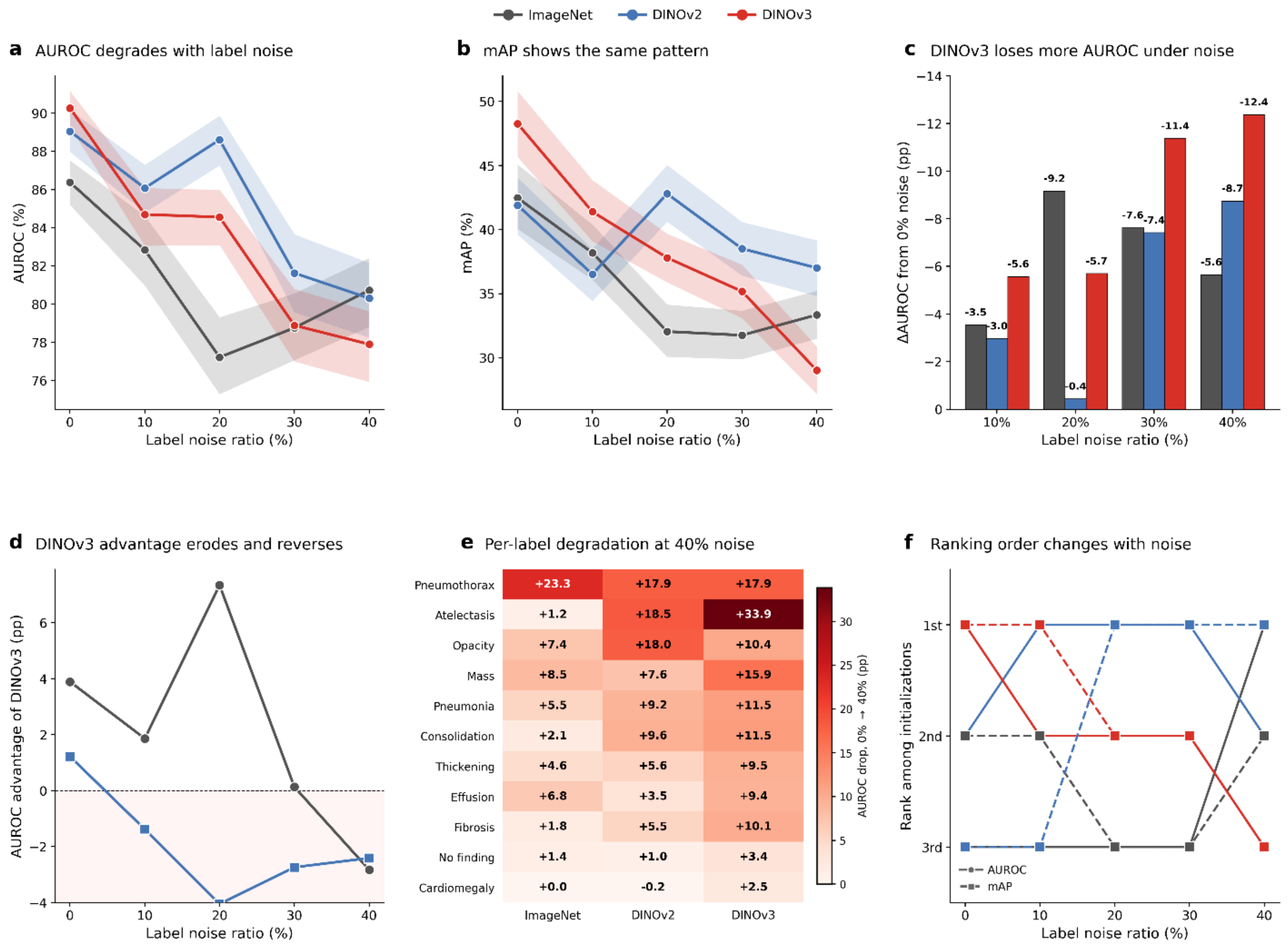

**Figure 5**: DINOv3's advantage over earlier initializations diminishes and reverses under label noise. All panels use VinDr-CXR (full training set contains n=15,000; test set is fixed at n=3,000 for all experiments) (ViT-B/16, 512 × 512) with synthetic symmetric label noise introduced on the training set at five ratios (0, 10, 20, 30, 40%). **a** Macro-averaged AUROC as a function of label noise ratio for each initialization, with 95% bootstrap confidence intervals shown as shaded bands. DINOv3 achieves the highest clean-label AUROC (90.27%) but degrades most steeply, falling below ImageNet at 40% noise. **b** Macro-averaged mAP as a function of noise ratio. The same crossover pattern holds under a class-imbalance-aware metric. **c** AUROC loss from the clean-label baseline at each noise level, per initialization. DINOv3 incurs the largest losses at three of the four noise levels and the largest loss overall (−12.4 pp at 40% noise). **d** DINOv3's AUROC advantage over ImageNet and over DINOv2 as a function of noise. The advantage over ImageNet starts at +3.9 pp, peaks at +7.3 pp at 20% noise, and reverses to −2.8 pp by 40%. The red-shaded region marks the range where DINOv3 has lost its advantage. **e** Per-label AUROC drop from 0% to 40% label noise, ordered by mean drop across initializations. Atelectasis shows an especially asymmetric pattern: ImageNet loses only 1.2 pp while DINOv3 loses 33.9 pp. **f** Configuration ranking at each noise level under AUROC (solid lines) and mAP (dashed lines). DINOv3 is the best-ranked initialization on clean labels under both metrics but drops to third at 40% noise under AUROC and to third under mAP.

This instability was not uniform across findings. At 40% noise, several labels showed substantially larger degradation for DINOv3 than for the other initializations, most notably atelectasis, where the AUROC drop was 33.9 points for DINOv3 compared with 18.5 for DINOv2 and 1.2 for ImageNet (**Figure 5e**). More broadly, DINOv3's advantage over ImageNet and DINOv2 narrowed with increasing corruption and eventually reversed, so that by 40% noise it ranked last under both AUROC and mAP (**Figure 5d,f**). These results establish the label-noise experiment as a boundary-condition analysis rather than a robustness result. DINOv3 was strongest when labels were clean, but its advantage diminished and could reverse under synthetic corruption. Accordingly, the gains observed on weakly labeled adult datasets in the main benchmark should not be interpreted simply as evidence that DINOv3 is intrinsically more robust to noisy supervision.

## DINOv3 generalizes across institutions under external validation

To test whether the main benchmark findings extended beyond within-dataset splits, we trained models on MIMIC-CXR and evaluated them externally on ChestX-ray14 and CheXpert using a harmonized seven-label setting comprising cardiomegaly, pleural effusion, pneumonia, atelectasis, no finding, consolidation, and pneumothorax (**Figure 6a-d, Table 4**). Across all five evaluated configurations, external performance remained close to the corresponding internal test performance, with AUROC drops of 0.9 to 2.7 percentage points at 512 × 512. The smallest external drop was observed for ConvNeXt-B with DINOv3 on ChestX-ray14, from 79.3 ± 0.2 internally to 78.3 ± 0.2 externally, corresponding to a gap of 1.0 ± 0.3. On CheXpert, the external gaps were somewhat larger but remained modest overall, ranging from 2.3 ± 0.3 to 2.7 ± 0.3.

Importantly, the relative ranking of configurations was largely preserved under domain shift. At 512 × 512, DINOv3 remained the strongest initialization for both backbones on both external target datasets. For ViT-B/16, DINOv3 reached 78.0 ± 0.2 on ChestX-ray14 and 80.6 ± 0.2 on CheXpert, exceeding both ImageNet and DINOv2. The same pattern held for ConvNeXt-B, where DINOv3 achieved the highest external AUROC on ChestX-ray14 (78.3 ± 0.2) and CheXpert (80.7 ± 0.2). Thus, the DINOv3 advantage observed in the internal benchmark was not lost when transferring across institutions.

The resolution effect also persisted externally. Across both target datasets and all configurations, moving from 224 × 224 to 512 × 512 improved external AUROC by 0.8 to 1.5 percentage points (**Figure 6b, Table 4**). For example, ViT-B with DINOv3 improved from 76.5 ± 0.2 to 78.0 ± 0.2 on ChestX-ray14 and from 79.2 ± 0.2 to 80.6 ± 0.2 on CheXpert. ConvNeXt-B with DINOv3 showed a similar pattern, increasing from 76.8 ± 0.2 to 78.3 ± 0.2 on ChestX-ray14 and from 79.4 ± 0.2 to 80.7 ± 0.2 on CheXpert. This indicates that the 512 × 512 advantage was not specific to in-domain evaluation. Class-imbalance-aware metrics supported the same conclusion. External PR-AUC and mAP closely tracked AUROC across configurations, with DINOv3 again ranking highest at 512 × 512 for both backbones and both target datasets. For example, ConvNeXt-B with DINOv3 achieved the best external PR-AUC and mAP on both ChestX-ray14 (35.1 ± 0.3 and 35.2 ± 0.3) and CheXpert (42.4 ± 0.3 and 42.4 ± 0.3), and the external AUROC-mAP relationship remained strongly aligned across institutions (**Figure 6d**). These findings show that the main adult benchmark result, namely the advantage of DINOv3 at 512 × 512, remained visible under external validation rather than being confined to internal splits.

**Table 4:** External validation across target datasets and resolutions. External evaluation results for models trained on MIMIC-CXR (training n=153,255; validation n=18,139; test n=43,793) and tested on ChestX-ray14 (training n=77,870; validation n=8,654; test n=25,596) and CheXpert (training n=115,449; validation n=13,098; test n=29,318) using the overlapping seven-label classification setting comprising cardiomegaly, pleural effusion, pneumonia, atelectasis, no finding, consolidation, and pneumothorax. Results are shown for ViT-B/16 initialized from ImageNet, DINOv2, and DINOv3, and for ConvNeXt-B initialized from ImageNet and DINOv2, each evaluated at 224 × 224 and 512 × 512 resolution. For each configuration, the table reports the internal test AUROC on MIMIC-CXR together with the external AUROC, precision-recall area under the curve (PR-AUC), and mean average precision (mAP) on ChestX-ray14 and CheXpert.

| Resolution | Backbone | Initialization | Internal ChestX-ray14 | External ChestX-ray14 | Internal - External | Internal CheXpert | External CheXpert | Internal - External |
|---|---|---|---|---|---|---|---|---|
| AUROC values | | | | | | | | |
| 224 × 224 | ViT-B | ImageNet | 77.4 ± 0.2 | 75.9 ± 0.2 | 1.5 ± 0.4 | 81.2 ± 0.2 | 78.5 ± 0.2 | 2.7 ± 0.3 |
| 224 × 224 | ViT-B | DINOv2 | 78.0 ± 0.2 | 76.4 ± 0.2 | 1.6 ± 0.3 | 81.8 ± 0.2 | 79.1 ± 0.2 | 2.7 ± 0.3 |
| 224 × 224 | ViT-B | DINOv3 | 77.8 ± 0.2 | 76.5 ± 0.2 | 1.3 ± 0.3 | 81.5 ± 0.2 | 79.2 ± 0.2 | 2.4 ± 0.3 |
| 224 × 224 | ConvNeXt-B | ImageNet | 77.7 ± 0.2 | 76.1 ± 0.2 | 1.6 ± 0.3 | 81.2 ± 0.2 | 78.9 ± 0.2 | 2.3 ± 0.3 |
| 224 × 224 | ConvNeXt-B | DINOv3 | 78.0 ± 0.2 | 76.8 ± 0.2 | 1.2 ± 0.3 | 81.9 ± 0.2 | 79.4 ± 0.2 | 2.5 ± 0.3 |
| 512 × 512 | ViT-B | ImageNet | 77.6 ± 0.2 | 76.7 ± 0.2 | 0.9 ± 0.3 | 82.0 ± 0.2 | 79.7 ± 0.2 | 2.3 ± 0.3 |
| 512 × 512 | ViT-B | DINOv2 | 78.7 ± 0.2 | 77.3 ± 0.2 | 1.4 ± 0.3 | 82.8 ± 0.2 | 80.1 ± 0.2 | 2.7 ± 0.3 |
| 512 × 512 | ViT-B | DINOv3 | 79.2 ± 0.2 | 78.0 ± 0.2 | 1.2 ± 0.3 | 83.0 ± 0.2 | 80.6 ± 0.2 | 2.3 ± 0.3 |
| 512 × 512 | ConvNeXt-B | ImageNet | 78.6 ± 0.2 | 77.0 ± 0.2 | 1.5 ± 0.3 | 82.5 ± 0.2 | 79.8 ± 0.2 | 2.7 ± 0.3 |
| 512 × 512 | ConvNeXt-B | DINOv3 | 79.3 ± 0.2 | 78.3 ± 0.2 | 1.0 ± 0.3 | 83.3 ± 0.2 | 80.7 ± 0.2 | 2.6 ± 0.3 |
| PR-AUC values | | | | | | | | |
| 224 × 224 | ViT-B | ImageNet | 33.3 ± 0.3 | 31.3 ± 0.3 | 2.0 ± 0.5 | 42.3 ± 0.3 | 38.2 ± 0.3 | 4.1 ± 0.4 |
| 224 × 224 | ViT-B | DINOv2 | 34.2 ± 0.3 | 32.5 ± 0.3 | 1.7 ± 0.5 | 43.5 ± 0.3 | 39.3 ± 0.3 | 4.2 ± 0.4 |
| 224 × 224 | ViT-B | DINOv3 | 34.2 ± 0.3 | 32.3 ± 0.3 | 1.9 ± 0.5 | 42.5 ± 0.3 | 38.9 ± 0.3 | 3.6 ± 0.4 |
| 224 × 224 | ConvNeXt-B | ImageNet | 33.6 ± 0.3 | 31.3 ± 0.3 | 2.3 ± 0.5 | 42.0 ± 0.3 | 38.5 ± 0.3 | 3.5 ± 0.4 |
| 224 × 224 | ConvNeXt-B | DINOv3 | 34.8 ± 0.3 | 33.0 ± 0.3 | 1.8 ± 0.5 | 44.2 ± 0.3 | 40.0 ± 0.3 | 4.2 ± 0.5 |
| 512 × 512 | ViT-B | ImageNet | 33.6 ± 0.3 | 32.7 ± 0.3 | 1.0 ± 0.5 | 44.2 ± 0.3 | 40.3 ± 0.3 | 4.0 ± 0.5 |
| 512 × 512 | ViT-B | DINOv2 | 34.7 ± 0.3 | 33.9 ± 0.3 | 0.8 ± 0.5 | 45.2 ± 0.3 | 41.0 ± 0.3 | 4.2 ± 0.5 |
| 512 × 512 | ViT-B | DINOv3 | 35.2 ± 0.3 | 34.3 ± 0.3 | 0.8 ± 0.5 | 45.8 ± 0.3 | 41.9 ± 0.3 | 3.9 ± 0.5 |
| 512 × 512 | ConvNeXt-B | ImageNet | 34.8 ± 0.3 | 33.4 ± 0.3 | 1.4 ± 0.5 | 44.5 ± 0.3 | 40.5 ± 0.3 | 3.9 ± 0.4 |
| 512 × 512 | ConvNeXt-B | DINOv3 | 36.1 ± 0.3 | 35.1 ± 0.3 | 1.0 ± 0.5 | 46.9 ± 0.3 | 42.4 ± 0.3 | 4.5 ± 0.5 |
| mAP values | | | | | | | | |
| 224 × 224 | ViT-B | ImageNet | 33.4 ± 0.3 | 31.4 ± 0.3 | 2.0 ± 0.5 | 42.4 ± 0.3 | 38.3 ± 0.3 | 4.1 ± 0.4 |
| 224 × 224 | ViT-B | DINOv2 | 34.3 ± 0.3 | 32.6 ± 0.3 | 1.7 ± 0.5 | 43.5 ± 0.3 | 39.4 ± 0.3 | 4.1 ± 0.4 |
| 224 × 224 | ViT-B | DINOv3 | 34.2 ± 0.3 | 32.4 ± 0.3 | 1.9 ± 0.5 | 42.6 ± 0.3 | 39.0 ± 0.3 | 3.6 ± 0.4 |
| 224 × 224 | ConvNeXt-B | ImageNet | 33.7 ± 0.3 | 31.4 ± 0.3 | 2.3 ± 0.5 | 42.1 ± 0.3 | 38.5 ± 0.3 | 3.6 ± 0.4 |
| 224 × 224 | ConvNeXt-B | DINOv3 | 34.8 ± 0.3 | 33.1 ± 0.3 | 1.8 ± 0.5 | 44.2 ± 0.3 | 40.0 ± 0.3 | 4.2 ± 0.5 |
| 512 × 512 | ViT-B | ImageNet | 33.7 ± 0.3 | 32.7 ± 0.3 | 1.0 ± 0.5 | 44.3 ± 0.3 | 40.3 ± 0.3 | 4.0 ± 0.5 |
| 512 × 512 | ViT-B | DINOv2 | 34.8 ± 0.3 | 33.9 ± 0.3 | 0.8 ± 0.5 | 45.3 ± 0.3 | 41.1 ± 0.3 | 4.2 ± 0.5 |
| 512 × 512 | ViT-B | DINOv3 | 35.2 ± 0.3 | 34.4 ± 0.3 | 0.8 ± 0.5 | 45.9 ± 0.3 | 42.0 ± 0.3 | 3.9 ± 0.5 |
| 512 × 512 | ConvNeXt-B | ImageNet | 34.8 ± 0.3 | 33.5 ± 0.3 | 1.4 ± 0.5 | 44.5 ± 0.3 | 40.6 ± 0.3 | 3.9 ± 0.4 |
| 512 × 512 | ConvNeXt-B | DINOv3 | 36.2 ± 0.3 | 35.2 ± 0.3 | 1.0 ± 0.5 | 46.9 ± 0.3 | 42.4 ± 0.3 | 4.5 ± 0.5 |

## Frozen billion-parameter features remain inferior to task-adapted mid-sized models

We finally compared frozen DINOv3-7B features coupled to a lightweight classifier with the best task-adapted ViT-B/16 and ConvNeXt-B models obtained under full fine-tuning across all seven datasets and both universal resolutions (**Figure 6e,f, Supplementary Table 6, Supplementary Figure 9**). Despite its much larger scale, the frozen encoder consistently underperformed the best fine-tuned mid-sized models. At 224 × 224, the gap to the best task-adapted model ranged from 4.6 percentage points in Pedi-CXR (68.8 ± 1.3 vs. 73.4 ± 1.2) to 7.4 percentage points in VinDr-CXR (82.8 ± 0.8 vs. 90.2 ± 0.6). Similar deficits were seen in adult cohorts such as ChestX-ray14 (76.1 ± 0.2 vs. 80.4 ± 0.2), CheXpert (76.8 ± 0.2 vs. 80.5 ± 0.2), MIMIC-CXR (77.1 ± 0.2 vs. 81.2 ± 0.1), and UKA-CXR (83.8 ± 0.1 vs. 88.2 ± 0.1).

The same pattern persisted at 512 × 512. Frozen DINOv3-7B reached 69.4 ± 1.3 in Pedi-CXR, 86.8 ± 0.6 in VinDr-CXR, 77.7 ± 0.3 in ChestX-ray14, 78.9 ± 0.2 in CheXpert, 78.9 ± 0.2 in MIMIC-CXR, and 85.4 ± 0.1 in UKA-CXR, whereas the best task-adapted models achieved 74.1 ± 1.2, 90.5 ± 0.5, 82.3 ± 0.2, 82.5 ± 0.1, 82.7 ± 0.2, and 88.5 ± 0.1, respectively. Thus, the frozen-to-fine-tuned gap remained substantial at higher resolution, typically around 3 to 5 percentage points and reaching 4.6 points in ChestX-ray14 and 3.8 points in MIMIC-CXR.

Increasing resolution from 224 × 224 to 512 × 512 did not close this gap systematically (**Figure 6f**). In some datasets the difference narrowed modestly, such as VinDr-CXR and UKA-CXR, whereas in others it remained nearly unchanged or widened slightly, including ChestX-ray14. The overall conclusion therefore did not depend on resolution: in the present setting, larger frozen natural-image representations did not substitute for downstream adaptation. These results show that model scale alone was insufficient to match the performance of task-adapted mid-sized backbones in chest radiograph classification. Even against the 7B-parameter natural-image-pretrained foundation encoder evaluated here, full adaptation of substantially smaller ViT-B/16 and ConvNeXt-B models remained decisively superior, underscoring the importance of domain- and task-specific fine-tuning in medical imaging.

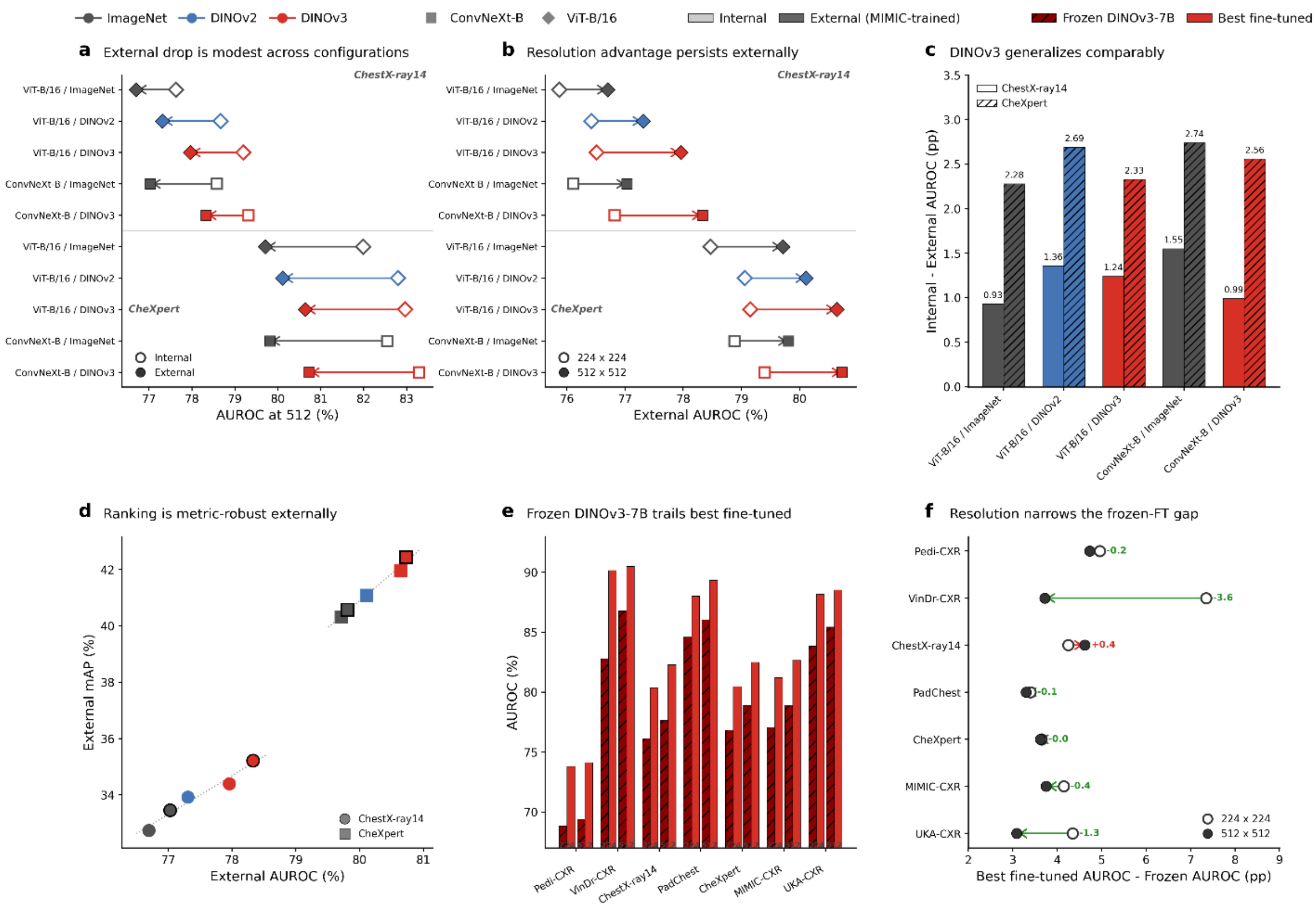

**Figure 6**: External validation and frozen feature comparison. **a** Internal vs. external AUROC at 512×512 for five configurations tested on ChestX-ray14 (test n=25,596), and CheXpert (test n=29,318) after training on MIMIC-CXR (training n=153,255; validation n=18,139; test n=43,793). Dumbbells show internal performance (open circles) and external performance (filled circles) connected by arrows indicating the magnitude of domain shift. External drops range from 0.9 to 2.7 percentage points. ConvNeXt-B/DINOv3 exhibits the smallest external drop on ChestX-ray14 (0.99 pp). **b** Resolution advantage from 224×224 (open) to 512×512 (filled) persists on external test sets, with lifts ranging from +0.83 to +1.51 pp across all configurations and both target datasets. **c** Internal-external AUROC gap per configuration, grouped by target dataset. Hatching distinguishes CheXpert from ChestX-ray14. DINOv3 configurations generalize comparably to or better than ImageNet baselines. **d** External mAP vs. AUROC at 512×512 for all configuration-dataset combinations. Dotted regression lines show strong ranking correspondence. Dataset shapes (circles for ChestX-ray14, squares for CheXpert) confirm metric robustness holds across institutions, addressing concerns about AUROC reliability under class imbalance. **e** Frozen DINOv3-7B linear probe performance (hatched bars) vs. best fine-tuned configuration (solid bars) across seven datasets at 224×224 and 512×512. Frozen features consistently lag fine-tuned mid-sized models by 3 to 7 percentage points at both resolutions. Dataset order: Pedi-CXR, VinDr-CXR, ChestX-ray14, PadChest, CheXpert, MIMIC-CXR, UKA-CXR. **f** Change in frozen-to-fine-tuned gap from 224×224 (open) to 512×512 (filled) per dataset. Green arrows indicate gap narrowing; red indicates widening.

# Discussion

In this large-scale chest radiograph benchmark across n=816,183 images from seven datasets, we found that the benefit of DINOv3 is conditional rather than universal. Six observations emerged. First, DINOv3 transferred well to adult chest radiograph classification, but its advantage over DINOv2 was weak at 224 × 224 pixels and became consistent only at 512 × 512 pixels. Second, this pattern did not extend to the pediatric cohort, which showed little sensitivity to either resolution or initialization. Third, scaling further to 1024 × 1024 pixels rarely improved performance and incurred a marked computational penalty. Fourth, ConvNeXt-B remained stronger than ViT-B/16 under both full fine-tuning and parameter-efficient adaptation. Fifth, the 512 × 512 DINOv3 advantage persisted under external validation, but it could not be explained simply by superior robustness to noisy labels. Sixth, frozen 7B features remained clearly inferior to task-adapted mid-sized models. The results show that modern self-supervised pretraining can improve chest radiograph classification, but its value depends on resolution, cohort type, adaptation strategy, and label quality rather than model scale alone.

Resolution dependence is the central result. At 224 × 224 pixels, DINOv3 generally improved over ImageNet, yet it did not consistently exceed DINOv2. At 512 × 512 pixels, however, DINOv3 became the strongest initialization across most adult cohorts, with the clearest gains seen when paired with ConvNeXt-B. This pattern is consistent with the design of DINOv3, which emphasizes fine-grained feature preservation through Gram-anchored distillation and explicit high-resolution adaptation[22]. It also aligns with the fact that chest radiographs contain clinically relevant structures at much finer spatial scales than are typically preserved in natural-image benchmarks[37]. At the same time, the magnitude of the gain was not uniform across adult datasets, which likely reflects differences in finding composition, label provenance, and cohort characteristics rather than a single dataset-level factor alone. Our cross-dataset finding-group summary supports this interpretation: the largest average gains from 224 × 224 to 512 × 512 were seen for small focal lesions and boundary-dependent findings, whereas large-structure abnormalities such as cardiomegaly and pleural effusion changed little. In practical terms, DINOv3 appears to matter most when diagnostic information is spatially subtle.

The pediatric cohort provides an important boundary condition. Pedi-CXR was not merely lower performing overall, but qualitatively different. It showed minimal separation between initializations and almost no consistent improvement from 224 × 224 to 512 × 512 pixels. Several factors may contribute. The cohort is substantially smaller than the adult datasets, contains only three labels, and focuses on a pediatric disease spectrum that differs from the adult multi-label setting[26,38,39]. Moreover, the labels in Pedi-CXR do not emphasize the same small focal and boundary-dependent abnormalities that benefited most from higher resolution in adults. These findings suggest that the adult DINOv3-at-512 pattern should not be generalized automatically to pediatric chest radiography. More broadly, they indicate that transfer behavior in pediatric imaging may be shaped less by high-resolution representation quality alone and more by cohort size, label ontology, and disease phenotype.

Our targeted 1024 × 1024 analysis sharpened this picture by showing that more resolution is not necessarily better. Across the three representative cohorts evaluated at 1024 × 1024, formal cross-resolution testing showed only two significant gains, both in MIMIC-CXR with ConvNeXt-B, whereas the remaining comparisons were neutral or negative. At the same time, the cost increase was substantial: from 512 × 512 to 1024 × 1024, per-epoch training time rose by 6.2-fold overall and by 7.9-fold for ViT-B/16, while total training time to convergence increased by a median of 4.8-fold overall and 8.9-fold for ViT-B/16. In the present setting, 512 × 512 pixels was therefore the best performance-cost operating point. This conclusion should, however, be interpreted within the scope of the efficiency analysis performed here, which was based on measured training time rather than inference latency, GPU memory consumption, or FLOPs. It should also be interpreted within the present architectural setup, as alternative high-resolution designs, such as different patch sizes or adaptive pooling strategies, could potentially shift the resolution-performance trade-off.

The backbone findings were also robust to changes in adaptation regime. Under full fine-tuning, ConvNeXt-B consistently outperformed ViT-B/16, particularly at 512 × 512 pixels. Parameter-efficient adaptation, implemented here with LoRA, did not narrow this gap. Instead, it reduced performance in all evaluated settings and generally affected ViT-B/16 more strongly than ConvNeXt-B, especially for DINOv3 initialization. This suggests that the ConvNeXt advantage cannot be explained simply by the use of full end-to-end fine-tuning. A plausible interpretation is that convolutional inductive biases and hierarchical local feature aggregation remain advantageous in chest radiograph analysis, particularly when diagnostically relevant information is spatially local and when high-resolution inputs emphasize subtle boundaries and fine structures. This interpretation is also consistent with our scaling results, in which ConvNeXt-B incurred a smaller computational penalty than ViT-B/16 at higher resolution and was the only backbone to show significant gains at 1024 × 1024 in MIMIC-CXR. While this does not prove that convolutional backbones are universally superior to transformers in radiology, it does show that, at base scale and under both full and parameter-efficient adaptation, ConvNeXt-B provided the more reliable transfer behavior in our benchmark.

The additional experiments also clarify how DINOv3 should and should not be interpreted in the presence of weak supervision. In the controlled label-noise experiment on VinDr-CXR, DINOv3 performed best under clean labels but degraded more steeply as corruption increased, ultimately falling below both DINOv2 and ImageNet at 40% noise. This is an important cautionary result. The gains we observed on large report-labeled cohorts such as MIMIC-CXR and CheXpert should not be attributed simply to superior robustness to noisy labels. Instead, DINOv3 appears to be strongest when the downstream supervision remains reasonably informative, while its advantage can diminish or reverse under sufficiently corrupted targets. In other words, DINOv3 may help on weakly labeled clinical datasets, but that benefit is not equivalent to intrinsic noise immunity.

At the same time, the external-validation experiment shows that the main adult finding is not merely an artifact of within-dataset splits. When models were trained on MIMIC-CXR and evaluated on the fully held-out external datasets ChestX-ray14 and CheXpert using a harmonized seven-label setting, external AUROC drops were modest and the relative ordering of configurations was largely preserved. DINOv3 remained the strongest initialization at 512 × 512 pixels for both backbones on both external target datasets, and the external advantage of 512 × 512 over 224 × 224 also persisted. Importantly, PR-AUC and mAP tracked AUROC closely in these analyses, arguing against the possibility that the observed ranking depended mainly on the choice of metric. Together, the noise and external-validation results refine the overall interpretation: DINOv3 generalizes across institutions, but its gains are better understood as improved transferable representation quality than as broad robustness to corrupted supervision. At the same time, this experiment used a single training source and a harmonized label subset rather than a full all-to-all cross-dataset design, so broader external generalization should still be tested in future work.

The frozen 7B comparison provides a final practical lesson. Despite its scale, frozen DINOv3-7B with a lightweight classifier was consistently inferior to the best fine-tuned base models across all datasets and both universal resolutions. The gap remained substantial even at 512 × 512 pixels and did not close systematically with higher resolution. This result is not entirely surprising, because the frozen encoder was pretrained on natural images and never adapted to the clinical domain. Still, it is informative: in the present setting, model size alone did not compensate for the absence of task-specific adaptation. For chest radiograph classification, carefully adapted 86 to 89 million-parameter models delivered more diagnostic value than the much larger frozen vision encoder evaluated here. An important next step will be to compare such frozen natural-image features directly against frozen representations from medical-domain self-supervised models, in order to disentangle the effect of domain mismatch from the broader question of whether frozen representations can approach the performance of task-adapted models in radiology.

From a clinical perspective, the magnitude of the gains in adult cohorts was modest in absolute AUROC terms, often around 0.5 to 1.0 points, but the distribution of those gains matters. Improvements concentrated in small focal lesions and boundary-dependent findings are more likely to be relevant to emergency, inpatient, and triage settings than improvements on already easy large-structure labels. A system that better resolves subtle pneumothorax, small nodules, or faint focal opacities can be clinically valuable even when the mean AUROC shift seems small. We therefore do not propose a single universal probability threshold from the present benchmark, because the appropriate operating point depends on the intended clinical task, local prevalence, and the relative cost of false-negative and false-positive decisions. In practice, triage-oriented use cases would typically favor higher sensitivity, whereas workflow-prioritization or rule-out settings may require a different trade-off. Our results therefore support a pragmatic recommendation at the model-selection level: when using off-the-shelf self-supervised visual initialization for adult chest radiograph classification, prioritize 512 × 512 inputs and a ConvNeXt-B backbone, and treat further scaling in size or resolution cautiously unless a clear task-specific benefit is shown.

This study has several limitations. First, although we included seven datasets from multiple countries and both public and internal sources, the main benchmark remained limited to AP and PA chest radiographs and to classification tasks; the relevance of these conclusions to lateral views, localization, segmentation, or report generation remains to be established. Indeed, because DINOv3 is designed to preserve dense patch-level structure and our strongest gains were observed for small focal and boundary-dependent findings, extending this analysis to localization and segmentation tasks represents a particularly important future direction. In addition, the relative distribution of AP and PA projections differed across datasets, so projection type remained partially confounded with dataset and may have contributed to apparent cross-dataset differences in performance. Second, the 1024 × 1024 analysis was deliberately targeted rather than exhaustive, and only three cohorts were evaluated at that resolution. This design was chosen because of the substantial computational burden of full high-resolution training and was intended as a focused test of whether scaling beyond 512 × 512 yielded meaningful additional benefit under the present training setup. Third, the backbone comparison under parameter-efficient adaptation was restricted to LoRA and to three adult datasets, so broader conclusions about all parameter-efficient fine-tuning strategies or all model scales would be premature. Fourth, the external-validation analysis used a harmonized seven-label subset and one training source, not a full all-to-all cross-dataset design. Fifth, the frozen-feature comparison was restricted to a natural-image-pretrained DINOv3-7B encoder and did not include frozen medical-domain self-supervised encoders, including emerging DINOv3-based medical adaptations such as MedDINOv3[25], so it does not by itself distinguish domain mismatch from a more general limitation of frozen representations. Sixth, the label-noise experiment used synthetic symmetric corruption in a single dataset, which provides a controlled stress test but cannot capture all properties of real report-derived labeling noise. In addition, class imbalance was handled using inverse-frequency weighting, but we did not systematically compare alternative strategies such as focal loss, oversampling, or other cost-sensitive objectives, which may be relevant particularly for rare findings. Finally, although class-imbalance-aware metrics supported the main follow-up analyses, AUROC remained the primary metric for the benchmark as a whole, and the study did not include prospective clinical validation or workflow-level evaluation.

In conclusion, DINOv3 improves adult chest radiograph classification most reliably at 512 × 512 pixels, especially when paired with ConvNeXt-B, but this advantage is conditional rather than universal. It does not extend clearly to the pediatric cohort, it rarely improves further at 1024 × 1024, it is not explained by superior noise robustness, and it does not remove the need for task-specific fine-tuning. The strongest overall message is therefore practical: for adult chest radiograph classification, the best trade-off in our benchmark came from fully adapted mid-sized backbones operating at 512 × 512 pixels, whereas larger frozen encoders and higher resolutions offered limited additional value. These findings provide concrete guidance for applying modern self-supervised representations in radiology and clarify where their benefits are most likely to matter in practice.

# Materials and methods

## Ethics statement

All methods were carried out in accordance with relevant guidelines and regulations. Ethical approval for this retrospective study was obtained from the Ethics Committee of the Medical Faculty of RWTH Aachen University (Reference No. EK 22-319). The requirement for individual informed consent was waived by the committee.

# Patient datasets

This study included a total of n = 816,183 AP or PA chest radiographs from seven international cohorts encompassing both adult and pediatric populations. Patients ranged in age from infancy to over 111 years. The datasets span diverse geographic regions (Asia, Europe, and North America), label generation strategies (manual annotation, rule-based natural language processing (NLP), and hybrid approaches), and clinical contexts (inpatient, outpatient, intensive care, and pediatrics). A detailed overview of dataset characteristics is provided in **Table 1**. Below, we describe each dataset. For datasets with predefined train and test partitions but without patient identifiers (Pedi-CXR and VinDr-CXR), no additional validation split was created in order to avoid potential patient-level information leakage.

### *Pedi-CXR dataset*

The Pedi-CXR[26] dataset is the largest publicly available pediatric chest radiograph dataset with diagnostic labels. It contains 9,125 PA images from children under the age of 10 years, collected in Vietnam. All radiographs were manually annotated by three radiologists with at least 10 years of experience. For this study, we followed the dataset's original split into training (n = 7,728) and test (n = 1,397) sets. No separate validation split was created, because the dataset does not provide patient identifiers and therefore any additional random subdivision of the training set would have carried a risk of patient-level information leakage. Labels include pneumonia and related pediatric conditions (see **Table 1**).

### *VinDr-CXR dataset*

The VinDr-CXR[27] dataset comprises 18,000 adult radiographs, curated from more than 100,000 studies performed at two Vietnamese hospitals. Images were acquired on equipment from multiple manufacturers. Labeling was performed by 17 radiologists, with each image independently annotated by three experts. The dataset authors provided a split into n = 15,000 training and n = 3,000 test images, which we used directly. No separate validation split was created, because the dataset does not provide patient identifiers and therefore additional random sampling from the training set could have introduced patient-level information leakage across splits. Labels cover common thoracic diseases such as cardiomegaly, effusion, and pneumonia (see **Table 1**).

### *ChestX-ray14 dataset*

The ChestX-ray14[28] dataset, released by the National Institutes of Health, contains n = 112,120 radiographs from 30,805 patients. Fourteen thoracic pathologies were labeled using a two-stage NLP pipeline applied to corresponding radiology reports. Following prior work, we first generated a patient-wise 80%/20% train-test split and then reserved 10% of the training portion for validation, resulting in n = 77,870 training, n = 8,654 validation, and n = 25,596 test images. Labels span major cardiopulmonary conditions (see **Table 1**).

### *PadChest dataset*

The PadChest[29] dataset includes n = 110,525 AP or PA radiographs from 67,205 patients at Hospital Universitario de San Juan in Alicante, Spain. Labels were derived from radiology reports in Spanish: 27,593 studies were manually annotated by radiologists, and the remainder were automatically labeled using a text classifier trained on this subset. We performed a patient-wise 80%/20% split, stratified by manual vs. automatic labeling, and then reserved 10% of the training portion for validation, yielding n = 79,697 training, n = 8,783 validation, and n = 22,045 test images. Labels are diverse and include both common and less frequent findings (see **Table 1**).

### *CheXpert dataset*

The CheXpert[30] dataset consists of n = 157,865 AP or PA chest radiographs from 57,872 patients at Stanford Hospital in California, USA. Labels for common radiographic findings were extracted using a rule-based NLP system that categorized mentions as positive, negative, or uncertain. Following established practice, uncertain and negative mentions were grouped as negative. We used a patient-wise 80%/20% split and then reserved 10% of the training portion for validation, resulting in n = 115,449 training, n = 13,098 validation, and n = 29,318 test images. Labels include cardiomegaly, effusion, pneumonia, and others (see **Table 1**).

### *MIMIC-CXR dataset*

The MIMIC-CXR[6] dataset contains n = 215,187 AP or PA radiographs from 62,094 patients at Beth Israel Deaconess Medical Center in Boston, Massachusetts, USA, collected between 2011 and 2016. Images were de-identified and linked to associated reports. Labels were generated automatically using the same NLP system as CheXpert30, ensuring consistency. We created a patient-wise 80%/20% split and then reserved 10% of the training portion for validation, yielding n = 153,255 training, n = 18,139 validation, and n = 43,793 test images. Labels overlap with CheXpert, covering major cardiopulmonary findings (see **Table 1**).

### *UKA-CXR dataset*

The UKA-CXR[21,31–35] dataset is an internal cohort from University Hospital RWTH Aachen in Aachen, Germany. It includes n = 193,361 adult radiographs from 54,176 patients, collected between 2009 and 2020 across 10 intensive care units using 18 radiography systems. Images were labeled by radiologists within the clinical reporting workflow using a structured template with categories such as pleural effusion, pneumonia, atelectasis, congestion, and cardiomegaly. For this study, we defined a patient-wise 80%/20% split and then reserved 10% of the training portion for validation, yielding n = 137,902 training, n = 15,353 validation, and n = 40,106 test images. Labels reflect routine diagnostic categories from clinical reporting (see **Table 1**).

## Label system and preprocessing

As in our prior works[21,31–35,39,40], all datasets were mapped into a unified binary multilabel classification framework, in which each image was assigned a positive or negative label for every included condition. Only AP or PA views were used in all experiments. Pedi-CXR, VinDr-CXR, ChestX-ray14, and PadChest were provided in binary format by design and were used directly. In CheXpert, and consequently in MIMIC-CXR, the original four categories ("positive," "negative," "uncertain," and "not mentioned") were reduced to binary by treating "negative," "uncertain," and "not mentioned" as negative, and only "positive" as positive. For the UKA-CXR dataset, which contained multiple severity levels, "normal" and "uncertain" were classified as negative, whereas all severity categories above normal were classified as positive, for example "mild," "moderate," or "severe" for pleural effusion and "borderline," "enlarged," or "massively enlarged" for cardiomegaly. In addition, UKA-CXR contained separate left- and right-sided labels for several findings; in these cases, the presence of a finding on either side was counted as positive. In PadChest, where annotations were generated through a combination of manual labeling and NLP, only the subset of labels overlapping with the target label system was retained and binarized accordingly. Finally, whenever available, the "no finding" label was preserved as a separate category to indicate a completely normal radiograph without any imaging abnormality, rather than merely the absence of the labels considered in this study.

Images were supplied in mixed formats depending on the dataset. ChestX-ray14, PadChest, CheXpert, and MIMIC-CXR were already available as PNG or JPG files, whereas Pedi-CXR, VinDr-CXR, and UKA-CXR were provided in DICOM format and converted to PNG or JPG before analysis. For DICOM images, metadata were checked to ensure correct polarity; if pixel intensities were stored in inverted form, they were re-inverted to maintain consistent orientation. All radiographs were resized to 224 × 224, 512 × 512, or 1024 × 1024 pixels, depending on the experiment. To normalize intensity values, each image was shifted so that the minimum pixel value corresponded to zero, scaled by its maximum pixel value, and clipped to the valid range before conversion to 8-bit grayscale[6]. Contrast was then enhanced by histogram equalization implemented with the OpenCV library[6,32]. These steps yielded a uniform preprocessing pipeline across datasets, with strict separation of training, validation, and test cohorts wherever applicable.

# Experimental design

### ***Pretrained initialization strategies and backbone architectures***

This study was designed as a controlled transfer-learning benchmark rather than a medical-domain pretraining study. Accordingly, DINOv2[20] and DINOv3[22] were not retrained on chest radiographs; instead, we used the publicly released pretrained checkpoints directly as initialization for downstream chest radiograph classification. Relative to DINOv2, DINOv3 extends the same general self-supervised framework by introducing Gram-anchored refinement to stabilize dense patch representations during extended training and explicit high-resolution adaptation to support transfer to larger input sizes[22]. These properties motivated our evaluation across resolutions, architectures, and adaptation regimes.

The main benchmark compared three initialization strategies: supervised ImageNet-21K[7], self-supervised DINOv2, and self-supervised DINOv3. Two backbone families were evaluated: the Vision Transformer base model (ViT-B/16, approximately 86 million parameters) and ConvNeXt-B (approximately 89 million parameters). Because a matching public ConvNeXt-B DINOv2 checkpoint was not available, DINOv2 experiments were performed only with ViT-B/16. The full main benchmark therefore comprised five full fine-tuning configurations: ViT-B/16 initialized from ImageNet, DINOv2, or DINOv3, and ConvNeXt-B initialized from ImageNet or DINOv3.

### *Main full fine-tuning benchmark*

Universal experiments were performed at 224 × 224 and 512 × 512 pixels across all seven datasets. To probe whether the 512 × 512 advantage extended further, we additionally evaluated 1024 × 1024 pixels on three representative cohorts spanning pediatric and adult settings, namely Pedi-CXR, ChestX-ray14, and MIMIC-CXR.

All full fine-tuning experiments were optimized with AdamW[41] using a learning rate of $10^{-5}$ and no weight decay. This learning rate was chosen based on empirical testing of multiple candidate values in the present study, where it provided the most stable convergence across datasets and model configurations, and it is also consistent with the range used in our prior work on chest radiograph classification[21,32–34,39,40]. Across all experiments, data augmentation consisted of random horizontal flips and random rotations up to 7°. The loss function was binary weighted cross-entropy, with class weights set inversely proportional to the frequency of each label in the training set[42]. All training was performed in full 32-bit floating point precision on a single NVIDIA L40S GPU with 48 GB VRAM and Intel Xeon Silver 4310 CPUs. For full fine-tuning, batch size was 128 at 224 × 224, 64 at 512 × 512, and 16 at 1024 × 1024. To preserve comparability across initialization strategies, backbones, and resolutions, all main benchmark models were trained under this shared optimization protocol rather than under separate per-configuration hyperparameter searches.

### *Frozen-feature probing with DINOv3-7B*

To test whether a much larger frozen foundation encoder could substitute for downstream adaptation, we additionally evaluated frozen representations from the DINOv3-7B model. In this setting, the backbone remained fully frozen and only a compact multilayer classification head, referred to as DINO-Net, was optimized. DINO-Net consisted of layer normalization[43] applied to the 4096-dimensional backbone features, followed by a Gaussian error linear unit (GELU)[44] activation and dropout with p = 0.3, a linear projection to a 512-dimensional embedding, a second dropout layer with p = 0.3, a second layer normalization, and a final linear mapping to the target label space. This head added approximately 2.1 million trainable parameters. Frozen-feature experiments were performed on all seven datasets at 224 × 224 and 512 × 512 pixels. Only the DINO-Net classifier was optimized, using AdamW with a learning rate of $10^{-4}$ and weight decay of $5 \times 10^{-5}$.

### *Parameter-efficient adaptation with LoRA*

To test whether the observed backbone ranking depended on the use of full end-to-end fine-tuning, we performed a dedicated parameter-efficient adaptation analysis using low-rank adaptation (LoRA)[36]. These experiments were conducted at 512 × 512 pixels on three adult datasets, namely ChestX-ray14, CheXpert, and MIMIC-CXR. We evaluated the same five backbone-initialization combinations as in the full fine-tuning benchmark: ViT-B/16 with ImageNet, DINOv2, or DINOv3 initialization, and ConvNeXt-B with ImageNet or DINOv3 initialization.

For ViT-B/16, LoRA adapters were applied to the attention projection layers, specifically query and value projections or their architecture-equivalent implementation. For ConvNeXt-B, adapters were applied to the pointwise projection layers within the ConvNeXt blocks, again using the corresponding implementation-specific module names. In all LoRA experiments, the classification head remained trainable. LoRA hyperparameters were fixed to rank 16, alpha 32, and dropout 0.05. Training used AdamW with learning rate $10^{-4}$ and batch size 32.

### *Synthetic label-noise experiment*

To test whether gains from DINOv3 on weakly labeled adult datasets could be explained simply by improved tolerance to noisy supervision, we performed a controlled synthetic label-noise experiment on VinDr-CXR using ViT-B/16 at 512 × 512 pixels. This analysis compared the three ViT-B/16 initialization strategies, namely ImageNet, DINOv2, and DINOv3.

Noise was injected only into the training split, while the test set remained unchanged. Corruption was applied at predefined noise levels of 10%, 20%, 30%, and 40%, with the clean training labels serving as the 0% reference. Noise was introduced only for the 10 abnormal labels in the VinDr-CXR setup, excluding the "No finding" label. We used an asymmetric corruption scheme intended to better approximate weak-label errors: positive labels were flipped to negative at the specified noise rate, whereas negative labels were flipped to positive at one quarter of that rate. After corruption of the abnormal labels, the "No finding" label was recomputed on the training set so that it remained mutually consistent with the pathology labels. Test labels were never modified.

### *External validation across institutions*

To determine whether the main benchmark findings extended beyond within-dataset train-test splits, we performed an external validation experiment using a harmonized seven-label setting comprising cardiomegaly, pleural effusion, pneumonia, atelectasis, no finding, consolidation, and pneumothorax. Models were trained on MIMIC-CXR and evaluated both on the internal MIMIC-CXR test set and on the fully held-out external datasets ChestX-ray14 and CheXpert. External validation was performed at 224 × 224 and 512 × 512 pixels for the same five backbone-initialization combinations used in the main benchmark.

### *Computational efficiency analysis*

To place performance differences in a practical context, computational efficiency was assessed for all full fine-tuning experiments by measuring both per-epoch training time and total wall-clock time to convergence on the same hardware platform, namely a single NVIDIA L40S GPU with 48

GB VRAM and Intel Xeon Silver 4310 CPUs, using full 32-bit floating point precision throughout. This analysis was intended as a practical efficiency comparison within the present benchmark rather than as a formal systems-level profiling study; accordingly, it did not include inference latency, GPU memory traces, or FLOPs.

Per-epoch training time was recorded for each evaluated full fine-tuning configuration, defined by dataset, initialization, backbone, and image resolution. Across the universal 224 × 224 and 512 × 512 experiments, this yielded measurements for all seven datasets, three initialization strategies for ViT-B/16, and two initialization strategies for ConvNeXt-B. For the targeted 1024 × 1024 setting, measurements were obtained only for the three representative cohorts evaluated at that resolution, namely Pedi-CXR, ChestX-ray14, and MIMIC-CXR. For each configuration, we also recorded the number of epochs required to convergence, allowing computation of total wall-clock training time.

Relative computational cost was then summarized as ratios across resolution transitions, particularly for 224 × 224 to 512 × 512 and 512 × 512 to 1024 × 1024. These ratios were aggregated across the relevant sets of configurations and reported descriptively using means, standard deviations, medians, and ranges. This analysis allowed us to relate performance changes directly to measured training cost under a fixed implementation and hardware setting.

## Evaluation and statistical analysis

The primary evaluation metric was the area under the receiver operating characteristic curve (AUROC), which provides a threshold-independent measure of discrimination in multilabel classification. For the main benchmark, we summarized performance within each dataset using the mean AUROC across all labels. Per-label AUROC values are provided in the supplementary information. Accuracy, sensitivity, and specificity were computed as complementary metrics, with operating thresholds chosen according to Youden's criterion[45], i.e., the cut-off maximizing the difference between true-positive and false-positive rates. For follow-up analyses involving class imbalance, including parameter-efficient adaptation, synthetic label corruption, and external validation, we additionally report precision-recall area under the curve (PR-AUC) and mean average precision (mAP).

Statistical analysis was performed using Python 3.9 with NumPy 1.22, SciPy 1.10, and scikit-learn 1.2. Bootstrapping with 1,000 redraws was used to estimate means, standard deviations, and 95% confidence intervals (CI)[21,46]. For within-dataset comparisons between initialization strategies, backbones, resolutions, or adaptation regimes, paired[47] bootstrap testing was performed on AUROC differences using identical resampling across compared models[21,31,40]. Formal cross-resolution testing between 512 × 512 and 1024 × 1024 was performed for all configurations evaluated at both resolutions. To account for multiple comparisons across related groups of tests, p-values were adjusted using the Benjamini-Hochberg false discovery rate (FDR) procedure, with statistical significance defined as FDR-adjusted as $p < 0.05$[48].

External validation experiments were performed by training on MIMIC-CXR and testing on ChestX-ray14 and CheXpert using a harmonized seven-label setting comprising cardiomegaly, pleural effusion, pneumonia, atelectasis, no finding, consolidation, and pneumothorax. The synthetic label-noise experiment was conducted on VinDr-CXR by progressively corrupting training labels at predefined noise levels while keeping the test set unchanged. Computational efficiency was assessed using measured per-epoch training time and total wall-clock time to convergence for each evaluated configuration on the same hardware platform, and scaling behavior was summarized as relative cost ratios across image resolutions.

## Data availability

The datasets used in this study are either publicly accessible, available under controlled access, or internal. The ChestX-ray14 and PadChest datasets are publicly available at https://www.kaggle.com/datasets/nih-chest-xrays/data and https://bimcv.cipf.es/bimcv-projects/padchest/, respectively. The VinDr-CXR and MIMIC-CXR datasets are restricted-access resources hosted on PhysioNet and can be obtained by agreeing to the relevant data protection requirements at https://physionet.org/content/vindr-cxr/1.0.0/ and https://physionet.org/content/mimic-cxr-jpg/2.0.0/. The Pedi-CXR dataset (VinDr-PCXR) is also available through PhysioNet at https://physionet.org/content/vindr-pcxr/1.0.0/. The CheXpert dataset may be requested from Stanford University at https://stanfordmlgroup.github.io/competitions/chexpert/. The UKA-CXR dataset contains patient data from the University Hospital Aachen, Germany; access may be granted upon reasonable request to the corresponding authors and within a written cooperation agreement. A subset of the UKA-CXR dataset is publicly available on Hugging Face via https://huggingface.co/TLAIM.

## Code availability and reproducibility

All source code, configuration files, and instructions to reproduce the experiments are available at https://github.com/tayebiarasteh/vit-med. Training and evaluation were performed strictly in full 32-bit floating point (FP32) precision. Experiments were conducted between August 13, 2025, and September 22, 2025.

Implementation details: Python 3.9 with PyTorch 2.8 and torchvision 0.23. Core libraries: NumPy 1.22, SciPy 1.10, scikit-learn 1.2, pandas 1.4, timm 0.6, and OpenCV (cv2) 4.7. Hugging Face tooling: transformers 4.56, huggingface-hub 0.34, datasets 2.19, accelerate 1.10, tokenizers 0.21, and safetensors 0.4.

Pretrained initialization weights were obtained from official public repositories:

- ImageNet-21K:

- ViT-B/16 from timm (model identifier: vit_base_patch16_224_in21k), with the corresponding Hugging Face model repository:
        - https://huggingface.co/timm/vit_base_patch16_224.orig_in21k
    - ConvNeXt-B from Hugging Face (loaded using the safetensors format): https://huggingface.co/facebook/convnext-base-224-22k
- DINOv2:
    - ViT-B/16 from Hugging Face, configured with scaled-dot product attention: https://huggingface.co/facebook/dinov2-base
- DINOv3:
    - ViT-B/16 from Hugging Face, configured with scaled-dot product attention: https://huggingface.co/facebook/dinov3-vitb16-pretrain-lvd1689m
    - ViT-7B from Hugging Face, configured with scaled-dot product attention: https://huggingface.co/facebook/dinov3-vit7b16-pretrain-lvd1689m
    - ConvNeXt-B from Hugging Face (loaded using the safetensors format): https://huggingface.co/facebook/dinov3-convnext-base-pretrain-lvd1689m

# Additional information

## Funding

JNK is supported by the German Cancer Aid (DECADE, 70115166), the German Federal Ministry of Education and Research (PEARL, 01KD2104C; CAMINO, 01EO2101; SWAG, 01KD2215A; TRANSFORM LIVER, 031L0312A; TANGERINE, 01KT2302 through ERA-NET Transcan; Come2Data, 16DKZ2044A; DEEP-HCC, 031L0315A), the German Academic Exchange Service (SECAI, 57616814), the German Federal Joint Committee (TransplantKI, 01VSF21048) the European Union's Horizon Europe and innovation programme (ODELIA, 101057091; GENIAL, 101096312), the European Research Council (ERC; NADIR, 101114631), the National Institutes of Health (EPICO, R01 CA263318) and the National Institute for Health and Care Research (NIHR, NIHR203331) Leeds Biomedical Research Centre. The views expressed are those of the author(s) and not necessarily those of the NHS, the NIHR or the Department of Health and Social Care. This work was funded by the European Union. Views and opinions expressed are however those of the author(s) only and do not necessarily reflect those of the European Union. Neither the European Union nor the granting authority can be held responsible for them. SN was supported by grants from the Deutsche Forschungsgemeinschaft (DFG) (NE 2136/3-1, LI3893/6-1, TR 1700/7-1). DT was supported by grants from the DFG (NE 2136/3-1, LI3893/6-1, TR 1700/7-1) and is supported by the German Federal Ministry of Education (TRANSFORM LIVER, 031L0312A; SWAG, 01KD2215B) and the European Union’s Horizon Europe and innovation programme (ODELIA [Open Consortium for Decentralized Medical Artificial Intelligence], 101057091).

## Author contributions

The formal analysis was conducted by STA, SN, and DT. The original draft was written by STA and edited by STA, MS, SN, and DT. The code was developed by STA. The experiments were performed by STA. The illustrations were designed by MS. The statistical analyses were performed by STA, SN, and DT. STA, CK, JNK, SN, and DT provided clinical expertise. STA, MS, JNK, and DT provided technical expertise. The study was defined by STA, SN, and DT. All authors read the manuscript and agreed to the submission of this paper.

## Competing interests

STA is an editorial board at *Communications Medicine* and at *European Radiology Experimental*, and a trainee editorial board at *Radiology: Artificial Intelligence*. JNK declares consulting services for Bioptimus, France; Owkin, France; DoMore Diagnostics, Norway; Panakeia, UK; AstraZeneca, UK; Scailyte, Switzerland; Mindpeak, Germany; and MultiplexDx, Slovakia. Furthermore, he holds shares in StratifAI GmbH, Germany, and in Synagen GmbH, Germany, has received a research grant by GSK, and has received honoraria by AstraZeneca, Bayer, Eisai, Janssen, MSD, BMS, Roche, Pfizer and Fresenius. DT received honoraria for lectures by Bayer, GE, Roche, AstraZeneca, and Philips and holds shares in StratifAI GmbH, Germany, and in Synagen GmbH, Germany. The other authors do not have any competing interests to disclose.

## References

1. Rajpurkar, P., Chen, E., Banerjee, O. & Topol, E. J. AI in health and medicine. *Nat Med* **28**, 31–38 (2022).
2. Tayebi Arasteh, S. *et al.* Large language models streamline automated machine learning for clinical studies. *Nat Commun* **15**, 1603 (2024).
3. Haug, C. J. & Drazen, J. M. Artificial Intelligence and Machine Learning in Clinical Medicine, 2023. *N Engl J Med* **388**, 1201–1208 (2023).
4. Tayebi Arasteh, S. *et al.* The Treasure Trove Hidden in Plain Sight: The Utility of GPT-4 in Chest Radiograph Evaluation. *Radiology* **313**, e233441 (2024).
5. Chen, Z. *et al.* A Vision-Language Foundation Model to Enhance Efficiency of Chest X-ray Interpretation. Preprint at https://doi.org/10.48550/arXiv.2401.12208 (2024).
6. Johnson, A. E. W. *et al.* MIMIC-CXR, a de-identified publicly available database of chest radiographs with free-text reports. *Sci Data* **6**, 317 (2019).
7. Deng, J. *et al.* ImageNet: A large-scale hierarchical image database. in *2009 IEEE Conference on Computer Vision and Pattern Recognition* 248–255 (IEEE, Miami, FL, 2009). doi:10.1109/CVPR.2009.5206848.
8. Ke, A., Ellsworth, W., Banerjee, O., Ng, A. Y. & Rajpurkar, P. CheXtransfer: performance and parameter efficiency of ImageNet models for chest X-Ray interpretation. in *Proceedings of the Conference on Health, Inference, and Learning* 116–124 (ACM, Virtual Event USA, 2021). doi:10.1145/3450439.3451867.
9. Krishnan, R., Rajpurkar, P. & Topol, E. J. Self-supervised learning in medicine and healthcare. *Nat. Biomed. Eng* **6**, 1346–1352 (2022).
10. Hendrycks, D., Mazeika, M., Kadavath, S. & Song, D. Using self-supervised learning can improve model robustness and uncertainty. in *NIPS'19: Proceedings of the 33rd International Conference on Neural Information Processing Systems* vol. 1403 15663–15674 (2019).
11. He, K., Fan, H., Wu, Y., Xie, S. & Girshick, R. Momentum Contrast for Unsupervised Visual Representation Learning. in *Proceedings of the IEEE/CVF Conference on Computer Vision and Pattern Recognition (CVPR)* 9729–9738 (2020).
12. Chen, T., Kornblith, S., Norouzi, M. & Hinton, G. A Simple Framework for Contrastive Learning of Visual Representations. in *International Conference on Machine Learning* vol. 119 (Vienna, Austria, 2020).
13. Grill, J.-B. *et al.* Bootstrap your own latent-a new approach to self-supervised learning. *Advances in neural information processing systems* **33**, 21271–21284 (2020).
14. Caron, M. *et al.* Unsupervised Learning of Visual Features by Contrasting Cluster Assignments. in *Advances in neural information processing systems 33* 9912–9924 (2020).
15. Wen, Y., Chen, L., Deng, Y. & Zhou, C. Rethinking pre-training on medical imaging. *Journal of Visual Communication and Image Representation* **78**, 103145 (2021).
16. Vaswani, A. *et al.* Attention Is All You Need. in *NIPS'17: Proceedings of the 31st International Conference on Neural Information Processing Systems* 6000–6010 (2017).
17. Dosovitskiy, A. *et al.* An Image is Worth 16x16 Words: Transformers for Image Recognition at Scale. Preprint at http://arxiv.org/abs/2010.11929 (2021).
18. Liu, Z. *et al.* A convnet for the 2020s. in *Proceedings of the IEEE/CVF conference on computer vision and pattern recognition* 11976–11986 (2022).
19. Caron, M. *et al.* Emerging Properties in Self-Supervised Vision Transformers. in *Proceedings of the IEEE/CVF International Conference on Computer Vision (ICCV)* 9650–9660 (2021).
20. Oquab, M. *et al.* DINOv2: Learning Robust Visual Features without Supervision. Preprint at http://arxiv.org/abs/2304.07193 (2023).

21. Tayebi Arasteh, S., Misera, L., Kather, J. N., Truhn, D. & Nebelung, S. Enhancing diagnostic deep learning via self-supervised pretraining on large-scale, unlabeled non-medical images. *Eur Radiol Exp* **8**, 10 (2024).
22. Siméoni, O. *et al.* DINOv3. Preprint at https://doi.org/10.48550/arXiv.2508.10104 (2025).
23. Liu, C. *et al.* Does DINOv3 Set a New Medical Vision Standard? Preprint at https://doi.org/10.48550/arXiv.2509.06467 (2025).
24. Yang, S., Wang, H., Xing, Z., Chen, S. & Zhu, L. SegDINO: An Efficient Design for Medical and Natural Image Segmentation with DINO-V3. Preprint at https://doi.org/10.48550/arXiv.2509.00833 (2025).
25. Li, Y., Wu, Y., Lai, Y., Hu, M. & Yang, X. MedDINOv3: How to adapt vision foundation models for medical image segmentation? Preprint at https://doi.org/10.48550/arXiv.2509.02379 (2025).
26. Nguyen, N. H., Pham, H. H., Tran, T. T., Nguyen, T. N. M. & Nguyen, H. Q. *VinDr-PCXR: An Open, Large-Scale Chest Radiograph Dataset for Interpretation of Common Thoracic Diseases in Children*. http://medrxiv.org/lookup/doi/10.1101/2022.03.04.22271937 (2022) doi:10.1101/2022.03.04.22271937.
27. Nguyen, H. Q. *et al.* VinDr-CXR: An open dataset of chest X-rays with radiologist's annotations. *Sci Data* **9**, 429 (2022).
28. Wang, X. *et al.* ChestX-ray8: Hospital-scale Chest X-ray Database and Benchmarks on Weakly-Supervised Classification and Localization of Common Thorax Diseases. in *2017 IEEE Conference on Computer Vision and Pattern Recognition (CVPR)* 3462–3471 (2017). doi:10.1109/CVPR.2017.369.
29. Bustos, A., Pertusa, A., Salinas, J.-M. & de la Iglesia-Vayá, M. PadChest: A large chest x-ray image dataset with multi-label annotated reports. *Medical Image Analysis* **66**, 101797 (2020).
30. Irvin, J. *et al.* CheXpert: A Large Chest Radiograph Dataset with Uncertainty Labels and Expert Comparison. *AAAI* **33**, 590–597 (2019).
31. Khader, F. *et al.* Artificial Intelligence for Clinical Interpretation of Bedside Chest Radiographs. *Radiology* **307**, e220510 (2022).
32. Tayebi Arasteh, S. *et al.* Collaborative training of medical artificial intelligence models with non-uniform labels. *Sci Rep* **13**, 6046 (2023).
33. Tayebi Arasteh, S. *et al.* Preserving fairness and diagnostic accuracy in private large-scale AI models for medical imaging. *Commun Med* **4**, 46 (2024).
34. Tayebi Arasteh, S. *et al.* Securing Collaborative Medical AI by Using Differential Privacy: Domain Transfer for Classification of Chest Radiographs. *Radiology. Artificial Intelligence* **6**, e230212 (2024).
35. Tayebi Arasteh, S., Isfort, P., Kuhl, C., Nebelung, S. & Truhn, D. Automatic Evaluation of Chest Radiographs – The Data Source Matters, But How Much Exactly? in *RöFo-Fortschritte auf dem Gebiet der Röntgenstrahlen und der bildgebenden Verfahren* vol. 195 ab99 (Georg Thieme Verlag, RheinMain CongressCenter (RMCC) in Wiesbaden, 2023).
36. Hu, E. J. *et al.* Lora: Low-rank adaptation of large language models. *Iclr* **1**, 3 (2022).
37. Chiarenza, A. *et al.* Chest imaging using signs, symbols, and naturalistic images: a practical guide for radiologists and non-radiologists. *Insights Imaging* **10**, 114 (2019).
38. Capitanio, M. A. Pitfalls in Pediatric Chest Radiography. *Radiology* **137**, 656–656 (1980).
39. Lotfinia, M., Tayebiarasteh, A., Samiei, S., Joodaki, M. & Tayebi Arasteh, S. Boosting multi-demographic federated learning for chest radiograph analysis using general-purpose self-supervised representations. *European Journal of Radiology Artificial Intelligence* **3**, 100028 (2025).
40. Tayebi Arasteh, S. *et al.* Enhancing domain generalization in the AI-based analysis of chest radiographs with federated learning. *Sci Rep* **13**, 22576 (2023).

41. Loshchilov, I. & Hutter, F. Decoupled Weight Decay Regularization. in *Proceedings of Proceedings of Seventh International Conference on Learning Representations (ICLR) 2019* (New Orleans, LA, USA, 2019).
42. Rezaei-Dastjerdehei, M. R., Mijani, A. & Fatemizadeh, E. Addressing Imbalance in Multi-Label Classification Using Weighted Cross Entropy Loss Function. in *2020 27th National and 5th International Iranian Conference on Biomedical Engineering (ICBME)* 333–338 (IEEE, Tehran, Iran, 2020). doi:10.1109/ICBME51989.2020.9319440.
43. Ba, J. L., Kiros, J. R. & Hinton, G. E. Layer Normalization. Preprint at https://doi.org/10.48550/arXiv.1607.06450 (2016).
44. Hendrycks, D. & Gimpel, K. Gaussian Error Linear Units (GELUs). Preprint at https://doi.org/10.48550/arXiv.1606.08415 (2023).
45. Unal, I. Defining an Optimal Cut-Point Value in ROC Analysis: An Alternative Approach. *Comput Math Methods Med* **2017**, 3762651 (2017).
46. Konietschke, F. & Pauly, M. Bootstrapping and permuting paired t-test type statistics. *Stat Comput* **24**, 283–296 (2014).
47. Dietterich, T. G. Approximate Statistical Tests for Comparing Supervised Classification Learning Algorithms. *Neural Computation* **10**, 1895–1923 (1998).
48. Tayebi Arasteh, S. *et al.* RadioRAG: Online Retrieval–Augmented Generation for Radiology Question Answering. *Radiology: Artificial Intelligence* **7**, e240476 (2025).

# Supplementary Figures

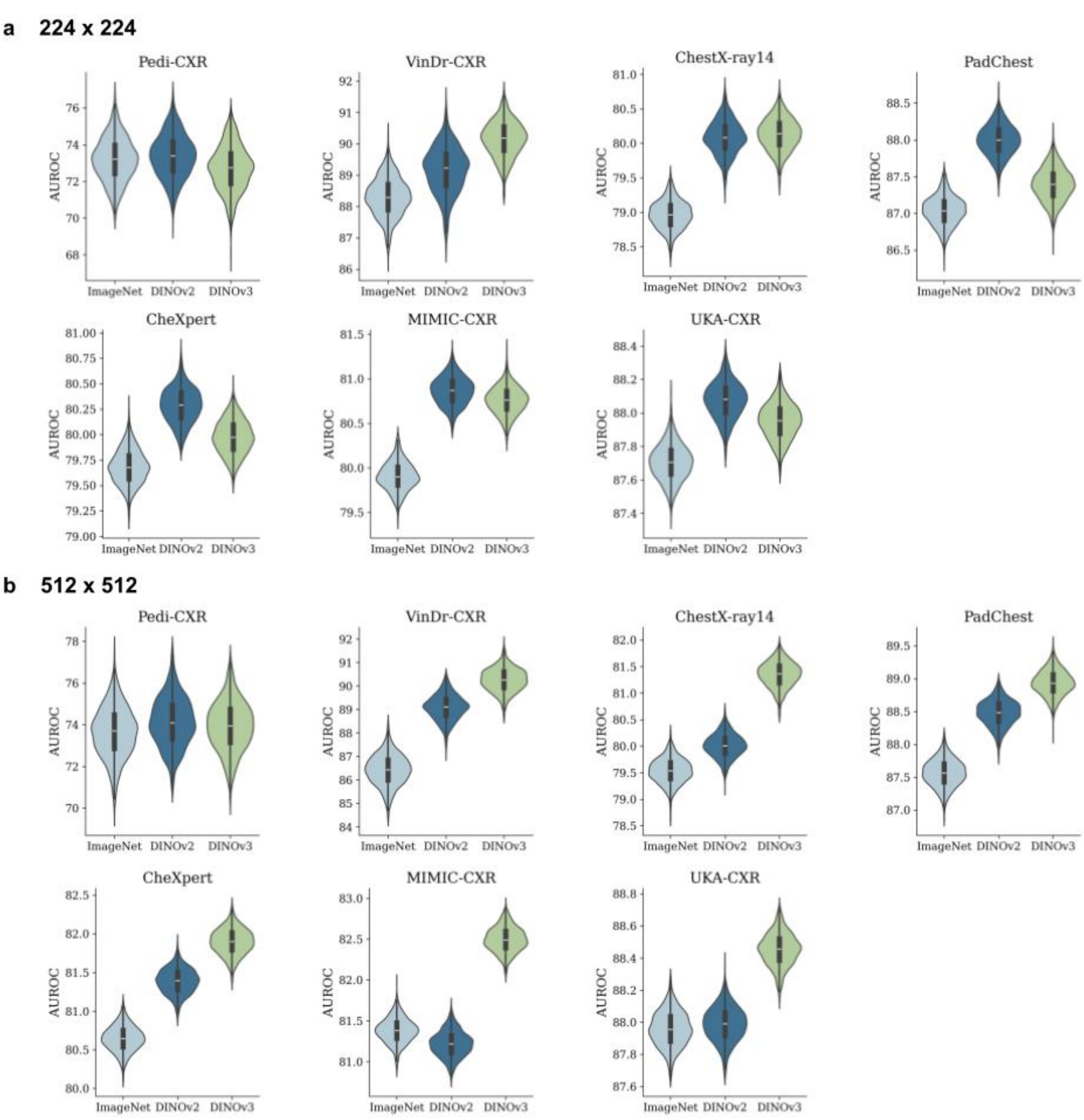

**Supplementary Figure 1**: Overall performance distributions across datasets. **a** Violin plots of bootstrap distributions (n = 1,000 resamples) for average AUROC values across all labels, comparing ImageNet, DINOv2, and DINOv3 initializations at 224 × 224 resolution with the ViT-B/16 backbone. At this resolution, DINOv2 often retained a slight edge, with DINOv3 performing comparably. **b** Corresponding bootstrap distributions at 512 × 512 resolution. Results are shown for all seven datasets: Pedi-CXR (training n=7,728; test n=1,397), VinDr-CXR (training n=15,000; test n=3,000), ChestX-ray14 (training n=77,870; validation n=8,654; test n=25,596), PadChest (training n = 79,697; validation n=8,783; test n=22,045), CheXpert (training n=115,449; validation n=13,098; test n=29,318), MIMIC-CXR (training n=153,255; validation n=18,139; test n=43,793), and UKA-CXR (training n=137,902; validation n=15,353; test n=40,106).

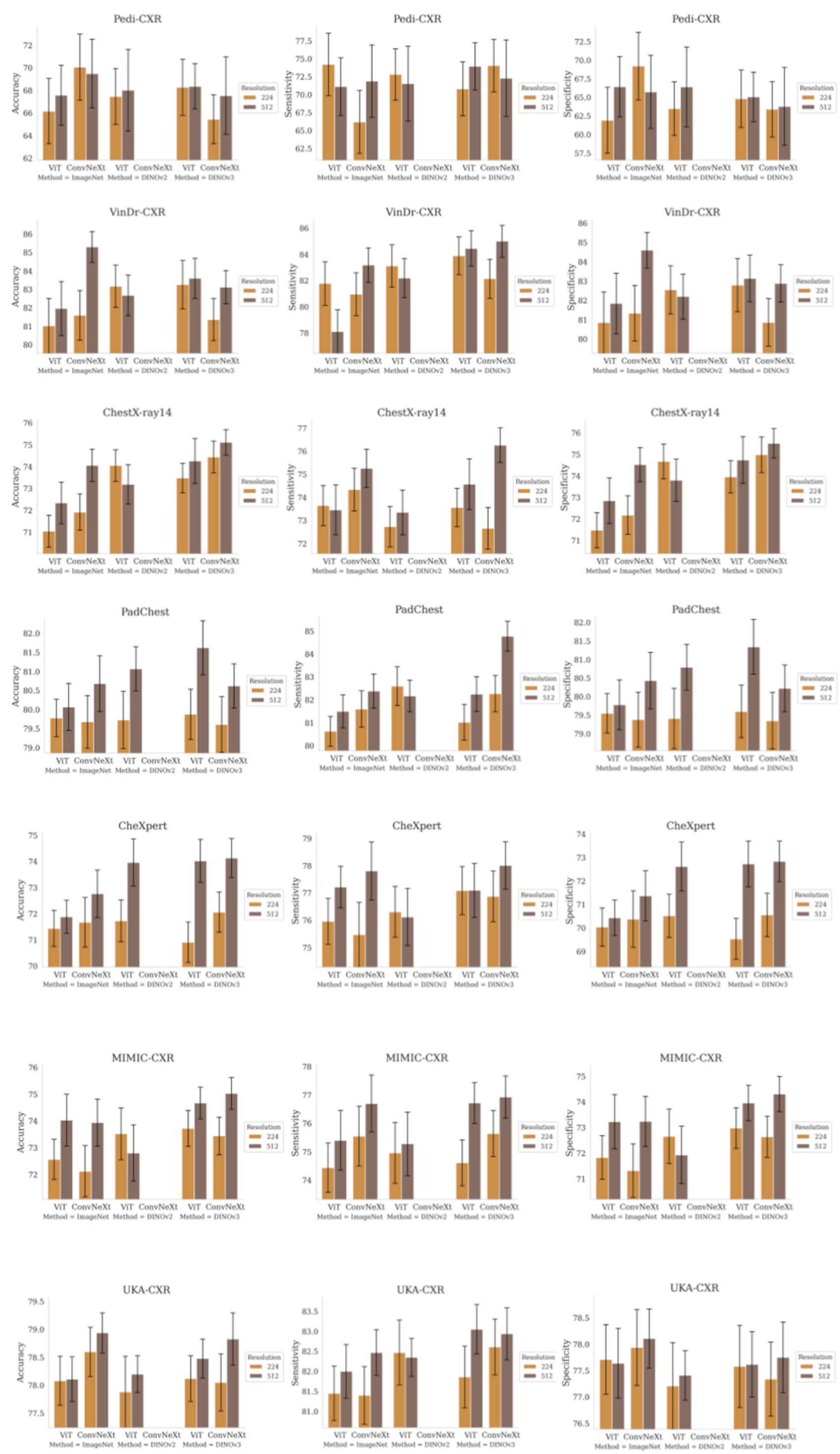

**Supplementary Figure 2:** Accuracy, sensitivity, and specificity across initialization strategies. Each panel corresponds to one dataset, showing bar plots for ImageNet, DINOv2, and DINOv3 at 224 × 224 and 512 × 512 resolutions. Bars display mean values with standard deviations from 1,000 bootstrap resamples.

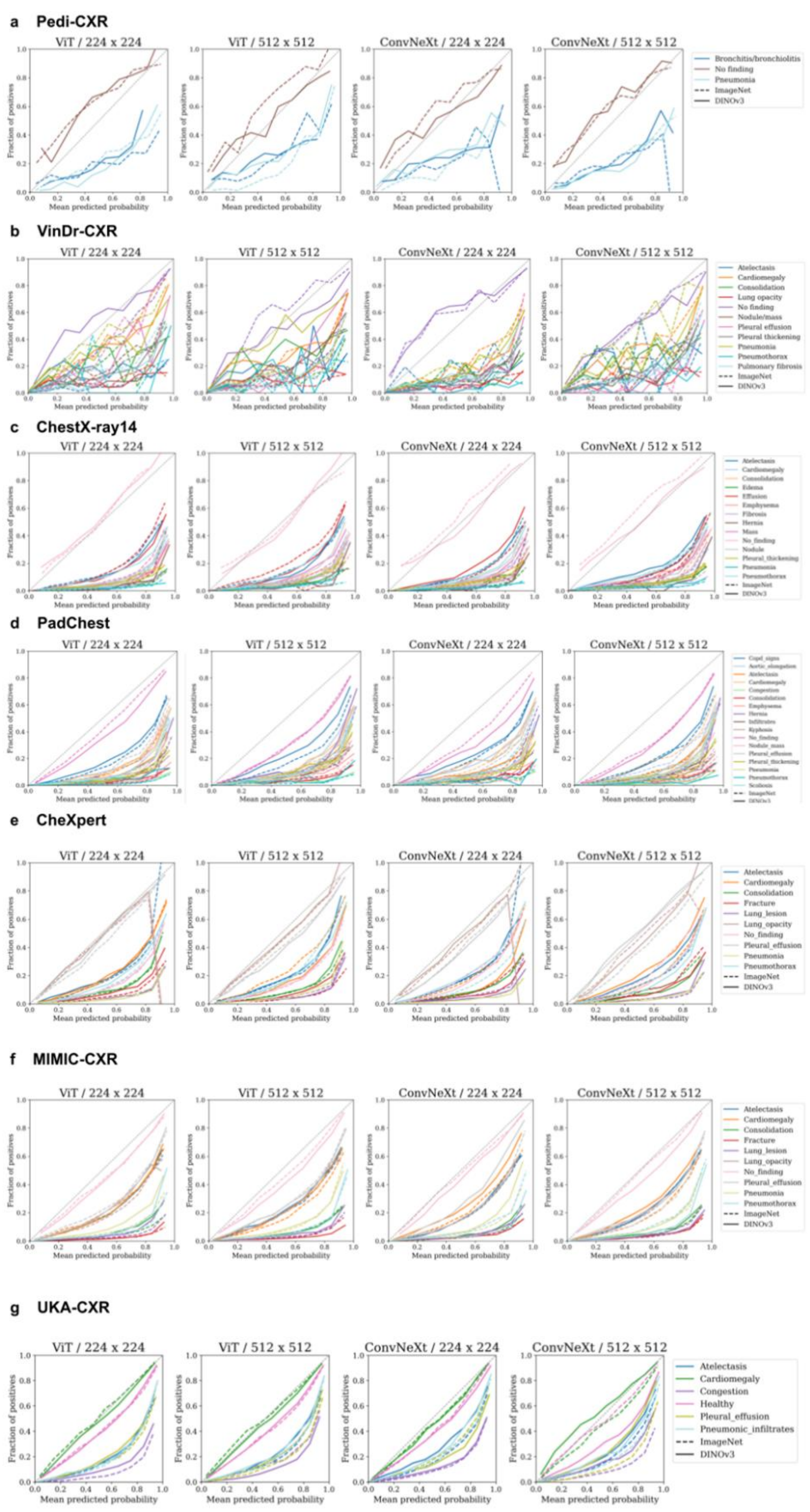

**Supplementary Figure 3:** Calibration analysis. Reliability diagrams comparing expected vs. observed probabilities for ImageNet and DINOv3 initializations under full finetuning. Plots are stratified by dataset (rows) and resolution (224 × 224, 512 × 512; columns), with separate panels for ViT-B/16 **(a)** and ConvNeXt-B **(b)**. The diagonal line indicates perfect calibration. Results are averaged across all labels within each dataset, with bootstrapped confidence shading from 1,000 resamples.

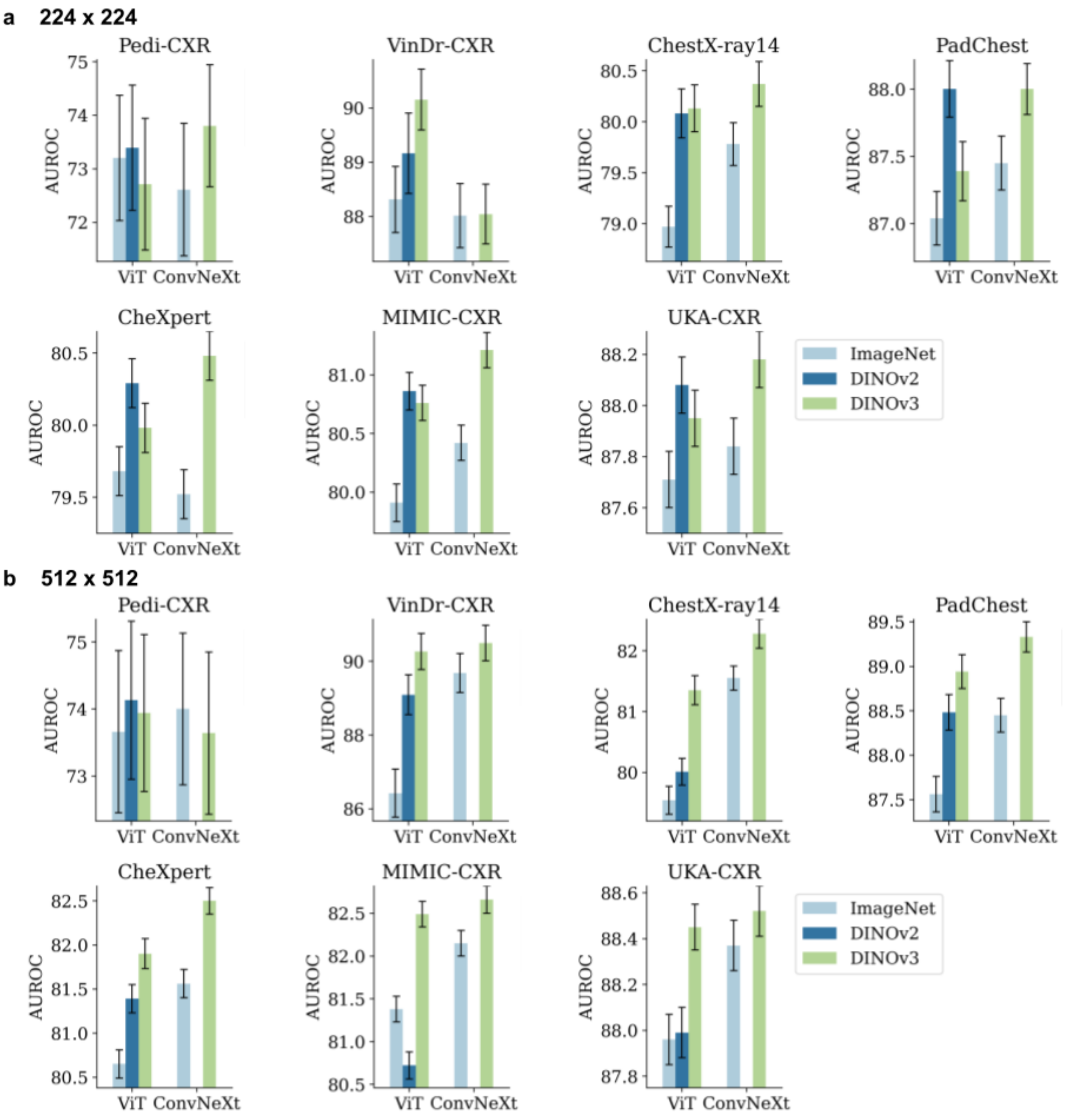

**Supplementary Figure 4:** Backbone comparison across datasets. **(a)** Mean AUROC values across all labels with standard deviations from 1,000 bootstrap resamples for ViT-B/16 and ConvNeXt-B backbones at 224 × 224 resolution. Results for ImageNet, DINOv2, and DINOv3 initializations are shown side by side within each backbone. **(b)** Corresponding results at 512 × 512 resolution. Results are shown for all seven datasets: Pedi-CXR (training n=7,728; test n=1,397), VinDr-CXR (training n=15,000; test n=3,000), ChestX-ray14 (training n=77,870; validation n=8,654; test n=25,596), PadChest (training n = 79,697; validation n=8,783; test n=22,045), CheXpert (training n=115,449; validation n=13,098; test n=29,318), MIMIC-CXR (training n=153,255; validation n=18,139; test n=43,793), and UKA-CXR (training n=137,902; validation n=15,353; test n=40,106).

**a** **224 x 224**

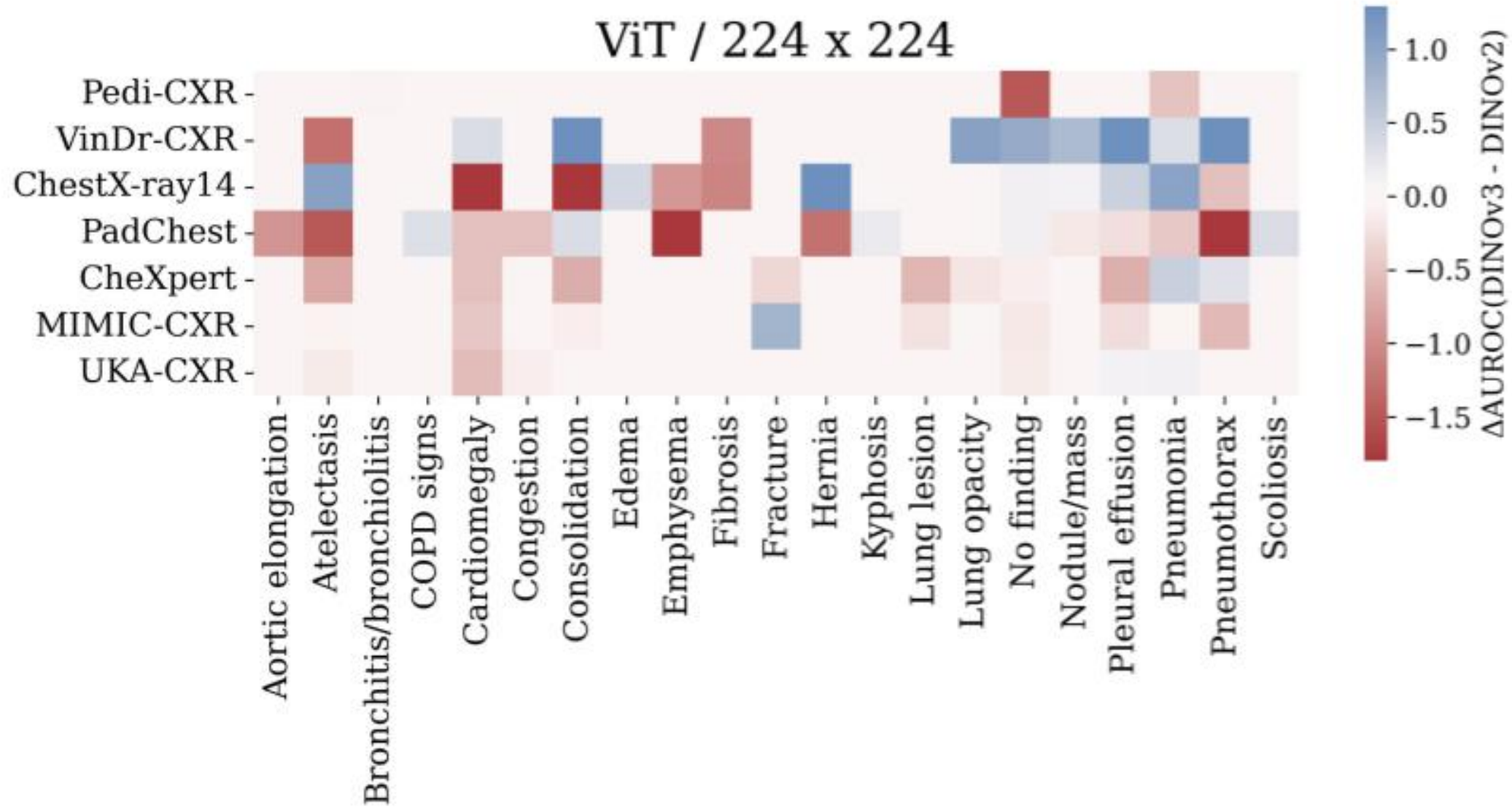

**b** **512 x 512**

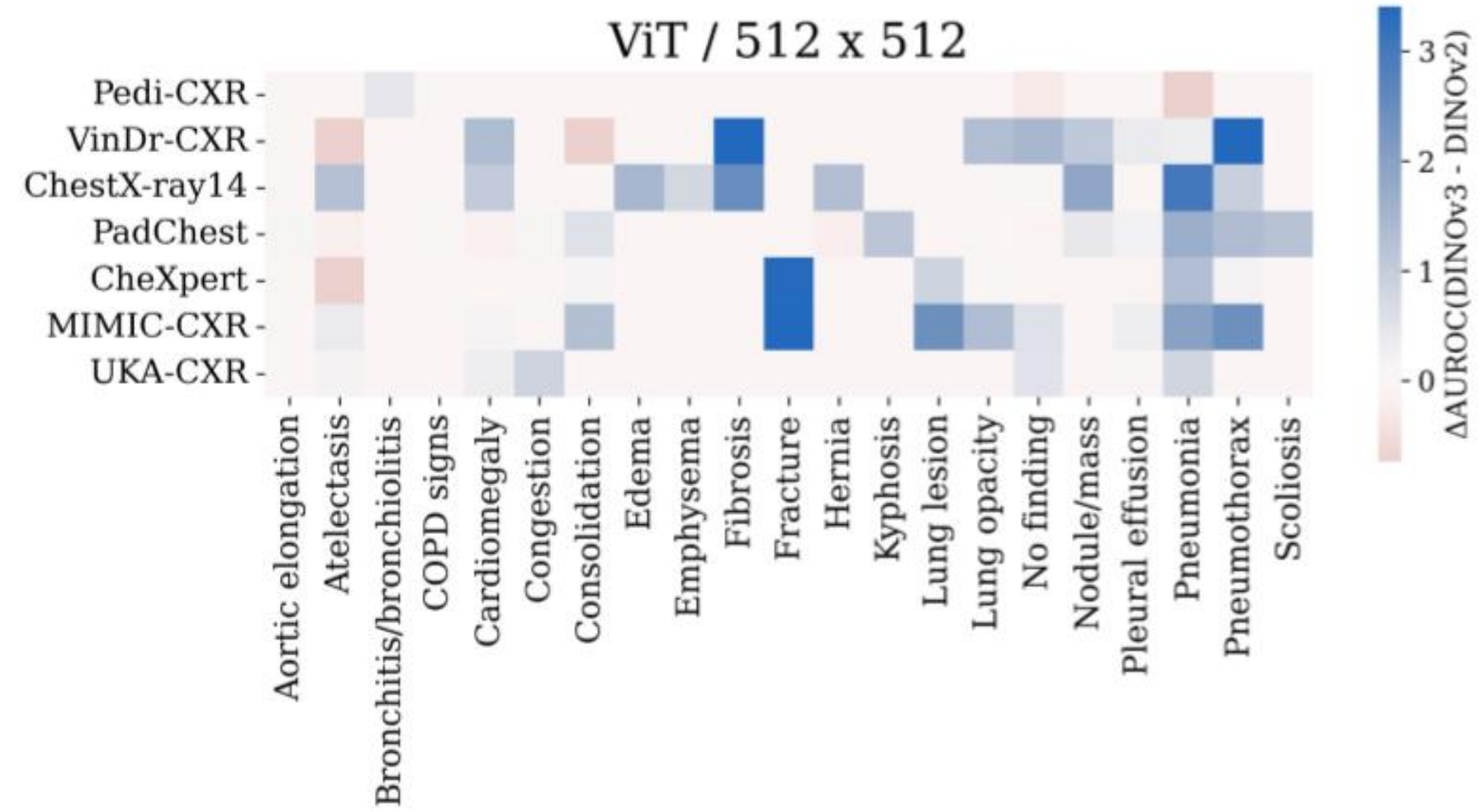

**Supplementary Figure 5**: Label-wise performance analysis. **a** Heatmap of differences in AUROC (ΔAUROC) between DINOv3 and DINOv2 across all labels and datasets at 224 × 224 resolution. Results are shown for all seven datasets: Pedi-CXR (training n=7,728; test n=1,397), VinDr-CXR (training n=15,000; test n=3,000), ChestX-ray14 (training n=77,870; validation n=8,654; test n=25,596), PadChest (training n = 79,697; validation n=8,783; test n=22,045), CheXpert (training n=115,449; validation n=13,098; test n=29,318), MIMIC-CXR (training n=153,255; validation n=18,139; test n=43,793), and UKA-CXR (training n=137,902; validation n=15,353; test n=40,106).

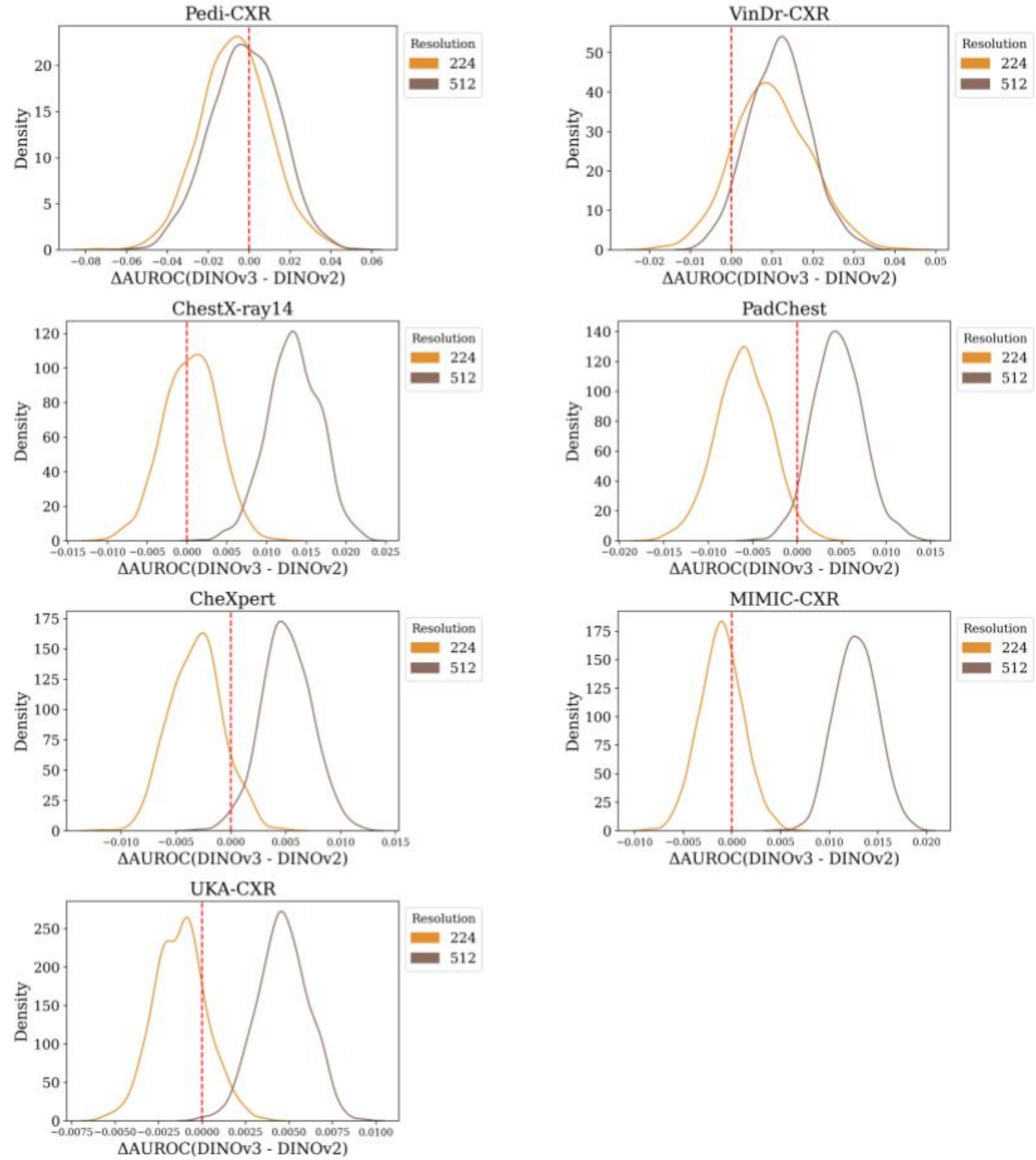

**Supplementary Figure 6:** Distribution of ΔAUROC values (DINOv3 – DINOv2) for the ViT-B backbone. Kernel density plots show bootstrap-sampled differences in average AUROC (n = 1,000 resamples) across all labels for each dataset, comparing 224 × 224 and 512 × 512 resolutions. A vertical dashed line at zero indicates no difference between methods. Results are shown for all seven datasets: Pedi-CXR (training n=7,728; test n=1,397), VinDr-CXR (training n=15,000; test n=3,000), ChestX-ray14 (training n=77,870; validation n=8,654; test n=25,596), PadChest (training n = 79,697; validation n=8,783; test n=22,045), CheXpert (training n=115,449; validation n=13,098; test n=29,318), MIMIC-CXR (training n=153,255; validation n=18,139; test n=43,793), and UKA-CXR (training n=137,902; validation n=15,353; test n=40,106).

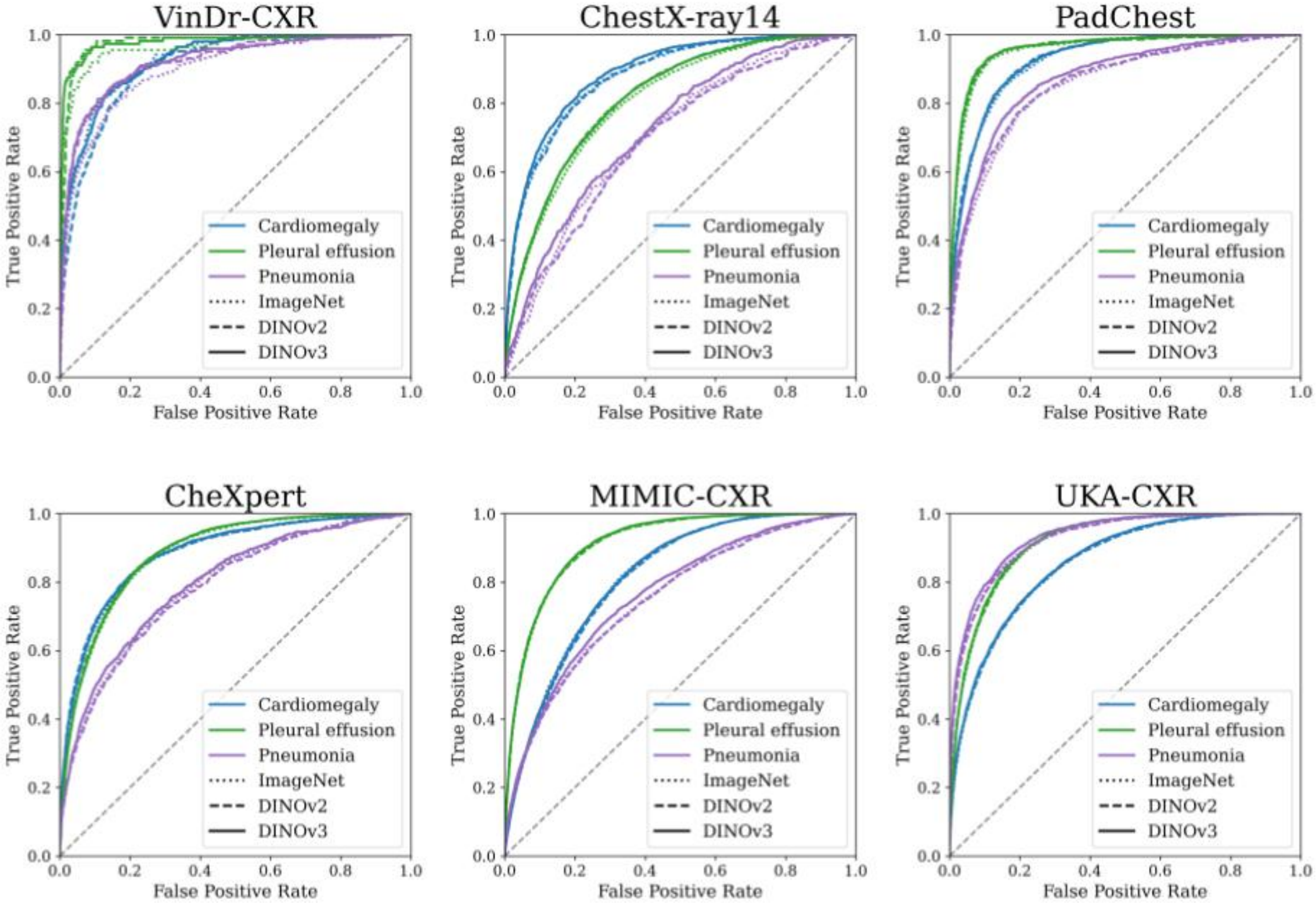

**Supplementary Figure 7:** Representative receiver operating characteristic (ROC) curves for selected labels (cardiomegaly, pleural effusion, pneumonia) at 512 × 512 resolution. Results are shown for all adult datasets: VinDr-CXR (training n=15,000; test n=3,000), ChestX-ray14 (training n=77,870; validation n=8,654; test n=25,596), PadChest (training n = 79,697; validation n=8,783; test n=22,045), CheXpert (training n=115,449; validation n=13,098; test n=29,318), MIMIC-CXR (training n=153,255; validation n=18,139; test n=43,793), and UKA-CXR (training n=137,902; validation n=15,353; test n=40,106).

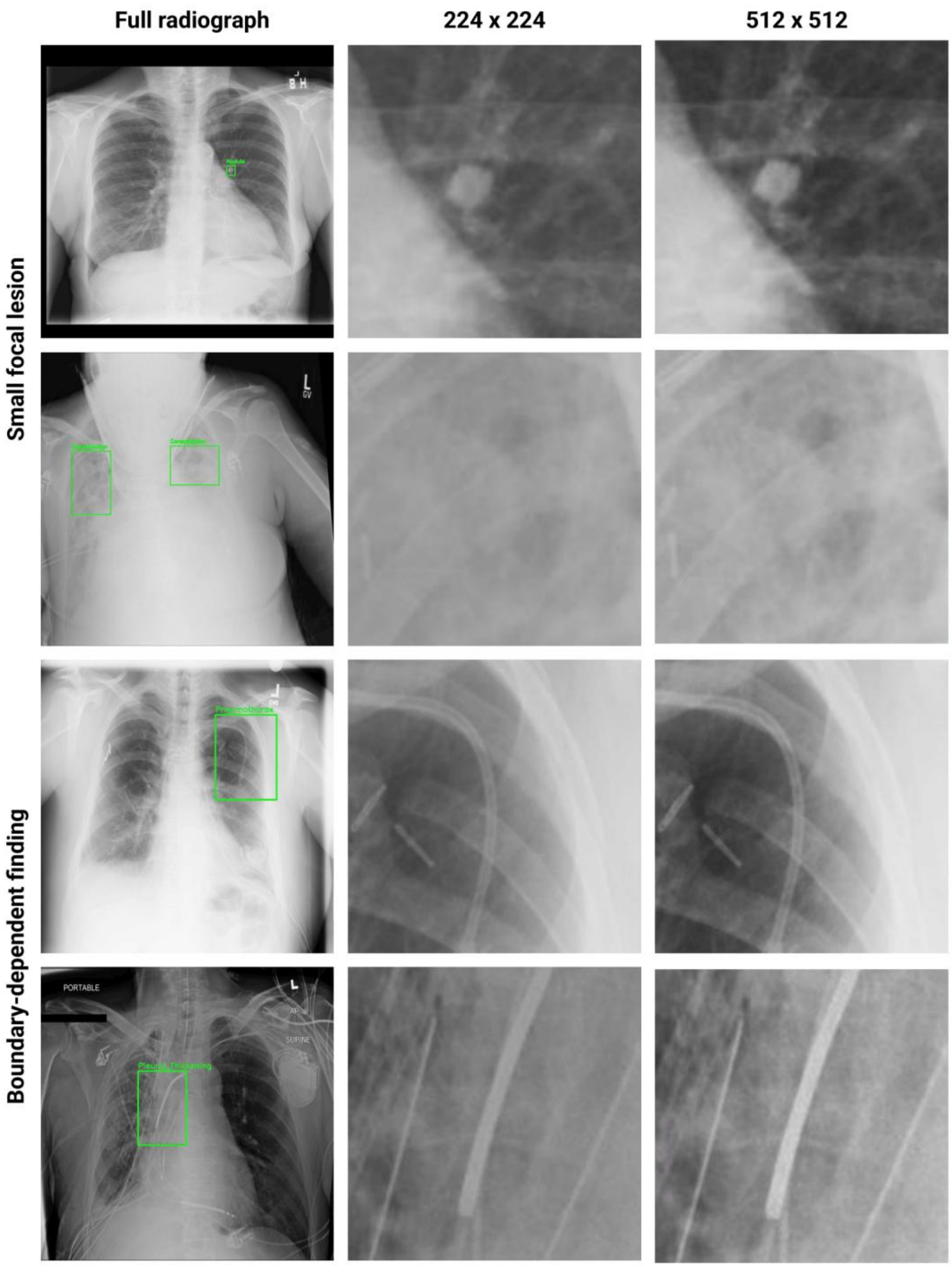

**Supplementary Figure 8:** Illustrative adult chest radiographs highlighting regions corresponding to findings with the greatest apparent benefit from higher input resolution. Shown are four representative examples from the adult ChestX-ray14 cohort, selected to visualize regions corresponding to small focal and boundary-dependent findings, the two groups that showed the largest average resolution-related gains in the quantitative cross-dataset summary (**Supplementary Table 2**). The left column shows the full radiograph with a manually chosen region of interest and the associated dataset label. The middle and right columns show the corresponding region after standard preprocessing and resizing to 224 × 224 and 512 × 512 pixels, respectively. The cropped regions from the resized images were enlarged to a common display size for visual comparison. Across these illustrative examples, the 512 × 512 versions preserve finer local structure and boundary detail more clearly than the 224 × 224 versions. This figure is intended as a qualitative complement to the quantitative analysis, not as a model-explanation figure and not as a separate performance experiment.

**a 224 x 224**

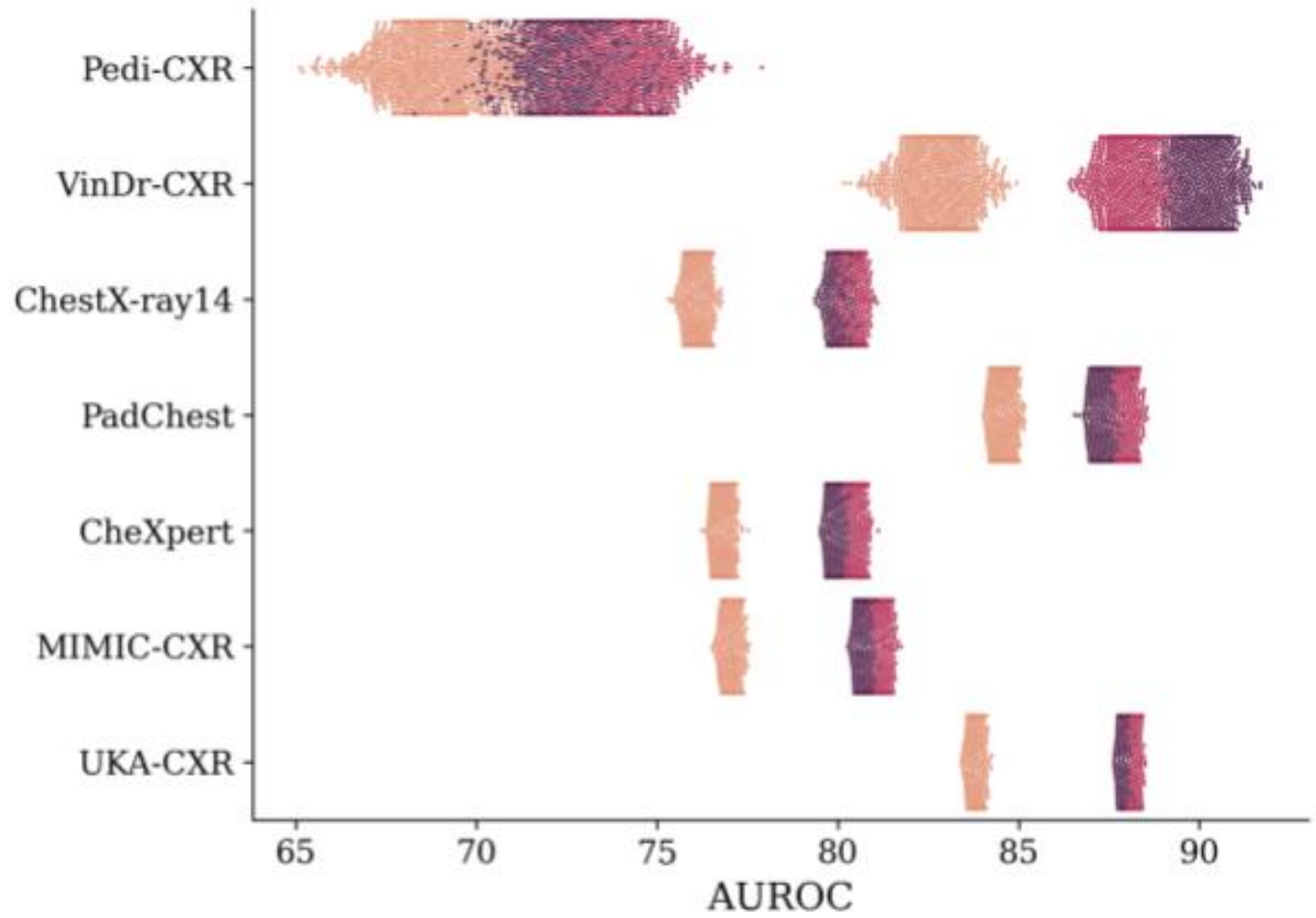

**b 512 x 512**

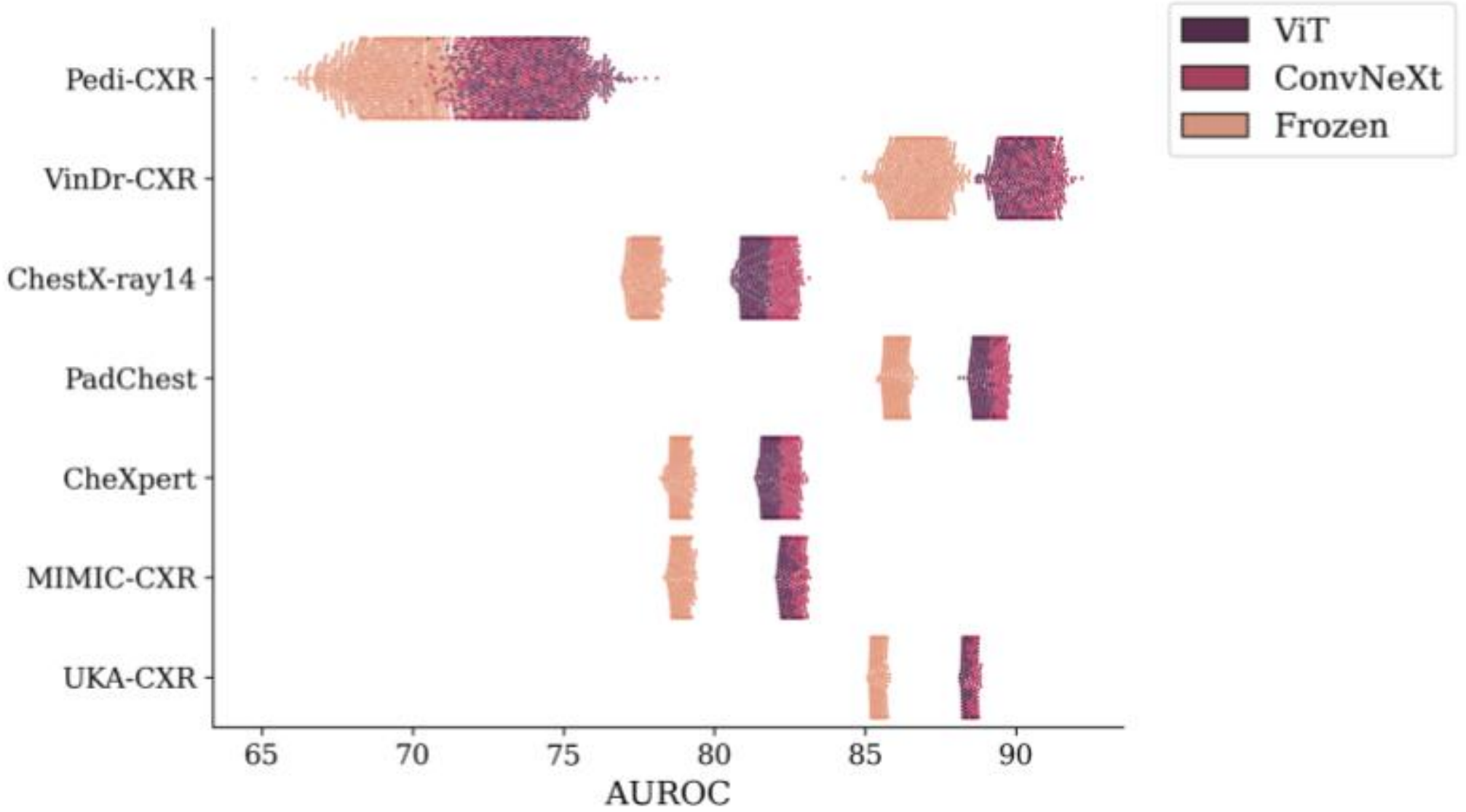

**Supplementary Figure 9**: Classification using frozen DINOv3-7B features vs. full finetuning of smaller models. Bootstrap distributions of AUROC values (n = 1,000 resamples) comparing classifiers trained on frozen DINOv3-7B features with lightweight heads (~2M parameters) against full finetuning of ViT-B/16 and ConvNeXt-B backbones (~86–87M parameters). Across datasets, full finetuning consistently outperforms frozen representations, despite the much smaller backbone size. Results are shown for all seven datasets: Pedi-CXR (training n=7,728; test n=1,397), VinDr-CXR (training n=15,000; test n=3,000), ChestX-ray14 (training n=77,870; validation n=8,654; test n=25,596), PadChest (training n = 79,697; validation n=8,783; test n=22,045), CheXpert (training n=115,449; validation n=13,098; test n=29,318), MIMIC-CXR (training n=153,255; validation n=18,139; test n=43,793), and UKA-CXR (training n=137,902; validation n=15,353; test n=40,106).

# Supplementary Tables

**Supplementary Table 1:** Pairwise bootstrap p-values across datasets, initialization strategies, and input resolutions. Two-sided p-values were obtained from paired bootstrap tests using 1,000 resampled AUROC pairs per model, ensuring identical resampling across initialization strategies for fair comparison. Results are reported separately for ViT-B/16 and ConvNeXt-B under full finetuning, as well as for frozen DINOv3-7B encoders with linear classifiers. Each comparison was performed at 224 × 224, 512 × 512, and, where available, 1024 × 1024 input resolutions. p-values were adjusted for multiple comparisons within coherent families of related tests (e.g., per-resolution comparisons across the six adult datasets) using the Benjamini-Hochberg false discovery rate (FDR) procedure, with FDR-adjusted $p < 0.05$ considered statistically significant. “N/A” indicates configurations that were not evaluated. Results are shown for all seven datasets: Pedi-CXR (training n=7,728; test n=1,397), VinDr-CXR (training n=15,000; test n=3,000), ChestX-ray14 (training n=77,870; validation n=8,654; test n=25,596), PadChest (training n=79,697; validation n=8,783; test n=22,045), CheXpert (training n=115,449; validation n=13,098; test n=29,318), MIMIC-CXR (training n=153,255; validation n=18,139; test n=43,793), and UKA-CXR (training n=137,902; validation n=15,353; test n=40,106).

| Backbone/fine tuning | Resolution | Comparison | Pedi-CXR | VinDr-CXR | ChestX-ray14 | PadChest | CheXpert | MIMIC-CXR | UKA-CXR |
|---|---|---|---|---|---|---|---|---|---|
| ViT full fine tuning | 224 x 224 | ImageNet vs. DINOv3 | 0.31 | 0.006 | 0.006 | 0.012 | 0.006 | 0.006 | 0.006 |
| | | ImageNet vs. DINOv2 | 0.61 | 0.11 | 0.006 | 0.006 | 0.006 | 0.006 | 0.006 |
| | | DINOv3 vs. DINOv2 | 0.78 | 0.050 | 0.40 | 0.003 | 0.003 | 0.20 | 0.008 |
| | 512 x 512 | ImageNet vs. DINOv3 | 0.37 | 0.006 | 0.006 | 0.006 | 0.006 | 0.006 | 0.006 |
| | | ImageNet vs. DINOv2 | 0.32 | 0.006 | 0.006 | 0.006 | 0.006 | 0.044 | 0.26 |
| | | DINOv3 vs. DINOv2 | 0.54 | 0.019 | 0.006 | 0.006 | 0.006 | 0.006 | 0.006 |
| | 1024 x 1024 | ImageNet vs. DINOv3 | 0.001 | N/A | 0.002 | N/A | N/A | 0.002 | N/A |
| | | ImageNet vs. DINOv2 | 0.001 | N/A | 0.002 | N/A | N/A | 0.40 | N/A |
| | | DINOv3 vs. DINOv2 | 0.001 | N/A | 0.002 | N/A | N/A | 0.002 | N/A |
| ConvNeXt full fine tuning | 224 x 224 | ImageNet vs. DINOv3 | 0.080 | 0.45 | 0.001 | 0.001 | 0.001 | 0.001 | 0.001 |
| | 512 x 512 | ImageNet vs. DINOv3 | 0.66 | 0.013 | 0.001 | 0.001 | 0.001 | 0.001 | 0.001 |
| | 1024 x 1024 | ImageNet vs. DINOv3 | 0.76 | N/A | 0.002 | N/A | N/A | 0.038 | N/A |
| ViT frozen features | 224 x 224 | Frozen vs. DINOv3 | 0.001 | 0.006 | 0.006 | 0.006 | 0.006 | 0.006 | 0.006 |
| | | Frozen vs. DINOv2 | 0.001 | 0.006 | 0.006 | 0.006 | 0.006 | 0.006 | 0.006 |
| | | Frozen vs. ImageNet | 0.001 | 0.006 | 0.006 | 0.006 | 0.006 | 0.006 | 0.006 |
| | 512 x 512 | Frozen vs. DINOv3 | 0.001 | 0.006 | 0.006 | 0.006 | 0.006 | 0.006 | 0.006 |
| | | Frozen vs. DINOv2 | 0.001 | 0.006 | 0.006 | 0.006 | 0.006 | 0.006 | 0.006 |
| | | Frozen vs. ImageNet | 0.001 | 0.23 | 0.006 | 0.006 | 0.006 | 0.006 | 0.006 |
| ConvNeXt frozen features | 224 x 224 | Frozen vs. DINOv3 | 0.001 | 0.006 | 0.006 | 0.006 | 0.006 | 0.006 | 0.006 |
| | | Frozen vs. ImageNet | 0.001 | 0.006 | 0.006 | 0.006 | 0.006 | 0.006 | 0.006 |
| | 512 x 512 | Frozen vs. DINOv3 | 0.001 | 0.006 | 0.006 | 0.006 | 0.006 | 0.006 | 0.006 |
| | | Frozen vs. ImageNet | 0.001 | 0.006 | 0.006 | 0.006 | 0.006 | 0.006 | 0.006 |

**Supplementary Table 2:** Compact cross-dataset per-label and per-finding-group summary. Summary results are shown for selected clinically important labels and grouped abnormality categories across the evaluated datasets. For each row, the table reports the datasets contributing to the summary, the best observed AUROC at 224 × 224 and 512 × 512 resolution, the corresponding change in AUROC between resolutions, and the difference between DINOv3 and DINOv2 at 224 × 224 and 512 × 512. AUROC values are reported as mean ± SD [95% CI] across contributing datasets. For each dataset and resolution, the best AUROC denotes the highest observed value among the evaluated model configurations. Differences between DINOv3 and DINOv2 were computed using architecture-matched ViT-B models, as DINOv2 results were available only for ViT-B. For grouped rows, clinically harmonized labels were first averaged within each dataset and then summarized across datasets.

| Label or finding group | Label group category | Datasets included | Best AUROC at 224 × 224 | Best AUROC at 512 × 512 | ΔAUROC (best 512 − best 224) | ΔAUROC (DINOv3 − DINOv2 at 224 × 224) | ΔAUROC (DINOv3 − DINOv2 at 512 × 512) |
|---|---|---|---|---|---|---|---|
| Cardiomegaly | Large-structure | UKA-CXR, CheXpert, ChestX-ray14, MIMIC-CXR, PadChest, VinDr-CXR | 88.7 ± 4.7 [83.8, 93.6] | 88.7 ± 4.4 [84.1, 93.4] | 0.0 ± 0.5 [-0.5, 0.5] | -0.7 ± 0.9 [-1.6, 0.3] | 0.5 ± 0.6 [-0.2, 1.1] |
| Pleural effusion | Large-structure | UKA-CXR, CheXpert, ChestX-ray14, MIMIC-CXR, PadChest, VinDr-CXR | 91.2 ± 5.6 [85.3, 97.1] | 91.6 ± 5.3 [86.1, 97.2] | 0.4 ± 0.4 [0.0, 0.9] | 0.1 ± 0.7 [-0.6, 0.8] | 0.1 ± 0.2 [-0.1, 0.3] |
| Pneumonia | Diffuse / textural | UKA-CXR, CheXpert, ChestX-ray14, MIMIC-CXR, PadChest, Pediatric CXR, VinDr-CXR | 82.5 ± 8.4 [74.7, 90.3] | 83.7 ± 7.9 [76.4, 91.0] | 1.2 ± 0.6 [0.7, 1.7] | 0.1 ± 0.5 [-0.4, 0.6] | 1.1 ± 1.2 [0.1, 2.2] |
| Atelectasis | Diffuse / textural | UKA-CXR, CheXpert, ChestX-ray14, MIMIC-CXR, PadChest, VinDr-CXR | 81.1 ± 6.7 [74.1, 88.1] | 82.7 ± 6.9 [75.5, 89.9] | 1.6 ± 1.0 [0.5, 2.7] | -0.4 ± 0.9 [-1.4, 0.5] | -0.3 ± 1.2 [-1.5, 1.0] |
| Consolidation | Diffuse / textural | CheXpert, ChestX-ray14, MIMIC-CXR, PadChest, VinDr-CXR | 82.8 ± 7.9 [72.9, 92.7] | 83.7 ± 8.2 [73.5, 93.9] | 0.9 ± 0.4 [0.4, 1.4] | -0.1 ± 1.3 [-1.7, 1.5] | -0.2 ± 1.4 [-2.0, 1.6] |
| Pneumothorax | Boundary-dependent | CheXpert, ChestX-ray14, MIMIC-CXR, PadChest, VinDr-CXR | 89.2 ± 2.8 [85.7, 92.6] | 91.5 ± 3.1 [87.6, 95.3] | 2.3 ± 0.9 [1.2, 3.4] | 0.2 ± 2.5 [-2.9, 3.3] | 1.8 ± 1.7 [-0.3, 4.0] |
| No finding | Normal / reference | UKA-CXR, CheXpert, ChestX-ray14, MIMIC-CXR, PadChest | 84.2 ± 6.3 [76.4, 92.0] | 84.8 ± 5.9 [77.5, 92.1] | 0.6 ± 0.4 [0.1, 1.2] | -0.0 ± 0.2 [-0.2, 0.2] | 0.1 ± 0.3 [-0.2, 0.5] |
| Small focal lesions | Small focal | CheXpert, MIMIC-CXR, ChestX-ray14, PadChest, VinDr-CXR | 78.6 ± 3.9 [73.7, 83.4] | 82.1 ± 3.1 [78.3, 85.9] | 3.5 ± 1.1 [2.2, 4.9] | 0.0 ± 0.6 [-0.6, 0.7] | 1.2 ± 0.6 [0.4, 2.0] |
| Boundary-dependent findings | Boundary-dependent | CheXpert, MIMIC-CXR, ChestX-ray14, PadChest, VinDr-CXR | 87.2 ± 3.7 [82.7, 91.8] | 89.6 ± 3.8 [84.9, 94.3] | 2.3 ± 0.7 [1.4, 3.3] | 0.3 ± 1.9 [-2.1, 2.7] | 1.7 ± 1.5 [-0.2, 3.5] |
| Diffuse / textural findings | Diffuse / textural | UKA-CXR, CheXpert, ChestX-ray14, MIMIC-CXR, PadChest, Pediatric CXR, VinDr-CXR | 81.7 ± 6.3 [75.8, 87.6] | 82.6 ± 6.3 [76.8, 88.4] | 0.9 ± 0.4 [0.5, 1.3] | -0.2 ± 0.2 [-0.4, 0.0] | 0.5 ± 0.6 [0.0, 1.1] |
| Large-structure findings | Large-structure | UKA-CXR, CheXpert, ChestX-ray14, MIMIC-CXR, PadChest, VinDr-CXR | 89.9 ± 4.4 [85.4, 94.5] | 90.2 ± 4.1 [85.9, 94.5] | 0.2 ± 0.3 [-0.1, 0.6] | -0.3 ± 0.6 [-0.9, 0.3] | 0.3 ± 0.3 [-0.1, 0.7] |

**Supplementary Table 3:** Relative computational scaling across resolution transitions, backbones, and initialization strategies. Summary of computational scaling for all full fine-tuning experiments in this study, expressed as resolution-transition ratios. For each transition, metric, backbone group, and initialization group, the table reports the mean ratio ± standard deviation (SD), median ratio, minimum ratio, maximum ratio, and the number of evaluated configurations (n). Ratios are shown for both per-epoch training time and total wall-clock training time to convergence. DINOv2 ratios are reported only for ViT-B/16, as DINOv2 was not evaluated with ConvNeXt-B. Results are shown for all seven datasets: Pedi-CXR (training n=7,728; test n=1,397), VinDr-CXR (training n=15,000; test n=3,000), ChestX-ray14 (training n=77,870; validation n=8,654; test n=25,596), PadChest (training n=79,697; validation n=8,783; test n=22,045), CheXpert (training n=115,449; validation n=13,098; test n=29,318), MIMIC-CXR (training n=153,255; validation n=18,139; test n=43,793), and UKA-CXR (training n=137,902; validation n=15,353; test n=40,106).

| Transition | Metric | Backbone | Initialization | Ratio (mean ± SD) | Ratio (Median | Ratio (Minimum) | Ratio (Maximum) | n |
|---|---|---|---|---|---|---|---|---|
| 224 → 512 | Epoch time | All | All | 3.3 ± 1.3 | 2.8 | 1.6 | 7.0 | 35 |
| 224 → 512 | Epoch time | ViT-B/16 | All | 3.5 ± 1.4 | 3.0 | 1.6 | 7.0 | 21 |
| 224 → 512 | Epoch time | ConvNeXt-B | All | 3.0 ± 1.0 | 2.5 | 2.1 | 4.8 | 14 |
| 224 → 512 | Epoch time | ViT-B/16 | ImageNet | 3.1 ± 1.2 | 3.0 | 1.6 | 5.2 | 7 |
| 224 → 512 | Epoch time | ViT-B/16 | DINOv2 | 3.7 ± 1.8 | 2.8 | 2.4 | 7.0 | 7 |
| 224 → 512 | Epoch time | ViT-B/16 | DINOv3 | 3.7 ± 1.3 | 3.1 | 2.7 | 6.2 | 7 |
| 224 → 512 | Epoch time | ConvNeXt-B | ImageNet | 2.9 ± 0.9 | 2.5 | 2.2 | 4.8 | 7 |
| 224 → 512 | Epoch time | ConvNeXt-B | DINOv3 | 3.0 ± 1.1 | 2.6 | 2.1 | 4.8 | 7 |
| 224 → 512 | Total time to converge | All | All | 3.6 ± 1.5 | 3.5 | 0.6 | 7.9 | 35 |
| 224 → 512 | Total time to converge | ViT-B/16 | All | 4.0 ± 1.6 | 4.1 | 0.6 | 7.9 | 21 |
| 224 → 512 | Total time to converge | ConvNeXt-B | All | 3.0 ± 1.3 | 2.8 | 1.4 | 4.8 | 14 |
| 224 → 512 | Total time to converge | ViT-B/16 | ImageNet | 4.4 ± 1.4 | 4.9 | 1.8 | 5.7 | 7 |
| 224 → 512 | Total time to converge | ViT-B/16 | DINOv2 | 4.1 ± 2.3 | 4.1 | 0.6 | 7.9 | 7 |
| 224 → 512 | Total time to converge | ViT-B/16 | DINOv3 | 3.6 ± 0.7 | 3.5 | 2.5 | 4.5 | 7 |
| 224 → 512 | Total time to converge | ConvNeXt-B | ImageNet | 2.7 ± 1.2 | 2.7 | 1.5 | 4.8 | 7 |
| 224 → 512 | Total time to converge | ConvNeXt-B | DINOv3 | 3.4 ± 1.4 | 4.0 | 1.4 | 4.8 | 7 |
| 512 → 1024 | Epoch time | All | All | 6.2 ± 4.4 | 5.1 | 2.4 | 18.8 | 15 |
| 512 → 1024 | Epoch time | ViT-B/16 | All | 7.9 ± 5.1 | 6.0 | 3.2 | 18.8 | 9 |
| 512 → 1024 | Epoch time | ConvNeXt-B | All | 3.6 ± 0.9 | 3.6 | 2.4 | 5.1 | 6 |
| 512 → 1024 | Epoch time | ViT-B/16 | ImageNet | 9.3 ± 8.3 | 6.0 | 3.2 | 18.8 | 3 |
| 512 → 1024 | Epoch time | ViT-B/16 | DINOv2 | 6.2 ± 2.4 | 6.5 | 3.7 | 8.3 | 3 |
| 512 → 1024 | Epoch time | ViT-B/16 | DINOv3 | 8.1 ± 4.5 | 5.7 | 5.4 | 13.3 | 3 |
| 512 → 1024 | Epoch time | ConvNeXt-B | ImageNet | 4.2 ± 0.8 | 3.7 | 3.7 | 5.1 | 3 |
| 512 → 1024 | Epoch time | ConvNeXt-B | DINOv3 | 3.1 ± 0.6 | 3.3 | 2.4 | 3.5 | 3 |
| 512 → 1024 | Total time to converge | All | All | 7.7 ± 5.8 | 4.8 | 2.2 | 18.9 | 15 |
| 512 → 1024 | Total time to converge | ViT-B/16 | All | 10.6 ± 5.8 | 8.9 | 4.7 | 18.9 | 9 |
| 512 → 1024 | Total time to converge | ConvNeXt-B | All | 3.3 ± 1.1 | 2.9 | 2.2 | 4.8 | 6 |
| 512 → 1024 | Total time to converge | ViT-B/16 | ImageNet | 11.1 ± 5.5 | 14.0 | 4.8 | 14.6 | 3 |
| 512 → 1024 | Total time to converge | ViT-B/16 | DINOv2 | 13.9 ± 7.9 | 18.0 | 4.7 | 18.9 | 3 |
| 512 → 1024 | Total time to converge | ViT-B/16 | DINOv3 | 6.7 ± 1.9 | 5.7 | 5.4 | 8.9 | 3 |
| 512 → 1024 | Total time to converge | ConvNeXt-B | ImageNet | 3.1 ± 1.1 | 2.8 | 2.2 | 4.4 | 3 |
| 512 → 1024 | Total time to converge | ConvNeXt-B | DINOv3 | 3.5 ± 1.2 | 3.0 | 2.5 | 4.8 | 3 |
| 224 → 1024 | Epoch time | All | All | 17.4 ± 10.3 | 16.4 | 7.2 | 40.4 | 15 |
| 224 → 1024 | Epoch time | ViT-B/16 | All | 22.4 ± 10.6 | 18.2 | 9.6 | 40.4 | 9 |
| 224 → 1024 | Epoch time | ConvNeXt-B | All | 10.0 ± 2.0 | 10.1 | 7.2 | 12.1 | 6 |
| 224 → 1024 | Epoch time | ViT-B/16 | ImageNet | 22.1 ± 16.2 | 16.4 | 9.6 | 40.4 | 3 |
| 224 → 1024 | Epoch time | ViT-B/16 | DINOv2 | 19.7 ± 1.5 | 20.1 | 18 | 20.8 | 3 |
| 224 → 1024 | Epoch time | ViT-B/16 | DINOv3 | 25.3 ± 12.7 | 18.2 | 17.8 | 40.0 | 3 |
| 224 → 1024 | Epoch time | ConvNeXt-B | ImageNet | 11.2 ± 1.6 | 12.0 | 9.4 | 12.1 | 3 |
| 224 → 1024 | Epoch time | ConvNeXt-B | DINOv3 | 8.8 ± 1.9 | 8.3 | 7.2 | 10.9 | 3 |
| 224 → 1024 | Total time to converge | All | All | 29.7 ± 26.3 | 20.9 | 4.6 | 94.0 | 15 |
| 224 → 1024 | Total time to converge | ViT-B/16 | All | 42.9 ± 26.4 | 37.5 | 17.8 | 94.0 | 9 |
| 224 → 1024 | Total time to converge | ConvNeXt-B | All | 9.9 ± 6.2 | 7.3 | 4.6 | 20.9 | 6 |
| 224 → 1024 | Total time to converge | ViT-B/16 | ImageNet | 40.2 ± 26.4 | 25.2 | 24.6 | 70.7 | 3 |
| 224 → 1024 | Total time to converge | ViT-B/16 | DINOv2 | 63.3 ± 28.6 | 58.5 | 37.5 | 94.0 | 3 |
| 224 → 1024 | Total time to converge | ViT-B/16 | DINOv3 | 25.3 ± 12.7 | 18.2 | 17.8 | 40.0 | 3 |
| 224 → 1024 | Total time to converge | ConvNeXt-B | ImageNet | 6.9 ± 1.0 | 6.5 | 6.1 | 8.0 | 3 |
| 224 → 1024 | Total time to converge | ConvNeXt-B | DINOv3 | 13.0 ± 8.2 | 13.6 | 4.6 | 20.9 | 3 |

**Supplementary Table 4:** Results for the optimization-regime comparison at 512 × 512 resolution. Results are reported for ChestX-ray14, CheXpert, and MIMIC-CXR, comparing full end-to-end fine-tuning (full FT) and low-rank adaptation (LoRA) for ViT-B/16 and ConvNeXt-B across ImageNet, DINOv2, and DINOv3 initialization strategies. For each dataset and configuration, the table reports mean area under the receiver operating characteristic curve (AUROC) from 1,000 bootstrap resamples as mean ± standard deviation with 95% confidence intervals (CIs). Gap rows summarize the absolute AUROC difference between optimization regimes, defined as full FT − LoRA. Results are shown for all seven datasets: Pedi-CXR (training n=7,728; test n=1,397), VinDr-CXR (training n=15,000; test n=3,000), ChestX-ray14 (training n=77,870; validation n=8,654; test n=25,596), PadChest (training n=79,697; validation n=8,783; test n=22,045), CheXpert (training n=115,449; validation n=13,098; test n=29,318), MIMIC-CXR (training n=153,255; validation n=18,139; test n=43,793), and UKA-CXR (training n=137,902; validation n=15,353; test n=40,106).

| Dataset | Initialization | ChestX-ray14 | CheXpert | MIMIC-CXR |
|---|---|---|---|---|
| ViT-B/16 full FT | ImageNet | 79.5 ± 0.2 [79.1, 80.0] | 80.7 ± 0.2 [80.3, 81.0] | 81.4 ± 0.1 [81.1, 81.7] |
| ViT-B/16 LoRA | ImageNet | 76.1 ± 0.2 [75.7, 76.5] | 77.7 ± 0.2 [77.4, 78.0] | 77.1 ± 0.2 [76.8, 77.5] |
| Gap: ViT-B/16 full FT − ViT-B/16 LoRA | ImageNet | 3.4 ± 0.3 [2.8, 4.0] | 3.0 ± 0.2 [2.5, 3.4] | 4.2 ± 0.2 [3.8, 4.7] |
| ViT-B/16 full FT | DINOv3 | 81.3 ± 0.2 [80.9, 81.8] | 81.9 ± 0.2 [81.6, 82.2] | 82.5 ± 0.1 [82.2, 82.8] |
| ViT-B/16 LoRA | DINOv3 | 76.3 ± 0.2 [75.8, 76.7] | 77.6 ± 0.2 [77.2, 77.9] | 78.4 ± 0.2 [78.1, 78.7] |
| Gap: ViT-B/16 full FT − ViT-B/16 LoRA | DINOv3 | 5.1 ± 0.3 [4.4, 5.7] | 4.3 ± 0.2 [3.8, 4.8] | 4.1 ± 0.2 [3.7, 4.5] |
| ViT-B/16 full FT | DINOv2 | 80.0 ± 0.2 [79.6, 80.4] | 81.4 ± 0.2 [81.1, 81.7] | 81.2 ± 0.1 [80.9, 81.5] |
| ViT-B/16 LoRA | DINOv2 | 78.7 ± 0.2 [78.3, 79.1] | 79.2 ± 0.2 [78.8, 79.5] | 79.5 ± 0.1 [79.2, 79.8] |
| Gap: ViT-B/16 full FT − ViT-B/16 LoRA | DINOv2 | 1.3 ± 0.3 [0.7, 1.9] | 2.2 ± 0.2 [1.8, 2.7] | 1.7 ± 0.2 [1.3, 2.1] |
| ConvNeXt-B full FT | ImageNet | 81.5 ± 0.2 [81.1, 81.9] | 81.6 ± 0.2 [81.2, 81.9] | 82.2 ± 0.1 [81.9, 82.4] |
| ConvNeXt-B LoRA | ImageNet | 77.5 ± 0.2 [77.0, 78.0] | 78.6 ± 0.2 [78.2, 78.9] | 78.8 ± 0.2 [78.5, 79.1] |
| Gap: ConvNeXt-B full FT − ConvNeXt-B LoRa | ImageNet | 4.1 ± 0.3 [3.4, 4.7] | 3.0 ± 0.2 [2.6, 3.4] | 3.4 ± 0.2 [3.0, 3.8] |
| ConvNeXt-B full FT | DINOv3 | 82.3 ± 0.2 [81.8, 82.7] | 82.5 ± 0.1 [82.2, 82.8] | 82.7 ± 0.2 [82.3, 83.0] |
| ConvNeXt-B LoRA | DINOv3 | 79.4 ± 0.2 [79.0, 79.9] | 79.2 ± 0.2 [78.9, 79.5] | 80.6 ± 0.2 [80.3, 80.9] |
| Gap: ConvNeXt-B full FT − ConvNeXt-B LoRA | DINOv3 | 2.8 ± 0.3 [2.2, 3.5] | 3.3 ± 0.2 [2.9, 3.7] | 2.1 ± 0.2 [1.6, 2.5] |

**Supplementary Table 5:** Exact results for the controlled synthetic label-noise experiment on the VinDr-CXR (training n=15,000; test n=3,000) dataset at 512 × 512 resolution. Results are shown for ViT-B/16 models initialized from ImageNet, DINOv2, and DINOv3 across increasing levels of synthetic training-label corruption from 0% to 40%. For each noise level and initialization strategy, the table reports area under the receiver operating characteristic curve (AUROC), precision-recall area under the curve (PR-AUC), mean average precision (mAP), and retained AUROC relative to the corresponding clean-label (0% noise) setting. All values are reported as mean ± standard deviation with 95% confidence intervals (CIs), based on 1,000 bootstrap resamples.

| Metric / Configuration | 0% noise | 10% noise | 20% noise | 30% noise | 40% noise |
|---|---|---|---|---|---|
| ImageNet AUROC | 86.4 ± 0.6 [85.2, 87.5] | 82.8 ± 0.9 [81.0, 84.6] | 77.2 ± 1.0 [75.3, 79.3] | 78.8 ± 0.9 [77.1, 80.6] | 80.7 ± 0.9 [78.8, 82.4] |
| DINOv2 AUROC | 89.0 ± 0.6 [88.0, 90.2] | 86.1 ± 0.7 [84.8, 87.3] | 88.6 ± 0.6 [87.3, 89.9] | 81.6 ± 1.0 [79.6, 83.7] | 80.3 ± 1.0 [78.2, 82.2] |
| DINOv3 AUROC | 90.3 ± 0.5 [89.3, 91.2] | 84.7 ± 0.8 [83.1, 86.1] | 84.6 ± 0.8 [83.1, 86.0] | 78.9 ± 1.0 [77.0, 80.8] | 77.9 ± 0.9 [75.9, 79.6] |
| ImageNet PR-AUC | 41.9 ± 1.2 [39.5, 44.5] | 37.7 ± 1.1 [35.6, 39.9] | 31.6 ± 1.0 [29.6, 33.6] | 31.3 ± 1.0 [29.4, 33.1] | 32.8 ± 0.9 [31.0, 34.7] |
| DINOv2 PR-AUC | 41.3 ± 1.1 [39.0, 43.4] | 35.9 ± 1.0 [33.8, 37.8] | 42.2 ± 1.1 [40.1, 44.4] | 38.0 ± 1.0 [35.9, 39.9] | 36.4 ± 1.1 [34.3, 38.5] |
| DINOv3 PR-AUC | 47.7 ± 1.4 [45.0, 50.3] | 40.8 ± 1.1 [38.6, 42.9] | 37.3 ± 1.0 [35.4, 39.1] | 34.7 ± 1.1 [32.7, 36.8] | 28.5 ± 0.9 [26.7, 30.3] |
| ImageNet mAP | 42.5 ± 1.2 [40.1, 45.1] | 38.2 ± 1.1 [36.1, 40.4] | 32.0 ± 1.0 [30.1, 34.1] | 31.8 ± 1.0 [29.9, 33.7] | 33.4 ± 1.0 [31.5, 35.2] |
| DINOv2 mAP | 41.9 ± 1.1 [39.6, 44.0] | 36.5 ± 1.0 [34.4, 38.5] | 42.8 ± 1.1 [40.6, 45.0] | 38.5 ± 1.1 [36.4, 40.6] | 37.0 ± 1.1 [34.8, 39.2] |
| DINOv3 mAP | 48.3 ± 1.4 [45.7, 50.8] | 41.4 ± 1.2 [39.1, 43.9] | 37.8 ± 1.0 [35.9, 39.7] | 35.2 ± 1.1 [33.1, 37.3] | 29.0 ± 0.9 [27.2, 30.9] |
| ImageNet retained performance vs 0% | 100.0 ± 0.0 [100.0, 100.0] | 95.9 ± 1.1 [93.7, 98.0] | 89.4 ± 1.2 [87.2, 91.8] | 91.2 ± 1.0 [89.2, 93.3] | 93.5 ± 1.1 [91.2, 95.4] |
| DINOv2 retained performance vs 0% | 100.0 ± 0.0 [100.0, 100.0] | 96.7 ± 0.7 [95.2, 98.0] | 99.5 ± 0.7 [98.0, 100.9] | 91.7 ± 1.1 [89.4, 93.9] | 90.2 ± 1.1 [87.9, 92.3] |
| DINOv3 retained performance vs 0% | 100.0 ± 0.0 [100.0, 100.0] | 93.8 ± 0.8 [92.0, 95.4] | 93.7 ± 0.8 [92.1, 95.3] | 87.4 ± 1.1 [85.3, 89.5] | 86.3 ± 1.0 [84.1, 88.2] |

**Supplementary Table 6:** Comparison of frozen DINOv3-7B features vs. task-adapted mid-sized models across datasets and resolutions. Average area under the receiver operating characteristic curve (AUROC) derived from 1,000 bootstrap resamples for frozen DINOv3-7B features with lightweight classification heads, compared against the best-performing fully adapted ViT-B/16 and ConvNeXt-B models obtained under full fine-tuning. Results are shown for all seven datasets: Pedi-CXR (training n=7,728; test n=1,397), VinDr-CXR (training n=15,000; test n=3,000), ChestX-ray14 (training n=77,870; validation n=8,654; test n=25,596), PadChest (training n=79,697; validation n=8,783; test n=22,045), CheXpert (training n=115,449; validation n=13,098; test n=29,318), MIMIC-CXR (training n=153,255; validation n=18,139; test n=43,793), and UKA-CXR (training n=137,902; validation n=15,353; test n=40,106), at the universal input resolutions of 224 × 224 and 512 × 512 pixels. For each dataset and resolution, the table reports the frozen DINOv3-7B AUROC as mean ± standard deviation with 95% confidence intervals (CIs), together with the best fine-tuned ViT-B/16 AUROC, the best fine-tuned ConvNeXt-B AUROC, and the identity of the best overall task-adapted model.

| Dataset | Resolution | Frozen DINOv3-7B AUROC (mean ± std [95% CI] | Best fine-tuned ViT-B/16 AUROC | Best fine-tuned ConvNeXt-B AUROC | Best overall task-adapted model |
|---|---|---|---|---|---|
| Pedi-CXR | 224 × 224 | 68.8 ± 1.3 [66.4, 71.3] | 73.4 ± 1.2 [71.1, 75.6] | 73.8 ± 1.1 [71.6, 76.0] | ConvNeXt-B (DINOv3) |
| Pedi-CXR | 512 × 512 | 69.4 ± 1.3 [66.9, 71.8] | 74.1 ± 1.2 [72.0, 76.4] | 74.0 ± 1.1 [71.8, 76.2] | ViT-B/16 (DINOv2) |
| VinDr-CXR | 224 × 224 | 82.8 ± 0.8 [81.2, 84.3] | 90.2 ± 0.6 [89.0, 91.2] | 88.0 ± 0.6 [87.0, 89.1] | ViT-B/16 (DINOv3) |
| VinDr-CXR | 512 × 512 | 86.8 ± 0.6 [85.5, 88.0] | 90.3 ± 0.5 [89.3, 91.2] | 90.5 ± 0.5 [89.5, 91.5] | ConvNeXt-B (DINOv3) |
| ChestX-ray14 | 224 × 224 | 76.1 ± 0.2 [75.7, 76.5] | 80.1 ± 0.2 [79.7, 80.6] | 80.4 ± 0.2 [80.0, 80.8] | ConvNeXt-B (DINOv3) |
| ChestX-ray14 | 512 × 512 | 77.7 ± 0.3 [77.1, 78.2] | 81.3 ± 0.2 [80.9, 81.8] | 82.3 ± 0.2 [81.8, 82.7] | ConvNeXt-B (DINOv3) |
| PadChest | 224 × 224 | 84.6 ± 0.2 [84.2, 85.0] | 88.0 ± 0.2 [87.6, 88.4] | 88.0 ± 0.2 [87.6, 88.3] | ViT-B/16 (DINOv2) |
| PadChest | 512 × 512 | 86.0 ± 0.2 [85.7, 86.4] | 88.9 ± 0.2 [88.6, 89.3] | 89.3 ± 0.2 [89.0, 89.7] | ConvNeXt-B (DINOv3) |
| CheXpert | 224 × 224 | 76.8 ± 0.2 [76.5, 77.2] | 80.3 ± 0.2 [80.0, 80.6] | 80.5 ± 0.2 [80.2, 80.8] | ConvNeXt-B (DINOv3) |
| CheXpert | 512 × 512 | 78.9 ± 0.2 [78.5, 79.2] | 81.9 ± 0.2 [81.6, 82.2] | 82.5 ± 0.1 [82.2, 82.8] | ConvNeXt-B (DINOv3) |
| MIMIC-CXR | 224 × 224 | 77.1 ± 0.2 [76.8, 77.4] | 80.9 ± 0.2 [80.6, 81.2] | 81.2 ± 0.1 [80.9, 81.5] | ConvNeXt-B (DINOv3) |
| MIMIC-CXR | 512 × 512 | 78.9 ± 0.2 [78.6, 79.2] | 82.5 ± 0.1 [82.2, 82.8] | 82.7 ± 0.2 [82.3, 83.0] | ConvNeXt-B (DINOv3) |
| UKA-CXR | 224 × 224 | 83.8 ± 0.1 [83.6, 84.1] | 88.1 ± 0.1 [87.9, 88.3] | 88.2 ± 0.1 [88.0, 88.4] | ConvNeXt-B (DINOv3) |
| UKA-CXR | 512 × 512 | 85.4 ± 0.1 [85.2, 85.7] | 88.5 ± 0.1 [88.2, 88.7] | 88.5 ± 0.1 [88.3, 88.7] | ConvNeXt-B (DINOv3) |